\documentclass[conference]{IEEEtran}

\usepackage{cite}

\ifCLASSINFOpdf
\else
\fi
\usepackage{amsmath}
\usepackage{amssymb}
\usepackage{todonotes}
\usepackage{caption}
\usepackage{subcaption}
\usepackage{verbatim}
\usepackage{multicol}
\usepackage[ruled,linesnumbered]{algorithm2e}
\SetEndCharOfAlgoLine{}

\SetCommentSty{mycommfont}
\SetKwRepeat{Do}{do}{while}
\SetKw{KwBy}{by}

\newcommand{\linebreakand}{%
  \end{@IEEEauthorhalign}
  \hfill\mbox{}\par
  \mbox{}\hfill\begin{@IEEEauthorhalign}
}

\begin{document}
%
\title{Fixed-point Quantization of Convolutional Neural Networks for Quantized Inference on Embedded Platforms}

\author{\IEEEauthorblockN{Rishabh Goyal\IEEEauthorrefmark{1}, 
Dr. Joaquin Vanschoren\IEEEauthorrefmark{2}}
\IEEEauthorblockA{\textit{Mathematics and Computer Science} \\
\textit{Eindhoven University of Technology}\\
\IEEEauthorrefmark{1}rishabhgoyal56@gmail.com, 
\IEEEauthorrefmark{2}j.vanschoren@tue.nl}
\and
\IEEEauthorblockN{Victor Van Acht\IEEEauthorrefmark{3},
Stephan Nijssen\IEEEauthorrefmark{4}}
\IEEEauthorblockA{\textit{Philips Innovation Services} \\
Eindhoven, Netherlands \\
\IEEEauthorrefmark{3}victor.van.acht@philips.com, \IEEEauthorrefmark{4}stephan.nijssen@philips.com}
}

\maketitle
\thispagestyle{plain}
\pagestyle{plain}

\begin{abstract}
Convolutional Neural Networks (CNNs) have proven to be a powerful state-of-the-art method for image classification tasks. One drawback however is the high computational complexity and high memory consumption of CNNs which makes them unfeasible for execution on embedded platforms which are constrained on physical resources needed to support CNNs. Quantization has often been used to efficiently optimize CNNs for memory and computational complexity at the cost of a loss of prediction accuracy. We therefore propose a method to optimally quantize the weights, biases and activations of each layer of a pre-trained CNN while controlling the loss in inference accuracy to enable quantized inference. We quantize the 32-bit floating-point precision parameters to low bitwidth fixed-point representations thereby finding optimal bitwidths and fractional offsets for parameters of each layer of a given CNN. We quantize parameters of a CNN post-training without re-training it.
Our method is designed to quantize parameters of a CNN taking into account how other parameters are quantized because ignoring quantization errors due to other quantized parameters leads to a low precision CNN with accuracy losses of up to 50\% which is far beyond what is acceptable. Our final method therefore gives a low precision CNN with accuracy losses of less than 1\%. As compared to a method used by commercial tools that quantize all parameters to 8-bits, our approach provides quantized CNN with averages of 53\% lower memory consumption and 77.5\% lower cost of executing multiplications for the two CNNs trained on the four datasets that we tested our work on. We find that layer-wise quantization of parameters significantly helps in this process.
\end{abstract}


\IEEEpeerreviewmaketitle

\section{Introduction}
Recent developments in Deep Learning have drawn significant attention from research and industry, especially considering the ability of neural networks to efficiently deal with tasks such as image recognition and classification, object detection, speech recognition, word prediction etc. with high accuracy, even surpassing human capabilities. However, one shortcoming of these networks, as noted by many of the cited works in this paper, is the cost with respect to computational complexity and memory. While hardware such as GPUs and CPUs efficiently support these requirements for neural networks, these costs can be a problem when implementing these networks in practice for an embedded application. 

As compared to general purpose CPUs and GPUs, embedded systems are designed for specific applications and are restricted with respect to physical resources. They are quite small in size, have low on-chip memory, fewer arithmetic and logic units (ALUs) and therefore also consume less energy. They also lack the same level of parallelism found in CPUs and GPUs. Given the lack of ALUs, they often only support simple operations with low precision numbers. Many small embedded systems do not have ALUs for operations on commonly supported floating-point numbers and therefore require some software subroutines to manipulate floating-point numbers which is a more time consuming process. In such a case, a direct implementation of computationally expensive and high memory consuming algorithms such as neural networks is unfeasible without further optimizations. A solution to this problem is therefore to either design custom hardware to accelerate neural networks or to optimize neural networks for existing embedded hardware such as Field Programmable Gate Arrays (FPGAs) and micro-controllers or a combination of the two approaches.

We choose the latter approach by reducing numerical precision from the commonly used 32-bit floating-point precision to integer/fixed-point precision. This process of reducing the precision of numbers, done by mapping numbers from a larger set to a smaller and more discrete set is known as quantization. Since integer operations are simpler than floating-point operations, the process of quantization simplifies computational complexity and increases computational speed.

Quantizing to integer precision then removes the need for floating-point precision hardware which will in turn reduce energy consumption. We can also reduce the number of bits used to represent these numbers as much as possible, which would reduce the memory requirements of the model.

In signal processing, the process of quantization however leads to quantization errors. These errors are the differences between the original and the quantized values. In the context of neural networks, we can expect the quantization errors because of the quantized parameters to have a direct impact on the prediction accuracy of the network. We therefore must reduce the numerical precision of the network as much as possible, while minimizing the impact on the prediction accuracy. 

In this paper we propose a method that can optimally quantize the parameters, namely weights, biases and activations of each convolutional and dense layer in a pre-trained Convolutional Neural Network (CNN) in order to make it compatible for, and enable efficient quantized inference on embedded platforms. We quantize the parameters by reducing their precision from floating-point to the fixed-point number representation. Fixed-point numbers are treated like integers in hardware therefore allowing simple and fast operations, but include a scaling offset that allows for limited fractional precision. 

Our proposed method takes a CNN model, analyzes it, and returns the optimal fixed-point number representations for parameters of each layer in the network while ensuring an acceptable loss in inference accuracy. The resulting fixed-point representations can then be used to implement a low precision CNN in a lower level language like C, for deployment on an embedded platform. As opposed to compressing a given neural network for a specific embedded platform, our method is aimed at quantizing parameters of a given CNN for deployment on any embedded platform of choice. Our work is limited to the design of this method and is therefore not concerned with the implementation of the resulting quantized model on embedded platforms.

While multiple works in the literature train and re-train the CNN during the quantization process, our method relies on post-training quantization where we quantize the parameters of a pre-trained CNN without re-training the network. Additionally, even though all parameters of the CNN can be uniformly quantized to the same level of precision, we vary this for parameters of each layer as we find variance in the level of precision that parameters of certain layers require. As our results will show, this leads to our method giving quantized models with much lower memory consumption while ensuring an acceptable accuracy loss of the low precision network.

Although most CPUs and GPUs only support bitwidths of 4, 8, 16, 32, 64, we also consider arbitrary bitwidths that are not limited to these. We investigate this in order to understand if there is an incentive for developing dedicated hardware for these non-conventional bitwidths in the future. We also limit our scope to CNNs specifically for image classification problems.

We first present and discuss commonly referenced works in the literature in Section \ref{sec:relatedwork} that have approached neural network quantization with state-of-the-art results. In Section \ref{sec:probdef} we then present some preliminaries and precisely formulate the problem and research question we aim to address. Section \ref{sec:init_analysis} presents some initial analysis of experimentation that is used to understand how the quantization of parameters of the network affects the inference accuracy. Using the observations and conclusions from this analysis, we design an algorithm to efficiently find the optimal precision levels for each parameter of each layer in the network in Section \ref{sec:algo}. Section \ref{sec:experiments} presents the results of the algorithm that was tested on two different CNN architectures trained on four different datasets and we compare the low-precision model of our algorithm to low-precision CNNs generated by a simple baseline method commonly used by commercial tools. Finally, we present our conclusions in Section \ref{sec:conclusion}.

\section{Related Work}
\label{sec:relatedwork}

There is a significant amount of work in the literature that has explored reducing computational complexity and memory requirements of neural networks using quantization. \cite{guo2018survey} presents a clear overview of the topic of quantized neural networks, with details on the popular methods and techniques. Works have presented novel techniques to quantize parameters of neural networks. The proposed techniques tend to fall under two general approaches, namely Quantized training, or Post-training quantization.

Quantized training has been a popular approach that is focused on training neural networks using low-precision number representations for weights, activations, and in some cases gradients. Popular works have successfully used binary numbers \cite{courbariaux2015binaryconnect, courbariaux2016binarized, rastegari2016xnor} and integer arithmetic \cite{wu2018training, jacob2018quantization} reaching inference accuracy close to that of the original network. Works that have used this approach have therefore been able to replace many expensive operations with simple integer or bit-wise operations therefore reducing training time and complexity. One of the common arguments against this approach is the problem of gradient mismatch due to a difference between the quantized and full-precision activation function which causes a problem with gradient updates during gradient descent \cite{lin2016overcoming}. The aforementioned papers and many others using this approach also often need to rely on full-precision parameters for gradient and parameter updates to ensure convergence of gradient descent.

In post-training quantization, parameters of a network is quantized to a low-precision network after training the network with full precision parameters. Parameters of the network are then only quantized for efficient inference as compared to efficient training. There is however a limit to how much the arithmetic precision of the network can be lowered without further degradation in the inference accuracy of the network. Many works choose to then re-train the network post-quantization with the original training data, either once or multiple times during quantization \cite{shin2017fixed, anwar2015fixed, han2015deep, cheng2018differentiable}. This helps recover losses incurred due to quantization, allowing for harsher quantization of the parameters of the network. The resulting quantized network allows for significant model size reduction with prediction accuracy being on par with their floating-point precision counterparts. In \cite{han2015deep}, the authors of DeepCompression take a pre-trained model and pass it through their three stage pipeline that consists of pruning of the weights, quantization of the weights and finally Huffman encoding them for 35-49x reduction in model size. During this process, they re-train the network multiple times to ensure almost no accuracy is lost. 

However the aforementioned approaches are not ideal for practice, and especially not for a quick and efficient deployment of a low-precision CNN model. The reasons for this, as discussed by \cite{banner2018posttraining} is that these approaches require access to the original training data which for privacy reasons may not be possible. Secondly, the processes are also time consuming given the additional time needed for training and re-training. Finally, the networks might require additional optimizations for specific platforms or applications.

For quick and efficient deployment of a quantized model for efficient quantized inference, we may instead quantize CNNs post-training with acceptable accuracy losses \cite{lin2016fixed, banner2018posttraining, zhao2019improving, choukroun2019low}. Of the many highly cited works that utilize this approach, the most common technique uses k-means clustering for weights of a network that was, to the best of our knowledge, originally presented by \cite{gong2014compressing}. The authors use a code-book to store $k$ centroids, each of which replaces a group of weights. Since we now only need to store the code-book of $k$ centroids in memory, we can significantly reduce memory consumption of the CNN model. Therefore weights that pertain to the cluster with centroid $k$ can easily be referenced using the code-book. Many papers have used this approach and some have worked to improve upon it \cite{choi2016towards, stock2019and}. A disadvantage of using k-means as highlighted by \cite{choi2016towards} is that the k-means approach does not allow for control on the performance loss due to quantization.

Our work also takes inspiration from the conclusions found in \cite{krishnamoorthi2018quantizing}. They find that using a per-layer granularity for quantization is more beneficial than quantizing parameters of all layers to one representation, given that layers behave differently when subjected to quantization. This would allow for parameters of some layers to be quantized more than others ensuring that no layer forms a bottleneck. Many of the works cited above also use a per-layer approach to quantizing neural networks with low bitwidths and high inference accuracy. 

In using layer-wise quantization, works such as \cite{zhou2016dorefa} have empirically found the first and last layers of a CNN to be sensitive to quantization. As a result, they choose not to quantize these layers. \cite{choi2018pact} however finds that using conservative quantization for the first and last layer can still perform sufficiently well with minimal accuracy degradation. Our work takes note of this and also finds the first layer to be quite sensitive to quantization, therefore taking a more conservative approach for this layer. Our approach to dealing with quantization for each layer in the network is covered in more detail in Section \ref{sec:orderquant}.

\cite{krishnamoorthi2018quantizing} finds that simple post-training quantization can provide sufficiently quantized models with almost no accuracy losses (as low as 8-bits). Since we consider arbitrary bitwidths, and perform layer-wise quantization, the quantized models produced by Dependent Optimized Search have bitwidths as low as 2-bits for some parameters ranging and up to 5 or 7 bits for others. For quick deployment of a quantized model, post-training quantization without re-training is therefore a useful approach to quantize parameters of a given pre-trained model.

\begin{figure*}[t!]
    \centering
     \begin{subfigure}[t]{0.4\textwidth}
         \centering
         \includegraphics[width=0.6\textwidth]{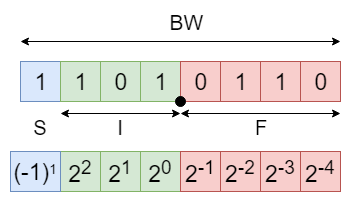}
         \caption{Binary representation of a fixed-point number consisting of a sign bit (S), integer part (I) and fractional part (F) making up the bitwidth (BW). Place value of each bit is provided below the example.}
         \label{fig:fxp_repr}
     \end{subfigure}
    \hspace{20mm}
     \begin{subfigure}[t]{0.4\textwidth}
         \centering
         \includegraphics[width=\textwidth]{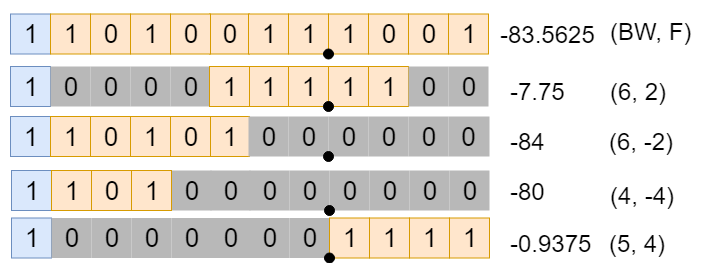}
         \caption{Examples of fixed-point representations for a given number. Bits stored in memory are in yellow and blue. Their floating-point equivalent value and corresponding BW and F are given to the right.}
         \label{fig:best_repr}
     \end{subfigure}
     \caption{Visualizations of a fixed-point number and their characteristics}
     \label{fig:fxp_vis}
\end{figure*}

\section{Problem Definition}
\label{sec:probdef}

In this section, we first present some preliminaries about the fixed-point number representation and the quantization function used to quantize any given floating-point number to fixed-point. Following this, we precisely formulate our intended goal for this paper.

\subsection{Fixed-point Number Representation}
\label{sec:fxp_nums}
Reducing the numerical precision from floating-point to integer precision results in a loss of the representation of fractional numbers. However, this results in an increase in computational speed because of the simplicity of integer operations over floating-point operations. Take for example the multiplication of two integers 3 and 4 as compared to the multiplication of the number $\pi$ up to a few digits, by itself. The latter would require more precision and the operations would be more complex given the fractional part of the number. \cite{nic:speedops} presents some experimental results showing that floating-point operations are slower than integer operations.

A middle ground between these two formats is the fixed-point precision. Fixed-point arithmetic is treated as integer precision in hardware but with an added fractional offset that fixes the number of binary digits after the radix. This allows for a limited fractional precision of the number. Since the fractional offset is a fixed value, integer operations can be applied to fixed-point numbers as the fractional offset can be accounted for after the arithmetic integer operations are executed.

The fixed-point system is advantageous over the floating-point system given that the latter system has different exponents for each number all of which need to be accounted for in each calculation, making calculations more complex and therefore more expensive. With a common exponent for multiple numbers, the fixed-point number system allows for simpler operations as the exponent is then accounted for after the arithmetic operations have occurred. \cite{courbariaux2014training} includes an insightful graphic on the difference between how floating-point and fixed-point numbers differ.

Fig.~\ref{fig:fxp_vis} illustrates the characteristics of a fixed-point number and some examples of it. We can characterize a fixed-point number using:
\begin{enumerate}
    \item \textbf{Bitwidth (BW):} The number of bits needed to store the number. The usual bitwidth for numbers in CPUs are 32 bits. In our model of the fixed-point number, the lowest possible bitwidth is 2 bits that consists of a sign bit and either an integer or fractional bit. In our paradigm, a 1 bit fixed-point number only includes the sign bit which we do not allow to be possible. Hence, no numbers can be represented using a 1 bit fixed-point number. 
    \item \textbf{Fractional Offset (F):} The integer position of the radix for the number. A positive value for F results in higher fractional precision (more precision), while a negative value of F results in storing more integer bits (more range). Fig.~\ref{fig:best_repr} illustrates the effect of changing F.
\end{enumerate}
The total bitwidth (BW) is therefore the sum of the number of bits allocated for the Sign (S), Integer part (I) and Fractional part (F). The Sign is always represented by 1 bit. The integer part of the number covers the range of numbers that can be represented, while the fractional part defines the precision. We then define a fixed-point representation as $(BW, F)$. 

We also discuss two elements of a binary number that will be useful later, namely the most significant bit (MSB) and least significant bit (LSB). The MSB is the bit with the largest numerical value, found to be the left most bit (excluding the sign). In Fig.~\ref{fig:fxp_repr} the value of the MSB is the bit with the place value of $2^2$. The LSB is the bit with the least numerical value, found to be the right most bit in the bitwidth. The LSB therefore defines the step size and precision, which is the difference between each consecutive binary number (1 bit). In Fig.~\ref{fig:fxp_repr}, the value of the LSB is the bit with place value of $2^{-4}$.

The fixed-point number representation is a data type that is only considered in binary, given that we fix the place of the radix based on the number of bits. However, we may equivalently represent them in floating-point format by summing up their respective place values (base 2) according to the binary number system. As an example, the number in Fig.~\ref{fig:fxp_repr} has an equivalent floating-point value of -5.375. Throughout this paper, we simulate fixed-point binary numbers by considering their floating-point equivalent values. By scaling numbers using base 2, we can effectively execute the same operations on floating-point that would be applied to the binary numbers with no differences in the final result.

Although the most common method of representing binary numbers in hardware is with the use of two's complement, we choose to represent our signed fixed-point binary numbers by using an explicit sign bit. This has the undesirable consequence of a possible representation of positive and negative 0. However, since our work is in simulation using a high level language like Python, this is not a problem because Python treats positive and negative zero as zero.

\subsection{Quantization function}
\label{sec:quant_func}
We quantize floating-point numbers to fixed-point by mapping floating-point numbers to their fixed-point equivalent representations in floating-point. A given floating-point number is quantized by first choosing the bitwidth and fractional offset $(BW, F)$ of the target fixed-point representation that the number must be quantized to. We generalize this by defining a function $Q(\mathbf{x})$ that quantizes a group of floating-point numbers $\mathbf{x}$ to their fixed-point equivalent values specified by the same target representation $(BW, F)$ while also taking into account our model of the fixed-point number.
\begin{equation}
    Q(\mathbf{x})_{(BW, F)} = \frac{C(R(\mathbf{x}\cdot 2^F), -t, t)}{2^F}
    \label{eq:quantfunc}
\end{equation}
where $t$ is defined as
\[ t = \begin{cases} 
        2^{BW - S} - 1 & BW > 1 \\
        0 & BW \leq 1
   \end{cases}
\]
and $R(x)$ is a function that rounds $x$ to the nearest integer, S is the number of bits for the sign with $S = 1$ and $C(x, a, b)$ is the clipping function defined as
\begin{equation}
    C(x, a, b) = 
    \begin{cases}
        a & \text{if $x \leq a$} \\
        x & \text{if $a < x < b$} \\
        b & \text{if $x \geq b$}
    \end{cases}
    \label{eq:clip}
\end{equation}

The piece-wise function of $t$ is a guard that ensures that the minimum bitwidth possible is 2 bits, which is a design choice.

By applying \eqref{eq:quantfunc} with choices for $(BW, F)$ we can observe the effect on the resulting quantized values in Fig.~\ref{fig:best_repr}. As an example, quantizing a number -83.5625 to a fixed-point representation $(BW = 6, F = 2)$ using \eqref{eq:quantfunc} gives a value of -7.75, which if converted to the fixed-point binary format would give the number seen in Example 2 of Fig.~\ref{fig:best_repr}. Alternatively, we may also apply the same operations to the binary numbers observed in Fig.~\ref{fig:best_repr}, and calculate their floating-point equivalent values as was done for Fig.~\ref{fig:fxp_vis} earlier. The results will be identical.

In order to simplify the experimental procedure, we use the Keras framework to work with CNN models. Since fixed-point or integer formats are not supported by Python or Keras, $Q(\mathbf{x})$ allows us to simulate fixed-point quantization on floating-point numbers that are equivalent to their binary fixed-point representations. The operations in $Q(\mathbf{x})$ simulate fixed-point quantization due to the scaling of $2^F$. The resulting quantized values are essentially the floating-point equivalent values of the fixed-point numbers. 

The two main operations in $Q(\mathbf{x})$ are the rounding and clipping functions $R(x)$ and $C(x, a, b)$ respectively. 
\subsubsection{Rounding}
By rounding the result of $\mathbf{x}\cdot 2^F$ to the nearest integer, we define the LSB $1~\mathit{LSB} = \frac{1}{2^F}$ and therefore the numerical precision. Since the LSB defines the step size, an increment of 1 bit in binary is the equivalent of an increase in $\frac{1}{2^F}$ in decimal. The value of $F$ determines the precision of the binary fixed-point representation as changing $F$ affects the value of the LSB and therefore the magnitude of the step size of 1 bit. To retain high precision, we therefore need a high value of F as that results in an extremely small step size between consecutive numbers. Conversely, to reduce precision we may decrease the value of F therefore increasing the step size between consecutive numbers, making the steps more discrete. 

Fig.~\ref{fig:best_repr} visually illustrates the effect of changing F as moving the set of bits that can be stored in memory (denoted in yellow and blue). For example, the representation of $(6, 2)$ has the smallest increment defined by $LSB = \frac{1}{2^{2}} = 0.25$, implying that a 1 bit increment is an increment of 0.25 in decimal. The representation $(6, -2)$ however (lower value of F) has the smallest increment defined by $LSB = 4$, similarly implying that a 1 bit increment is an increment of 4 in decimal. The latter is lower in precision as the former representation covers more intermediate values.

Decreasing F below a certain threshold $F_{min}$ would result in too low of a precision to represent the original number as the value of the LSB increases more than the value of the original number itself. This would automatically result in large differences between the quantized value and the original value and therefore result in high quantization error due to rounding.

\subsubsection{Clipping} 
Clipping occurs when a number is too large to be stored, given the range that the fixed-point number system allows for. In such a case, values outside this range are restricted to the maximum range of the fixed-point arithmetic system. With reference to our binary fixed-point number, this threshold is determined by the capacity of our bitwidth, namely the maximum absolute value that can be stored within the bitwidth. For instance, with a bitwidth of 3, the maximum absolute value that can be stored is a value of $2^{BW - S} - 1 = 2^{3 - 1} - 1 = 3$. All values larger than 3 or lower than -3 will therefore be restricted to those values respectively.

For a fixed value of F, with $F > F_{min}$, reducing $BW$ effectively reduces the range of values that can be stored because bits are removed from the left of the bitwidth. In order to avoid clipping, the bitwidth must account for the largest possible value in the given distribution of numbers, which occurs when we choose $BW \geq BW_{min}$ where $|2^{BW_{min} - 1} - 1| = max(|R(\mathbf{x \cdot 2^F})|)$ with reference to \eqref{eq:quantfunc}, \eqref{eq:clip}. As the bitwidth is reduced such that $BW < BW_{min}$, we find that $|2^{BW - 1} - 1| < max(|R(\mathbf{x \cdot 2^F})|)$ and values $|R(\mathbf{x \cdot 2^F})| > |2^{BW - 1} - 1|$ are restricted to $|2^{BW - 1} - 1|$. We also effectively reduce the most significant bit (MSB) by reducing the bitwidth where $1~MSB = \frac{2^{BW - 1}}{2^F}$. 

Conversely for a fixed $BW$, clipping can also occur by increasing the value of F in $R(\mathbf{x \cdot 2^F})$ such that $|R(\mathbf{x \cdot 2^F})| > |2^{BW - 1} - 1|$. Mathematically, values are scaled to be larger than what can be stored in the bitwidth therefore resulting in clipping. Visually in Fig.~\ref{fig:best_repr}, increasing F simply results in shifting the bitwidth window further to the right therefore changing the value of the MSB and decreasing it. This effect can be observed when increasing F from -2 to 2 in Fig.~\ref{fig:best_repr} as a value of -84 reduces to -7.75. We can note here that for no clipping to occur, for a fixed bitwidth $BW = 6$, $F = F_{max} = -2$ in order to ensure that no clipping occurs. The MSB plays an important role in the quantization error induced, since the MSB has the largest numerical value in the binary number.

\subsubsection{Range of values represented by $(BW, F)$}
A given fixed-point representation $(BW, F)$ can represent a range of values $(-m, m)$ with a step size of $LSB = \frac{1}{2^F}$, where $m$ is given as
\begin{equation}
\label{eq:max_val}
    m = \frac{2^{BW - 1} - 1}{2^F}
\end{equation}

Therefore, we can directly relate our choice of fixed-point representation with the distribution of the parameters we are trying to quantize. For example, using a fixed-point representation $(5, 7)$ will allow us to represent a range of values of $(-0.1171875, 0.1171875)$ with a step size of $1~LSB = \frac{1}{2^7}$. This will prove to be useful when trying to understand the range of values that a given fixed-point representation can cover.

\subsection{Problem Formalization}
\label{sec:problem_formal}
We can formalize our intended goal by using the quantization function given in \eqref{eq:quantfunc}. 

We characterize a floating-point precision pre-trained CNN model with $L$ layers by the parameters $p : \{ \mathbf{W}, \mathbf{B}, \mathbf{A} \}$ representing weights, biases and activations of the model respectively, where for each convolutional or dense layer $l$ in the network with a kernel size of $w_l \times h_l$ with $c_l$ channels and with output activations being of size $x \times y$, we have
\begin{equation}
\label{eq:params}
    \begin{split}
        \mathbf{W} = & ~ \{ \mathbf{W}_l~|~l = 1,\cdots,L, \mathbf{W}_l \in \mathbb{R}^{w_l \times h_l \times c_l} \}\\
        \mathbf{B} = & ~ \{ \mathbf{B}_l~|~l = 1,\cdots,L, \mathbf{B}_l \in \mathbb{R}^{1 \times c_l} \} \\
        \mathbf{A} = & ~ \{ \mathbf{A}_l~|~l = 1,\cdots,L, \mathbf{A}_l \in \mathbb{R}^{x \times y \times c_l} \}
    \end{split}
\end{equation}

The inference accuracy of the network is defined as $a(\mathbf{W}, \mathbf{B}, \mathbf{A})$. Using \eqref{eq:quantfunc} we may then quantize the parameters of the network to their respective fixed-point representations $\{\{(BW, F)_l\}_W, \{(BW, F)_l\}_B, \{(BW, F)_l\}_A ~ | ~ l = 1, \cdots, L\}$ giving a set of quantized parameter values defined as $p_Q : \{Q(\mathbf{W}), Q(\mathbf{B}), Q(\mathbf{A})\}$ where 
\begin{equation}
\label{eq:quantparams}
    \begin{split}
        Q(\mathbf{W}) = &\{Q(\mathbf{W}_l)_{(BW, F)_{l, W}}~|~l = 1,\cdots,L, \mathbf{W}_l \in \mathbb{R}^{w_l \times h_l \times c_l} \}\\
        Q(\mathbf{B}) = & \{Q(\mathbf{B}_l)_{(BW, F)_{l, B}}~|~l = 1,\cdots,L, \mathbf{B}_l \in \mathbb{R}^{1 \times c_l} \}\\
        Q(\mathbf{A}) = & \{Q(\mathbf{A}_l)_{(BW, F)_{l, A}}~|~l = 1,\cdots,L, \mathbf{A}_l \in \mathbb{R}^{x \times y \times c_l} \}\\
    \end{split}
\end{equation}
The inference accuracy of the quantized network would then be $a(Q(\mathbf{W}), Q(\mathbf{B}), Q(\mathbf{A}))$.

After quantizing all the parameters of the network using fixed-point representations representations $(BW, F)_{l, p}$ for the respective parameters of each layer, we can evaluate the loss in inference accuracy using
\begin{equation}
    \label{eq:acc_loss}
    \Delta a = \frac{a(\mathbf{W}, \mathbf{B}, \mathbf{A}) - a(Q(\mathbf{W}), Q(\mathbf{B}), Q(\mathbf{A}))}{a(\mathbf{W}, \mathbf{B}, \mathbf{A})}
\end{equation}

We aim to therefore find optimal fixed-point representations $\{\{(BW, F)^*_l\}_W, \{(BW, F)^*_l\}_B, \{(BW, F)^*_l\}_A ~ | ~ l = 1, \cdots, L\}$ subject to the following three constraints concerning inference accuracy loss, the memory consumed by the network and cost of multiplications respectively:
\begin{equation}
\label{eq:constraints}
    \begin{split}
         & \Delta a \leq \epsilon \\
         \min & \sum_{p = \mathbf{W}, \mathbf{B}, \mathbf{A}} \sum_{l=1}^L (BW_l)_p \cdot n(\mathbf{p}_l) \\
         \min & \sum_{l = 1}^L ((BW_l)_{p = W} \cdot n(\mathbf{W}_l)) \cdot ((BW_l)_{p = A} \cdot n(\mathbf{A}_l))\\
    \end{split}
\end{equation}
where $\epsilon$ is the acceptable loss in inference accuracy after quantizing all weights, biases and activations of the network, a value defined by the user and $n(\mathbf{x})$ returns the number of values in $\mathbf{x}$. 

For the constraint on memory consumption, we simply sum across parameters of all layers, the number of bits for parameters of each layer. For the cost of multiplications in bits, we represent this as the sum across all layers, of the multiplication between the number of bits for the weights of the layer and the number of bits for activations of the layer. Our goal is to minimize these two constraints.

The aforementioned constraints simply translate to trying to find the lowest possible bitwidths for all parameters in our network, such that the accuracy loss is within acceptable bounds.

Section \ref{sec:budget} will discuss how values of the inference accuracy loss $\epsilon$ are chosen.

Our primary objective can therefore be formulated with the following research question:

\textit{How can we find an optimal set of fixed-point representations $(BW, F)^*_{l, p}$ for parameters $p \in \{\mathbf{W}, \mathbf{B}, \mathbf{A}\}$ of each layer $l = 1,\cdots, L$ in a pre-trained Convolutional Neural Network such that we have the smallest possible bitwidths with an acceptable loss $\epsilon$ in the network's inference accuracy?}

Along with our primary research question, we aim to answer whether quantization of a certain parameter $p$ in layer $l$ is dependent on the quantization of some other parameter in the CNN. We may formulate it as follows:

\textit{Are the optimal fixed-point representations $(BW, F)^*_{l, p}$ for parameter $p$ of layer $l$ either dependent on fixed-point representations $(BW, F)^*_{k, p}$ for $k \neq l$ or dependent on fixed-point representations $(BW, F)^*_{l, q}$ for $q \neq p$ for all layers  $l$ and $k$?}

\section{Initial Analysis}
\label{sec:init_analysis}

In order to design a method for finding optimal fixed-point representations of all parameters in the network, we first performed some experiments to investigate the relationship between the fixed-point representation $(BW, F)$ and the accuracy loss of the low-precision CNN. 

We start by presenting our pipeline for the post-training quantization process. We then discuss two ways of evaluating the accuracy loss of the network after quantization. Using the pipeline, we then perform brute-force analysis on the fixed-point representation $(BW, F)$ to quantize a set of parameters of the network to observe the effect on the inference accuracy of the network $\Delta a$. These experiments will be the basis for the design of our method in Section \ref{sec:algodesign}.

\begin{figure}[t!]
    \centering
    \includegraphics[width=0.45\textwidth]{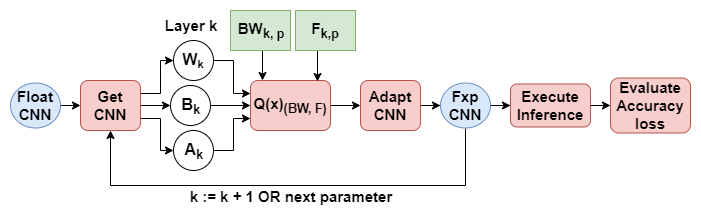}
    \caption{A visualization of our post-training quantization and inference pipeline used to quantize parameters of each layer $k$ of a CNN. Each layer $k$ has parameters ($\mathbf{W}_k, \mathbf{B}_k, \mathbf{A}_k$). Here $BW_{k, p}$ and $F_{k, p}$ are input parameters to our pipeline and the inference accuracy loss of the quantized network $\Delta a$ is the output parameter}
    \label{fig:ptq_process}
\end{figure}

\begin{figure}[t!]
    \centering
    \includegraphics[width=0.35\textwidth]{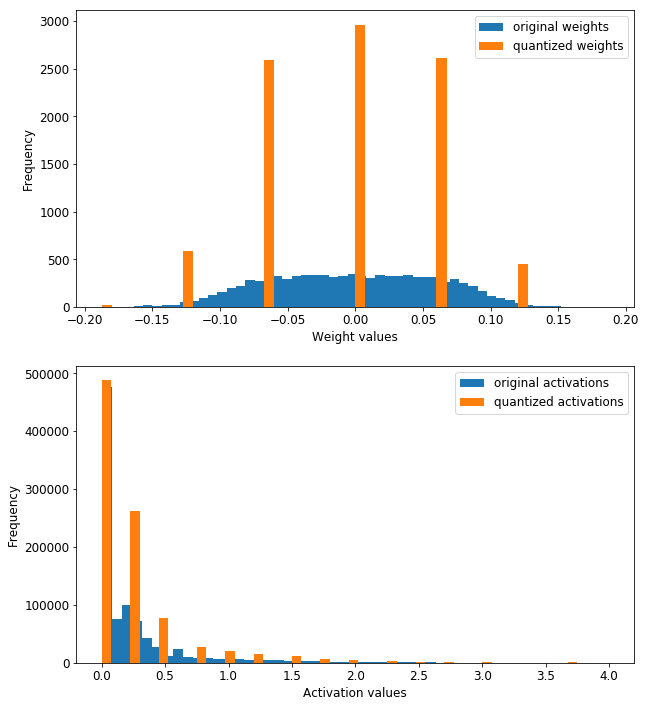}
    \caption{Histograms of the quantized and full precision weights and activations. Weights and activations are quantized to fixed-point representations $(BW, F)$ of (3, 4) and (5, 2) respectively. 1 LSB for weights is $1/2^4 = 0.0625$ and for activations is $1/2^2 = 0.25$}
    \label{fig:quant_dist}
\end{figure}

\begin{algorithm*}[htbp]
\KwData{$\mathbf{p}_l$ : parameter values for layer $l$, $BW_{l, p}$: Bitwidth, $F_{l, p}$: Fractional offset, $\mathit{IND}$: Boolean flag,  $r_Q$: Set of optimal $(BW, F)^*$ collected}
\KwResult{Inference accuracy loss for quantization of parameter $p$ of layer $l$:  $\Delta a_{l, p}$}
\uIf{IND}{ \tcp{Independent Quantization}
    Adapt the CNN model, replacing $\mathbf{p}_l$ with $Q(\mathbf{p}_l)_{(BW, F)_{l, p}}$ \tcp{Post-training quantization process}
    
    $a(Q(\mathbf{p}_l)_{(BW, F)_{l, p}}, \mathbf{q}) \gets $ Inference accuracy of the quantized model\\
    $\Delta a^I_{l, p} \gets (a(\mathbf{W}, \mathbf{B}, \mathbf{A}) - a(Q(\mathbf{p}_l)_{(BW, F)_{l, p}}, \mathbf{q})) ~/~ a(\mathbf{W}, \mathbf{B}, \mathbf{A}) $ \tcp{Accuracy loss from quantized values $Q(\mathbf{p}_l)$}
    
    $\Delta a_{l, p} \gets \Delta a^I_{l, p}$ \tcp{Independent quantization inference accuracy loss}
}
\Else{ \tcp{Dependent Quantization}
    \For{$(BW, F)^*_{k, q}$ in $r_Q$ }
    {\tcp{For $q \neq p$, $1 \leq k \leq L$}
    Adapt the CNN model, replacing $\mathbf{q}_k$ with $Q(\mathbf{q}_k)_{(BW, F)_{k, q}}$ \tcp{Post-training quantization process}
    
    }
    Adapt the CNN model again by replacing $\mathbf{p}_l$ by $Q(\mathbf{p}_l)_{(BW, F)_{l, p}}$ \tcp{Post-training quantization process}
    
    $a(Q(\mathbf{p}_l)_{(BW, F)_{l, p}}, Q(\mathbf{q}_k)) \gets$ Inference accuracy of quantized CNN model\\
    $\Delta a^D_{l, p} \gets (a(\mathbf{W}, \mathbf{B}, \mathbf{A}) - a(Q(\mathbf{p}_l)_{(BW, F)_{l, p}}, Q(\mathbf{q}_k))) ~/~ a(\mathbf{W}, \mathbf{B}, \mathbf{A}) $
    \tcp{Accuracy loss from quantized values $Q(\mathbf{p}_l), Q(\mathbf{q}_k)$}
    
    $\Delta a_{l, p} \gets \Delta a^D_{l, p}$ \tcp{Dependent quantization inference accuracy loss}
}
\caption{\texttt{EvalAccLossCNN}}
\label{alg:eval_acc_loss}
\end{algorithm*}

\subsection{Post-training quantization process}
\label{sec:PT_pipeline}
Fig.~\ref{fig:ptq_process} illustrates our post-training quantization pipeline for quantizing parameters of a pre-trained CNN. We perform the following steps when quantizing a parameter of a layer $\mathbf{p}_l$ of a given CNN to a fixed-point representation $(BW, F)_{k, p}$:

\begin{enumerate}
    \item Taking a pre-trained CNN, we first extract the parameter values $\mathbf{p}_k \in \{ \mathbf{W}_k, \mathbf{B}_k, \mathbf{A}_k\}$ of layer $k$. 
    \item We then use \eqref{eq:quantfunc} to quantize $\mathbf{p}_k$ to a fixed-point representation $(BW, F)_{k, p}$ therefore giving $Q(\mathbf{p}_k)_{(BW, F)_{k, p}}$.
    \item The network is then adapted with the respective quantized values. For weights and biases, we simply place the respective quantized parameters back in the model. For activations, we alter the architecture of the network by inserting a custom quantization layer (Lambda layer in Keras) after layer $k$. For convolutional layers, this layer is appended after the activation function (ReLU), while for the last dense layer, the layer is appended before the activation function (Softmax).
    \item We may then optionally feed the same quantized CNN back into the pipeline and repeat steps 2 and 3 to quantize other parameters in the CNN.
    \item We run inference on our quantized model with the original test data and measure the loss in inference accuracy lost $\Delta a$. We discuss this further in the following sub-section.
\end{enumerate}

Fig.~\ref{fig:quant_dist} shows the result of quantizing the weights and activations of a layer in a CNN to fixed-point representations of $(3, 4)$ and $(5, 2)$ respectively. Note the sparsity that is induced in the respective distribution due to the step size that was defined by their LSB values.

\subsection{Evaluation of the Inference Accuracy Loss}
\label{sec:eval_acc_loss}

From the post-training quantization pipeline designed in Section \ref{sec:PT_pipeline}, we note that the inference accuracy lost $\Delta a$ can be evaluated in two ways that we discuss below.

Algorithm \ref{alg:eval_acc_loss} illustrates the process for the two types of evaluation.
The algorithm takes as input parameter values $\mathbf{p} \in \{ \mathbf{W}, \mathbf{B}, \mathbf{A}\}$ for layer $l$, a fixed-point representation for quantization $(BW, F)_{l, p}$, a boolean flag that we will discuss and a set of fixed-point representations for the quantization of other parameters in the network that are not $p$. The two types of quantization follow directly from the post-training quantization process described in Section \ref{sec:PT_pipeline}.

\subsubsection{Independent Quantization}
By setting the boolean flag in \texttt{EvalAccLossCNN} $\mathit{IND} = \mathit{True}$ we define independent quantization as quantizing a parameter of a layer $\mathbf{p}_l$ to a fixed-point representation $(BW, F)_{l, p}$ while keeping all other parameters in the network at floating-point precision. We define it as independent quantization as we quantize $\mathbf{p}_l$ independent of how other parameters in the network are quantized. The inference accuracy loss $\Delta a$ is then measured for the quantization of $\mathbf{p}_l$ which we express as $\Delta a^I_{l, p}$. We will use the superscript $I$ to refer to Independent Quantization.

In this way, effects of quantization on other parameters are effectively ignored as they are left at full precision in the evaluation of the inference accuracy loss. Therefore the quantization errors because of the other quantized parameters are not taken into account during the evaluation of $\Delta a^I_{l, p}$.

\begin{figure*}[t!]
    \centering
     \begin{subfigure}[t]{0.48\textwidth}
         \centering
         \includegraphics[width=\textwidth]{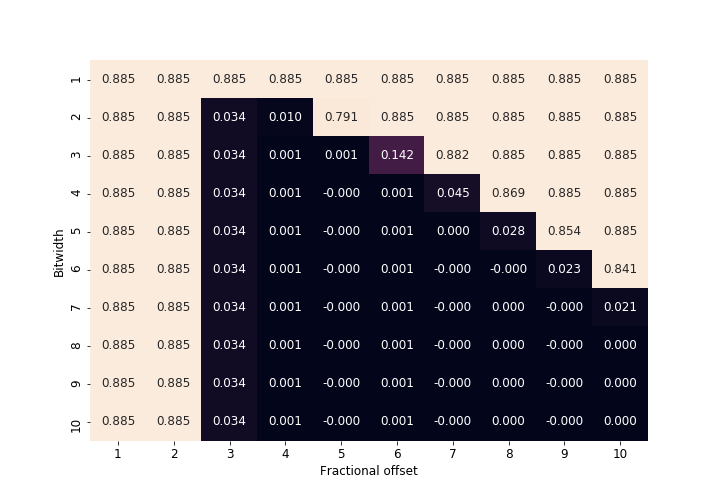}
         \caption{Brute-force of $(BW, F)$ for quantization of weights of a layer. All other biases and activations at full precision}
         \label{fig:bf_w}
     \end{subfigure}
     \hfill
     \begin{subfigure}[t]{0.48\textwidth}
         \centering
         \includegraphics[width=\textwidth]{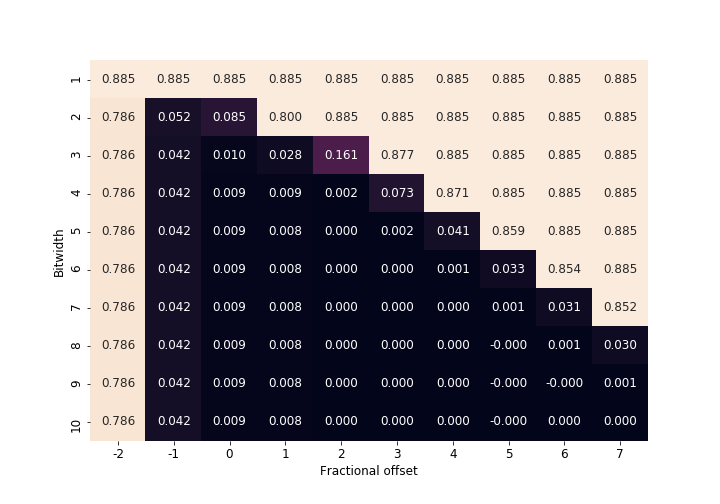}
         \caption{Brute-force of $(BW, F)$ for quantization of activations of a layer. All other weights and biases at full precision}
         \label{fig:bf_a}
     \end{subfigure}
     \caption{Heatmap of the inference accuracy loss $\Delta a_{l,p}$ when using grid search on $BW$ and $F$ to quantize weights or activations of a layer of a simple 5-layer sequential CNN trained on the MNIST dataset. All other parameters were kept at full-precision.}
     \label{fig:bf_results}
\end{figure*}

\subsubsection{Dependent Quantization}
By setting the boolean flag in \texttt{EvalAccLossCNN} $\mathit{IND} = \mathit{False}$, to quantize parameters $\mathbf{p}_l$ to fixed-point representation $(BW, F)_{l, p}$, we perform dependent quantization by also quantizing other parameters in the network $q \neq p$ to their fixed-point representations $(BW, F)_{l, q}$ (Step 4 of the Post-training quantization process). Lines 6-13 in Algorithm \ref{alg:eval_acc_loss} show how dependent quantization is performed.

We first take all parameters $q \neq p$ for layers $k$ ($1 \leq k \leq L$) and replace them by the quantized values $Q(\mathbf{q}_k)_{(BW, F)^*_{k, q}}$ using their fixed-point representations $(BW, F)_{k, q}$ stored in $r_Q$. We then perform a similar task by adapting the same CNN and replacing its parameters $\mathbf{p}_l$ of layer $l$ by $Q(\mathbf{p}_l)_{(BW, F)_{l, p}}$ given the fixed-point representation $(BW, F)_{l, p}$. We can then measure the loss in inference accuracy $\Delta a = \Delta a^D_{l, p}$ for the quantization of parameters $\mathbf{p}_l$ and parameters $\mathbf{q}_k$. We will use the superscript $D$ to refer to dependent quantization.

Unlike independent quantization, dependent quantization now includes the effects of errors because of other quantized parameters in the evaluation of $\Delta a = \Delta a^D_{l, p}$. With this process, we sequentially quantize a parameters of a network while still keeping the other quantized parameter values. 

Given the difference in methods of evaluating $\Delta a$ where we either do or do not take the other quantized parameters into account, we can expect discrepancies between $\Delta a^I_{l, p}$ and $\Delta a^D_{l, p}$, concretely with $\Delta a^D_{l, p} \geq \Delta a^I_{l, p}$ given that the former takes errors due to quantization of other parameters into account in the accuracy loss. We will show and discuss these discrepancies in more detail through our results.

\subsection{Brute Force Analysis}
\label{sec:bf_analysis}

With the post-training quantization pipeline and a method of evaluating the inference accuracy loss $\Delta a$ we now aim to to observe the effect of changing the bitwidth and fractional offset $(BW, F)$ for a group of parameters in our network on the inference accuracy. We associate this group of numbers by the parameter values pertaining to a layer in the network $\mathbf{p}_l$ where $1 \leq l \leq L$ and $p \in \{ \mathbf{W}, \mathbf{B}, \mathbf{A} \}$ thereby quantizing parameters of a network layer-wise.

We now perform a brute-force analysis of the bitwidth and fractional offset $(BW, F)_{l, p}$ used to quantize all parameters $p$ and all layers $l$ while measuring the accuracy loss $\Delta a^I_{l, p}$ to understand the independent effect of quantizing parameter $\mathbf{p}_l$ without considering quantization effects of other quantized parameters. Hence, other parameters are left at floating-point precision. We can then repeat this process for each parameter in the network for all layers.

Fig.~\ref{fig:bf_results} shows a heatmap of the resulting experiment with the values of $\Delta a^I_{l, p}$ for a given range of combinations of $(BW, F)$ on the second convolutional layer (arbitrary choice) of a 5-layer sequential CNN (4 convolutional and 1 dense layer) trained on the MNIST dataset for weights in Fig.~\ref{fig:bf_w} and activations in Fig.~\ref{fig:bf_a}. Similar results for other layers collected can be found in Appendix A.

We identify two major regions, namely the darker region representing the area with little to no accuracy loss compared to the original network, and the brighter region where the network can no longer make useful predictions. Note the negative values in the center of Fig.~\ref{fig:bf_w} indicating a possible improvement in the inference accuracy of the network compared to the original full precision network. The magnitude of this improvement however is quite small. Therefore we reason that this improvement could possibly be within a margin of error of the inference accuracy measurement that might vary across data. Another possibility is that the fractional offset may result in a set of weights around zero to be rounded to zero therefore resulting in some regularization of the network. We do not present the result for biases here because the experimental results showed that, for the most part, quantization of biases did not have a significant impact on the inference accuracy of the network. We will show this through our results in Section \ref{sec:experiments}.

The advantage of the plots in Fig.~\ref{fig:bf_results} is the fact that effects of rounding and clipping can quickly be observed using a visual representation. However, these plots are expensive to generate considering how long it takes to brute force these grid of values. 

We discuss the effects of rounding and clipping in detail below using the intuition generated from Section \ref{sec:quant_func}.

\subsubsection{Rounding and the value of $F_{min}$}
In Section \ref{sec:quant_func} we noted that the fractional offset affects the step size characterized by the LSB as $1~LSB = \frac{1}{2^F}$. Decreasing the fractional offset resulted in sparser distributions of values as each increment of 1 bit results in a large discrete steps.

Fig.~\ref{fig:bf_results} shows that reducing $F$ too much results in a high loss of numeric precision of the set of parameters that are quantized. With $F$ being too low, the value of the LSB increases and intermediate values are then rounded to extremely sparse and discrete values. These large step sizes result in too many intermediate values being rounded up or down. This has the effect of increasing the quantization error because the values are changed too far from their original value therefore resulting in a high drop in inference accuracy of the network. Fig.~\ref{fig:bf_w} shows that $F_{min} = 4$ for little to no accuracy loss for weights. Fig.~\ref{fig:bf_a} on the other hand shows that $F_{min} = 0$ for activations with an accuracy loss of 0.9\%. 

We can also directly relate the scale of fractional offsets for weights and activations in Fig.~\ref{fig:bf_results} to the distribution of weights and activations in Fig.~\ref{fig:quant_dist}. In Fig.~\ref{fig:quant_dist} since $|\mathbf{W}_l| << 1$, we require positive fractional offsets in Fig.~\ref{fig:bf_w} as we require more fractional precision. On the other hand, since $\mathbf{A}_l \geq 0$ with many activations that are also greater than 1, we note in Fig.~\ref{fig:bf_a} that we may also potentially use negative or null fractional offsets therefore only storing the integer part of the number and losing fractional precision. In this latter case however, given that a majority of the activations in Fig.~\ref{fig:quant_dist} are less than 1, the effect of rounding to integer precision results in a loss of 4.2\% in the inference accuracy.

\subsubsection{Clipping and the value of $BW_{min}$ and $F_{max}$}
From the discussion on clipping in Section \ref{sec:quant_func} we noted that clipping occurs when either $BW < BW_{min}$ for a fixed F or when $F > F_{max}$ for a fixed $BW$. Fig.~\ref{fig:bf_results} shows this behaviour along the diagonal line, where either increasing $F$ for the same $BW$ or decreasing $BW$ for the same $F$ results in clipping of the maximum and/or minimum values of respective weights or activations. The plot also shows that clipping has a very harsh effect on the inference accuracy as the loss $\Delta a$ increases substantially. 

Fig.~\ref{fig:bf_results} shows that the values of $F_{max}$ and $BW_{min}$ can vary depending on the choice of $BW$ and $F$ respectively. This can be explained by considering the fact that for a bitwidth and fractional offset $BW \geq BW_{min}$ and $F \leq F_{max}$, reducing $(BW, F)$ equally would result in bits being dropped from the right side of the number therefore avoiding the numbers from being subjected to the clipping operation in \eqref{eq:quantfunc}. This can also be observed in Fig.~\ref{fig:best_repr} where reducing $BW$ and $F$ by 2 (examples 3, 4) did not result in clipping as we retained the MSB of the original number but that bits were dropped from the right. The numeric value did however increase from -84 to -80 due to the reduction in precision caused by the decrease in F. Mathematically, given that $BW$ and $F$ are raised to the same base of 2 as long as $BW \geq BW_{min}$ and $F \leq F_{max}$, changing BW and F equally will maintain $|2^{BW - 1} - 1| > max(|R(\mathbf{x} \cdot 2^{F})|)$ and values will therefore not be clipped. Therefore, we note that in Fig.~\ref{fig:bf_results}, as long as we reduce $BW$ and $F$ equally while $BW \geq BW_{min}$ and $F \leq F_{max}$, we ensure that no values of the weights or activations are clipped or restricted.

We noted in Section \ref{sec:quant_func} that using \eqref{eq:max_val} we can evaluate the range of values a given fixed-point representation can cover. This allows us to directly relate the bitwidth to the distribution of parameters that we want to quantize. For example, using a fixed-point representation $(BW, F) = (4, 7)$ allows us to represent values within the range of $(-m, m) = (-0.0546875, 0.0546875)$. Comparing this to the original weights in Fig.~\ref{fig:quant_dist} we observe that quantizing the original weights to this fixed-point representation would automatically result in clipping of the weights distribution to the value of $(-m, m)$ because $m << \max(|\mathbf{W}_l|)$.

By observing the parameter distributions, we also noticed that slight clipping has a more significant effect on the accuracy loss of the network than as compared to effects of rounding. This can be explained by the fact that clipping values to those in a lower bitwidth results in much larger quantization errors considering the differences between the original and quantized values. Take for example the equivalent numbers in Fig.~\ref{fig:best_repr} (examples 2 and 5) where clipping of the original number to a lower value or bitwidth results in large differences between the original value (-83.5625) and the quantized number. In the same figure, reducing the precision by reducing F results in a much smaller quantization error (examples 3, 4) because the differences between -83.5625 and -84 or -80 are much smaller. As noted earlier, a major reason for this is the value of the most significant bit $1~MSB = \frac{2^{BW - 1}}{2^F}$.

\subsubsection{Conservative vs Harsh Quantization}
Generally, in Fig.~\ref{fig:bf_results}, we note that the accuracy loss increases as the bitwidth and possibly also the fractional offset are reduced. Fig.~\ref{fig:bf_w} shows that for a fixed-point representation of $(BW, F) = (2, 4)$, the accuracy loss is 1\%. Increasing this bitwidth to possibly 4 bits ($(BW, F) = (4, 4)$) would result in only a 0.1\% accuracy loss. Similarly, for activations in Fig.~\ref{fig:bf_a}, using a bitwidth lower than 5 bits results in accuracy losses of at least 0.9\%. 

From all our tests of brute force analysis, we noticed that quantizing conservatively (using larger bitwidths) led to lower accuracy drops while quantizing in a harsh manner resulted in larger accuracy losses. The reason for this is that with lower bitwidths, we are restricting the range of parameter values that can be represented therefore throttling information. We therefore require a larger bitwidth to ensure lower accuracy losses.

\subsubsection{Optimal Fixed-point representation} \label{sec:opt_repr}
We noted through the constraints in \eqref{eq:constraints} on memory consumption and multiplication complexity that we require the smallest possible bitwidths to quantize parameters in our network such that the loss in inference accuracy is acceptable. From Fig.~\ref{fig:bf_results} we note that we achieve this constraint when both the bitwidth and fractional offset are as low as possible while ensuring that the accuracy loss $\Delta a \leq \epsilon$. For instance, in Fig.~\ref{fig:bf_w}, with $\Delta a = 0.001$ (loss of 0.1\%), we can reduce the bitwidth of the quantized weights to as low as 3 bits with a fixed-point representation of $(BW, F)^*_{l, W} = (3, 4)$.

\subsubsection{Weights vs Activations}
The plots in Fig.~\ref{fig:bf_results} also show a clear discrepancy between the quantization of weights and activations. By representing all weights for the the given layer $l$ with 3 bits, the inference accuracy of the network drops by 0.1\% ($\Delta a^I_{l, W} = 0.001$). However, quantizing activations to a fixed-point point representation with $BW = 3$ would result in at least a 1\% loss in inference accuracy. In order to retain the original inference accuracy, activations would be required to use a minimum of 5 bits $(F = 2)$. We found a similar discrepancy between the two parameter types for the other models trained on the other 3 datasets. From this we conclude that activations are generally more sensitive to quantization than weights. Empirical data for this may be found in Appendix A. This conclusion was also supported by works in the literature.

We discuss the reason for this discrepancy in Section \ref{sec:discussion}.

\subsubsection{Observations for various layers}
Due to readability we only presented the result of brute-force analysis on the weights and activations of one layer in the CNN. Additional results can be found in Appendix A. While the experiment yielded similar plots for other layers as seen in Fig.~\ref{fig:bf_results}, the major discrepancy was observed for the parameters of layer 1. We generally observed that parameters of layer 1 required higher bitwidths and was therefore more sensitive to the operations of clipping and rounding. The results of our experiments in Section \ref{sec:experiments} will show this more clearly. We also discuss the reasons for these layer-wise discrepancies in Section \ref{sec:discussion}.

\subsubsection{Evaluation of the accuracy loss}\label{sec:quant_other_params}
For the results in Fig.~\ref{fig:bf_results} we performed independent quantization by measuring inference accuracy loss $\Delta a^I_{l, p}$ as discussed in Section \ref{sec:bf_analysis}. By performing a similar brute-force experiment, but instead measuring $\Delta a^D_{l, p}$ while quantizing other parameters $q \neq p$ to their respective fixed-point representations $(BW, F)_{l, q}$ would result in larger accuracy losses for the network. This is because the error due to quantization of other parameters may accumulate through the network.

With other parameters quantized, the same fixed-point representation $(BW, F)^*_{l, W} = (3, 4)$ may now result in a larger inference accuracy loss $\Delta a$ of the network if we evaluate $\Delta a^D_{l, p}$. To compensate, we may therefore be required to increase the bitwidth and fractional offset to retain or lower the loss in inference accuracy. The choice of dependent or independent quantization will therefore have an effect on the optimal fixed-point representation $(BW, F)^*_{l, p}$, introducing a dependency on other parts of the network. 

Through our experimentation we noted that $\Delta a^D_{l, p} \geq \Delta a^I_{l, p}$. The magnitude of the difference between the two depends on how aggressively or conservatively the other parameters are quantized. However, the pattern and shape found in Fig.~\ref{fig:bf_results} was consistent for both types of quantization.

We discuss our approach to investigating this in more detail in Section \ref{sec:orderquant}.

\subsubsection{Summary of Conclusions}
We now summarize the conclusions drawn from the tests on brute-force analysis briefly:
\begin{itemize}
    \item For a minimum loss in inference accuracy, we need to choose a fixed-point representation $(BW, F)$ such that $F_{min} \leq F \leq F_{max}$ and $BW \geq BW_{min}$. Decreasing $F$ below $F_{min}$ would significantly increase the step size characterized by the LSB. This would reduce the precision too much resulting in high quantization errors due to rounding. On the other hand, increasing $F$ above $F_{max}$ or decreasing $BW$ below $BW_{min}$ will result in clipping of the parameter distribution, in turn resulting in a high accuracy loss.
    \item We note that clipping the parameter distribution results in much larger accuracy losses as compared to the operation of rounding. This was because clipping creates much larger errors between the original and the quantized values whereas rounding results in smaller quantization errors. The major reason for this is the role that the value of MSB plays since it has the largest numerical value. Changing the MSB by clipping will have larger effects on the quantization errors. Reducing the precision by reducing $F$ results in smaller quantization errors and therefore smaller accuracy losses. On the other hand, reducing the $BW$ has larger effects on the accuracy losses since we effectively reduce the range of values that can be covered.
    \item The fixed-point representation $(BW, F)$ can be directly related to the distribution of the parameter we intend to quantize. With the bitwidth $BW$ we can evaluate the range of values that can be covered. With the fractional offset $F$, we can evaluate the number of bits needed for the fractional part of the number. The value of F could also be negative, which is means that we only store integer bits instead.
    \item Quantizing conservatively requires using larger bitwidths in order to minimize the effects of clipping and rounding on the distribution of the parameter values. Quantizing harshly, thus using lower bitwidths, results in more significant effects of clipping and rounding on the distribution of the parameter values.
    \item The optimal fixed-point representation is one that has the smallest possible bitwidth and fractional offset, for which the accuracy loss of the quantized network is still acceptable.
    \item There are discrepancies in the minimum bitwidths required to represent weights, biases and activations. We discuss the reason for this in Section \ref{sec:discussion}.
    \item While the shape of the plot is consistent for parameters of all layers, we observed a slightly different behaviour in layer 1. Concretely, parameters of layer 1 usually require larger bitwidths than parameters of the successive layers. We discuss this in more detail in Section \ref{sec:discussion}.
    \item The method of evaluating the accuracy loss will affect our choice of the optimal fixed-point representation. Using dependent quantization (measuring $\Delta a^D_{l, p}$) instead of independent quantization (measuring $\Delta a^I_{l, p}$) will influence our decision of the optimal fixed-point representation possible for the accuracy loss we are willing to accept. We noted that generally, including errors due to quantization of other parameters by evaluating $\Delta a^D_{l, p}$ will result in $\Delta a^D_{l, p} \geq \Delta a^I_{l, p}$ depending on how the other parameters are quantized.
\end{itemize}

\section{Algorithm Design}
\label{sec:algo}
\subsection{Algorithm for Efficient Search}
\label{sec:algodesign}
\begin{algorithm*}[t!]
\KwData{Acceptable loss $\epsilon_{l, p}$, Initial bitwidth $BW^0_{l, p}$, Boolean flag IND}
\KwResult{All optimal fixed-point representations $r_Q$ : $\{ \{(BW, F)^*_l~|~ l = [1, \cdots, L]\}_p~|~p = \{ \mathbf{W}, \mathbf{B}, \mathbf{A}\} \}$}
$r_Q = \{\}$ \tcp{Set of the optimal fixed-point representations $(BW, F)^*_{l, p}$}
\For{$p$ \textnormal{:} $\{ \mathbf{W}, \mathbf{B}, \mathbf{A} \}$}
    {
    \tcp{For each parameter}
    \For{layer $l$}{
    \tcp{For each layer}
        $F^0_{l, p} \gets BW^0_{l, p} - S - \lceil log_2 (\max (|\mathbf{p}_l|))\rceil$\\
        $\Delta a_{l, p} := \mathit{EvalAccLossCNN}(\mathbf{p}_l, BW^0_{l, p}, F^0_{l, p}, \mathit{IND}, r_Q)$
        
        \uIf{$\Delta a_{l, p} > \epsilon_{l, p}$}{
            End Algorithm and ask for a larger $BW^0_{l, p}$ or higher $\epsilon_{l, p}$
        }

        \While{$\Delta a_{l, p} \leq \epsilon_{l, p}$\tcp{Reduce the BW and F}}{
        $BW_{l, p}, F_{l, p} := BW_{l, p} - 1, F_{l, p} - 1$\\
        $\Delta a_{l, p} := \mathit{EvalAccLossCNN}(\mathbf{p}_l, BW_{l, p}, F_{l, p}, \mathit{IND}, r_Q)$
        }
        $(\hat{BW}, \hat{F})_{l, p} \gets$ lowest $(BW, F)_{l, p}$ for which $\Delta a_{l, p} \leq \epsilon_{l, p}$\\
        \While{$\Delta a_{l, p} \leq \epsilon_{l, p}$ \tcp{Reduce BW further if possible}}{
        $\hat{BW}_{l, p} := \hat{BW}_{l, p} - 1$\\
        $\Delta a_{l, p} := \mathit{EvalAccLossCNN}(\mathbf{p}_l, \hat{BW}_{l, p}, \hat{F}_{l, p}, \mathit{IND}, r_Q)$
        }
        $\hat{BW}_{l, p} \gets$ lowest $\hat{BW}_{l, p}$ for which $\Delta a_{l, p} \leq \epsilon_{l, p}$
        
        $\Delta \hat{a}_{l, p} \gets \mathit{EvalAccLossCNN}(\mathbf{p}_l, \hat{BW}_{l, p}, \hat{F}_{l, p}, \mathit{IND}, r_Q)$ \tcp{$(\hat{BW}, \hat{F})_{l, p}$ is the result of diagonal search}
        
        \For{$(BW, F)_{l, p} := (\hat{BW} \pm 1, \hat{F} \pm 1)_{l, p}$ }{ \tcp{Local search. Search neighbouring representations}
            $\Delta a_{l, p} \gets \mathit{EvalAccLossCNN}(\mathbf{p}_l, BW_{l, p}, F_{l, p}, \mathit{IND}, r_Q)$
        }
        $(\Tilde{BW}, \Tilde{F})_{l, p} \gets (BW, F)_{l, p}$ with lowest $\Delta a_{l, p}$ 
        \tcp{best of the nine $(\hat{BW}, \hat{F})_{l, p}$ representations}

         $\Delta \Tilde{a}_{l, p} \gets \mathit{EvalAccLossCNN}(\mathbf{p}_l, \Tilde{BW}_{l, p}, \Tilde{F}_{l, p}, \mathit{IND}, r_Q)$ \tcp{$(\Tilde{BW}, \Tilde{F})_{l, p}$ is the result of local search}
         
        \uIf{$\Tilde{BW}_{l, p} \neq \hat{BW}_{l, p}$ \&\& $\Tilde{F}_{l, p} \neq \hat{F}_{l, p}$ \tcp{Compare local and diagonal search}} {
            \uIf{ $\Delta \hat{a}_{l, p} - \Delta \Tilde{a}_{l, p} > \delta$ }{
                $(BW, F)^*_{l, p} \gets$ optimal representation $(BW, F)$ with the lower accuracy loss $\Delta a_{l, p}$
            }
            \Else{
                $(BW, F)^*_{l, p} \gets$ optimal representation $(BW, F)$ with the lower bitwidth $BW_{l, p}$
            }
        }
        \Else {
            Optimal $(BW, F)^*_{l, p} \gets$ Representation $(\Tilde{BW}, \Tilde{F})_{l, p}$
        }
        $r_Q := r_Q \cup \{\,(BW, F)^*_{l, p}\,\}$
    }
}
\caption{\texttt{OptSearchCNN}}
\label{alg:Opts}
\end{algorithm*}

We now present the design of our method aimed at finding the optimal fixed-point representations $\{ \{(BW, F)_l~|~ l = [1, \cdots, L]\}_p~|~p = \{ \mathbf{W}, \mathbf{B}, \mathbf{A}\} \}$ for all layers $l$ and all network parameters $p$ of a CNN to meet the constraints in \eqref{eq:constraints}. These constraints simply translate to determining the smallest possible bitwidths subject to an accuracy loss $\Delta a \leq \epsilon$ for the entire network, as simply stated in our primary research question. In Section \ref{sec:opt_repr} through Fig.~\ref{fig:bf_results}, the smallest possible bitwidths occurred for a fixed-point representation $(BW, F)^*_{l, p}$ that was lowest in value while ensuring $\Delta a_{l, p} \leq \epsilon_{l, p}$ (either independent or dependent) for any given layer $l$ and any given parameter $p$ in the CNN. 

For the design of the algorithm, we consider the predictable pattern of the accuracy loss $\Delta a^I_{l, p}$ of the network observed in Fig.~\ref{fig:bf_results} and the corresponding conclusions based on this pattern on the bounds of the fixed-point representation $(BW, F)_{l, p}$. We noted that as long as $BW \geq BW_{min}$ and $F_{min} \leq F \leq F_{max}$ the accuracy loss $\Delta a^I_{l, p}$ after quantizing parameters of the CNN is minimal. We use these bounds and this behaviour to design our main method. 

Algorithm \ref{alg:Opts} presents our method \texttt{OptSearchCNN} for finding optimal fixed-point representations $(BW, F)_{l, p}$ for all parameters $p \in \{ \mathbf{W}, \mathbf{B}, \mathbf{A} \}$ for all layers $l$ in a given CNN. 

\texttt{OptSearchCNN} requires the following inputs:
\begin{itemize}
    \item \textit{Acceptable Inference Accuracy Loss} ($\epsilon_{l, p}$): The stopping condition for the algorithm to find the optimal fixed-point representation $(BW, F)^*_{l, p}$ for quantization of parameter $p$ of layer $l$.
    \item \textit{Initial Bitwidth} ($BW^0_{l, p}$): A starting bitwidth from which the algorithm will work towards the smallest bitwidth. This needs to be sufficiently high. From our experimentation, 8-12 bits has sufficed. This value may however vary.
    \item \textit{Boolean flag} ($\mathit{IND}$): A boolean variable used to determine whether the inference accuracy loss evaluated after quantizing the parameter of a layer $\mathbf{p}_l$ must be computed using Independent or Dependent quantization. It is set to True for independent quantization. Based on the evaluation of $\Delta a$ we either perform independent or dependent optimized search.
\end{itemize}

In Section \ref{sec:quant_other_params} we noted that the optimal fixed-point representation $(BW, F)^*_{l, p}$ found for parameter values $\mathbf{p}_l$ will depend on how the inference accuracy loss is evaluated, either independent of or dependent on other quantized parameters. Given that we compare the inference accuracy loss $\Delta a_{l, p}$ at every stage with the acceptable loss $\epsilon_{l, p}$ our method of evaluating this loss is important. The flag $\mathit{IND}$ as inputs to \texttt{OptSearchCNN} and \texttt{EvalAccLossCNN} decide this.

In order to store the optimal fixed-point representations collected during the execution of the algorithm, we use a set $r_Q$ to store this. We now briefly summarize the main steps of \texttt{OptSearchCNN} in Algorithm \ref{alg:Opts}:
\begin{enumerate}
    \item By the order of the \texttt{for} loops in \texttt{OptSearchCNN}, we repeat the following process over all layers, one parameter at a time. 
    \item Start with a sufficiently large initial bitwidth $BW^0_{l, p}$ (8-12 bits suffices) and calculate the corresponding fractional offset $F^0_{l, p}$ such that we avoid clipping the maximum and minimum parameter values $\mathbf{p}_l$. To avoid clipping, we simply assign as many of the bits in our bitwidth to the integer part of the number such that we include the largest absolute valued number in the distribution. Therefore the number of integer bits $I = \lceil log_2(\max (|\mathbf{p}_l|)) \rceil$. $F^0_{l, p}$ is then given as
    \begin{equation}
    \label{eq:noclip}
        F^0_{l, p} = BW^0_{l, p} - S - I
    \end{equation}
    where S is the number of bits for the sign, $S = 1$. We then evaluate the accuracy loss for this representation $(BW, F)$. \textbf{(Lines 4-5)}
    \item Traverse diagonally with respect to $(BW, F)_{l, p}$ upwards by reducing the two values by 1 equally in Fig.~\ref{fig:bf_results} until the $\Delta a_{l, p} > \epsilon_{l, p}$. The resulting representation $(\hat{BW}, \hat{F})_{l, p}$ is the lowest $(BW, F)$ collected for which the accuracy loss is acceptable. \textbf{(Lines 8-13)}
    \item Attempt to reduce the bitwidth further while keeping the fractional offset F constant. The resulting fixed-point representation $(\hat{BW}, \hat{F})_{l, p}$ optimal result from our diagonal search with an accuracy loss $\Delta \hat{a}_{l, p} \leq \epsilon_{l, p}$. \textbf{(Lines 14-20)}
    \item With $(\hat{BW}, \hat{F})_{l, p}$, we now also look at neighbouring fixed-point representations with respect to Fig.~\ref{fig:bf_results} with a step of 1 in each direction for the bitwidth and fractional offset. The result of this local search is a fixed-point representation $(\Tilde{BW}, \Tilde{F})_{l, p}$ with an accuracy loss of $\Delta \Tilde{a}_{l, p} \leq \epsilon_{l, p}$  and is a minimum of the 9 values collected. We search for the minimum to attempt to reduce accuracy loss, if possible. \textbf{(Lines 21 - 25)}
    \item Compare the result of diagonal search ($(\hat{BW}, \hat{F})_{l, p}, \Delta \hat{a}_{l, p}$) and local search ($(\Tilde{BW}, \Tilde{F})_{l, p}, \Delta \Tilde{a}_{l, p}$) and if the accuracy losses differ by more than a hyper parameter $\delta$, we strategically make a choice as to the representation that we consider optimal. \textbf{(Lines 26 - 35)}
    \item Store the optimal fixed-point representation $(BW, F)^*_{l, p}$ recorded in $r_Q$.
\end{enumerate}

A successful termination of this algorithm will return all optimal fixed-point representations for the parameters $p$ for all layers $l$ in the CNN, represented by $r_Q$ : $\{ \{(BW, F)^*_l~|~ l = [1, \cdots, L]\}_p~|~p = \{ \mathbf{W}, \mathbf{B}, \mathbf{A}\} \}$. With these fixed-point representations, we may then quantize the respective parameters giving quantized parameters \eqref{eq:quantparams} for which the inference accuracy is $a(Q(\mathbf{W}), Q(\mathbf{B}), Q(\mathbf{A}))$ and the accuracy loss as seen in \eqref{eq:constraints} is $\Delta a \leq \epsilon$. The fixed-point representations in $r_Q$ should accordingly satisfy the constraints of \eqref{eq:constraints}.

The algorithm may also terminate unsuccessfully at the following lines of \texttt{OptSearchCNN}:
\begin{itemize}
    \item \textbf{Line 5-8:} The accuracy loss evaluated there $\Delta a_{l, p}$ is higher than the acceptable loss $\epsilon_{l, p}$.
    \item \textbf{Line 11:} The while loop terminates early because the accuracy loss for the fixed-point representation evaluated does not give any representations for which the accuracy loss is acceptable.
\end{itemize}

We may then have to either use a larger initial bitwidth $BW^0_{l, p}$, potentially increasing it to 16 bits, or increase the acceptable accuracy loss $\epsilon_{l, p}$. This results in a trade-off that needs to be made between inference accuracy loss and the lower memory consumption that can be achieved with the resulting bitwidth. 

With respect to time complexity, execution of \texttt{OptSearchCNN} is bounded as we use finite loops. The execution time of each instance of the loop depends on the starting bitwidth and will in the worst case terminate with a minimum possible bitwidth of 1 bit, representing no bits remaining. Additionally, this execution time also depends on the time required for running inference and the time required to adapt the CNN model. Given that we tested our algorithm on simple models with four simple datasets as discussed in Section \ref{sec:exp_setup}, the inference time of the entire test set was fairly short. For the datasets MNIST, Fashion-MNIST and CIFAR10, the time for inference on the two types of architectures was roughly 3s and 7s respectively. Inference on the SVHN dataset took the longest amount of time (7s and 15s for the two types of architectures) since it had many more test examples. The algorithm was executed on a CPU, while inference was executed on a GPU. We present the architectures used for experimentation in the following section.

\subsection{Investigating Dependence of Quantization}
\label{sec:orderquant}
In Section \ref{sec:problem_formal}, we also expressed the need to investigate whether the optimal fixed-point representations $(BW, F)^*_{l, p}$ for parameters of layers $\mathbf{p}_l$ was dependent on the optimal fixed-point representation $(BW, F)^*_{l, q}$ of another parameter in the network $q \neq p$. By defining our method of evaluating the accuracy loss $\Delta a_{l, p}$ independently or dependently in \texttt{EvalAccLossCNN} (Algorithm \ref{alg:eval_acc_loss}), we defined a method to partially address this question. We will now define our method of addressing these questions more concretely.

\texttt{OptSearchCNN} finds optimal fixed-point representations with minimal bitwidths subject to having an accuracy loss $\Delta a_{l, p} \leq \epsilon_{l, p}$. The evaluation of $\Delta a$, as mentioned earlier in the paper will affect the bitwidths that the algorithm is able to find.

\subsubsection{Independent Optimized Search}

We noted in Section \ref{sec:eval_acc_loss} that independent quantization ignored effects of quantization on other parameters by the evaluation of $\Delta a^I_{l, p}$. We can perform independent optimized search by evaluating $\Delta a^I_{l, p}$ and setting $\mathit{IND} = \mathit{True}$. Independent optimized search will then search for fixed-point representations with minimum bitwidths by evaluating if $\Delta a = \Delta a^I_{l, p} \leq \epsilon_{l, p}$. Therefore, the choice of optimal fixed-point representations for parameters of each layer will be independent of that of other parameters. In such a case the order in which parameters of the network are quantized should not matter. We will refer to this version of optimized search as independent optimized search.

\subsubsection{Dependent Optimized Search}

On the other hand, dependent quantization does take effects of quantization of other parameters $q \neq p$ into account in the evaluation of $\Delta a^D_{l, p}$ in \texttt{EvalAccLossCNN}. We can perform dependent optimized search by evaluating $\Delta a^D_{l, p}$ setting $\mathit{IND} = \mathit{False}$. The optimal $(BW, F)^*_{l, p}$ to quantize parameter $\mathbf{p}_l$ will then depend on how the other parameters have been quantized up to that point. In such a case, we aim to investigate whether parameters must be quantized in a specific order for the resulting fixed-point representations to either provide lower memory consumption or potentially a lower loss in inference accuracy. We consider the following two cases:
\begin{enumerate}
    \item \textbf{Layers:} Quantizing parameter $\mathbf{p}_l$ either forwards (layers $l = 1,\cdots,L$) or reverse (layers $l = L,\cdots,1$). We may also look at bi-directional or other (pseudo-)random orders across layers, however we limit our scope to the two orders only.
    \item \textbf{Parameters:} Order in which we choose to quantize weights, biases and activations. We note 6 different ways to order the quantization of these three parameter types. 
\end{enumerate}

Given the dependency on how other parameters are quantized in the CNN, we perform dependent optimized search in a more controlled manner. Concretely, we aim to find optimal fixed-point representations with minimal bitwidths such that the accuracy loss $\Delta a^D_{l, p}$ is controlled by our choice of $\epsilon^D_{l, p}$ at every stage of the execution of \texttt{OptSearchCNN}. This is done by setting an allocation scheme for the acceptable loss in inference accuracy $\epsilon^D_{l, p}$. Controlling  $\epsilon^D_{l, p}$ as the network is quantized will allow dependent optimized search to quantize the respective parameters conservatively or harshly depending on how the network has been quantized up until that point. A larger value of $\epsilon^D_{l, p}$ will result in harsher quantization with lower bitwidths while lower values will result in conservative quantization and larger bitwidths. We discuss our allocation schemes in Section \ref{sec:budget}.

By controlling the acceptable accuracy loss as parameters in the network are quantized, we can therefore expect a variance in the bitwidths that dependent optimized search is able to find. We will discuss these layer-wise discrepancies in more detail in Section \ref{sec:dep_opts}.

\section{Experiments and Evaluation}
\label{sec:experiments}

We now test the implementation of \texttt{OptSearchCNN} with independent or dependent \texttt{EvalAccLossCNN}.

We first present the CNNs and datasets used for testing \texttt{OptSearchCNN}. We then present the resulting accuracy loss, bitwidths and memory consumption of the quantized models resulting from the optimizations of fixed-point representations done by Independent and Dependent \texttt{OptSearchCNN}. Finally, we choose the better of the two types of \texttt{OptSearchCNN} and compare it against a baseline approach that is commonly used by commercial tools.

For readability and to keep the section concise, we only present a small subset of our results. However the conclusions and observations are based on experiments from tests on all models and datasets mentioned above. Additional empirical data are provided in Appendix B-D. For our results, we only present the values of the accuracy losses and the corresponding bitwidths that were found by \texttt{OptSearchCNN} for quantizing a given CNN. We do not present the values of the fractional offsets from the optimal fixed-point representations $(BW, F)_{l, p}$ given by \texttt{OptSearchCNN} as they are not important for our constraints in \eqref{eq:constraints}. 

\begin{figure*}[t!]
    \centering
     \begin{subfigure}[t]{0.46\textwidth}
         \centering
         \includegraphics[width=\textwidth]{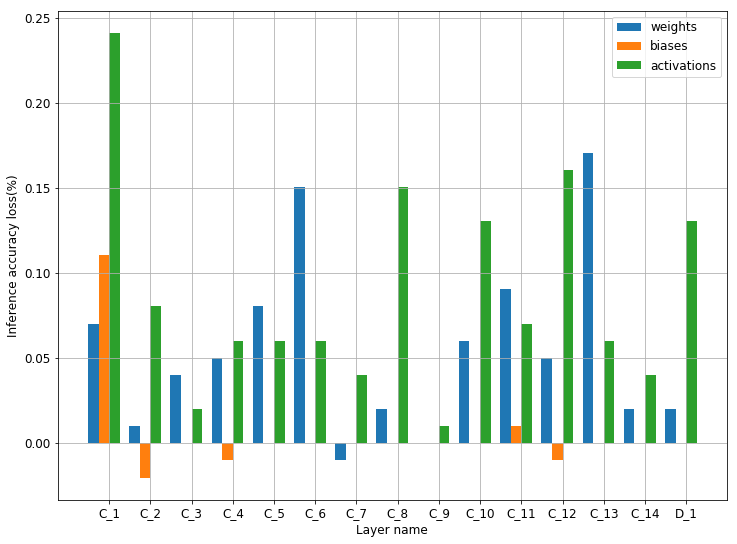}
         \caption{Inference accuracy loss $\Delta a^I_{l, p}$ for parameters of each layer after finding its optimal $(BW, F)_{l, p}$ using Independent Optimized Search while keeping other parameters at full precision.}
         \label{fig:ind_acc_loss_layer}
     \end{subfigure}
    \hspace{10mm}
     \begin{subfigure}[t]{0.46\textwidth}
         \centering
         \includegraphics[width=\textwidth]{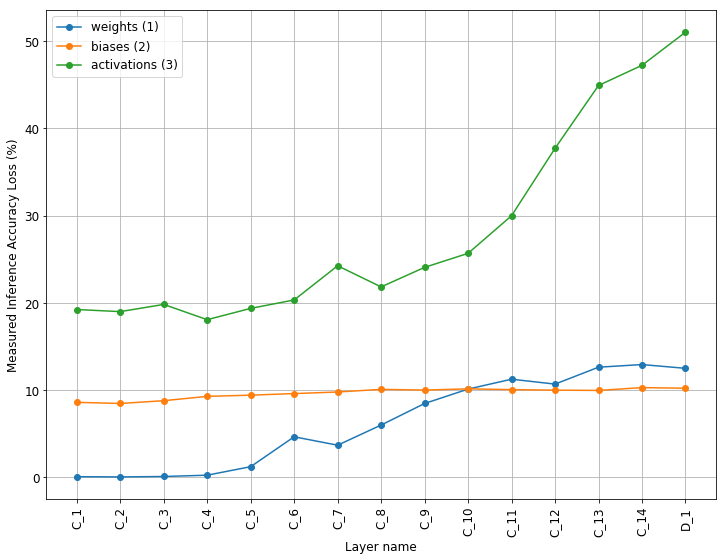}
         \caption{Inference accuracy loss of the CNN $\Delta a$ every time a parameter of a layer is quantized to the respective $(BW, F)^*_{l, p}$ found using Independent Optimized Search. Parameters of the network are quantized sequentially in the order of weights, biases followed by activations from layers 1 to L.}
         \label{fig:ind_seq_acc_loss}
     \end{subfigure}
     \caption{Inference accuracy loss measured using two ways after quantizing parameters to their fixed-point representations of the pre-trained 15-layer Sequential CNN trained on MNIST using independent Optimized Search. Acceptable loss $\epsilon_{l, p} = 0.3\%$.}
     \label{fig:ind_quant_results}
\end{figure*}

\begin{figure}
    \centering
    \includegraphics[width=0.45\textwidth]{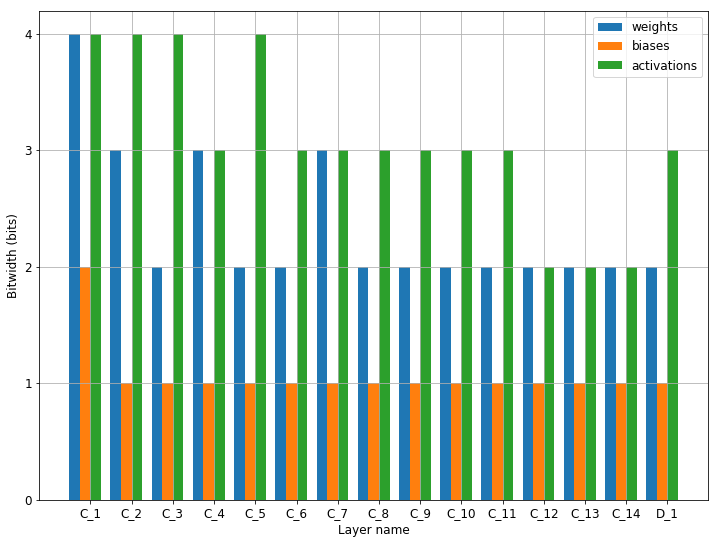}
    \caption{Optimal bitwidths $BW^*_{l, p}$ of parameters of each layer found using Independent Optimized Search for a 15 layer Sequential CNN trained on MNIST}
    \label{fig:ind_bw_seq}
\end{figure}
\subsection{Experimental Setup}
\label{sec:exp_setup}
For our experiments, we trained two CNN architectures on four different datasets (MNIST handwritten digits, CIFAR10, Fashion-MNIST and Street View House Numbers (SVHN)). \texttt{OptSearchCNN} then extracted the parameters of these models and optimized the fixed-point representations to obtain a quantized network. We use the following two architectures, which are representative of some commonly used CNNs for image classification and therefore suffice for the purposes of our tests:
\begin{enumerate}
    \item \textbf{Sequential Model:} For a simple architecture, we consider a single-branch sequential model with Convolutional layers, Max or Average Pooling and Batch Normalization. For some cases, we also included Dropout layers during training to ensure that the models did not over-fit the data. It consists of 14 convolutional layers with 1 dense layer for classification. The architecture has roughly 327k parameters.
    \item \textbf{Branched Model:} For a branched architecture we use the structure of the Inception model by Google \cite{szegedy2016rethinking}. The architecture has roughly 194k parameters and consists of 22 convolutional layers with 1 dense layer for classification.
\end{enumerate}

A visual illustration of the architectures are provided in Appendix E.

We also used Batch Normalization layers in all our models. We found that using Batch Normalization layers aided in keeping the distribution of weights of the CNN normalized to have equal magnitude and thus being less spread out, covering a smaller range of values. This allows us to use smaller bitwidths to cover the range of weight values with sufficient resolution.

The activation function used for all convolutional layers was the ReLU function, while classification in the dense layer at the end is done using Softmax. 

With respect to the quantization pipeline in Fig.~\ref{fig:ptq_process}, we quantize weights and/or biases and plug them back into the model. For activations, we add an additional layer that includes the quantization function $Q(\mathbf{x})_{(BW, F)_{l, p}}$. In Keras this is achieved by using a Lambda layer that passes all activations through $Q(\mathbf{x})_{(BW, F)_{l, p}}$. This layer is added after activations for convolutional layers, and before activations for the Dense layer. In addition, we limit the quantization process only to convolutional and dense layers and leave the other layers untouched.

\subsection{Independent Optimized Search}
\label{sec:ind_opts}
For independent \texttt{OptSearchCNN}, we instantiated Algorithm \ref{alg:Opts} with an acceptable accuracy loss for each parameter in each layer $\mathbf{p}_l$ to be $\epsilon_{l, p} = \epsilon^I_{l, p} = 0.003$ (0.3\%). Therefore \texttt{OptSearchCNN} will find fixed-point representations for all layers $l$ and all parameters $p$ such that the loss when quantizing that parameter is $\Delta a^I_{l, p} \leq \epsilon^I_{l, p}$. We specify an initial bitwidth $BW^0_{l, p} = 8$ and set the boolean flag $\mathit{IND} = \mathit{True}$. 

After successful termination of independent \texttt{OptSearchCNN}, we get optimal fixed-point representations $r_Q = r^I_Q$, where each $(BW, F)^*_{l, p}$ is determined independently of the optimal $(BW, F)^*_{l, q}$ for other parameters $q \neq p$, i.e. with other parameters at full precision. 

Fig.~\ref{fig:ind_acc_loss_layer} shows the accuracy losses $\Delta a^I_{l, p}$ measured when quantizing $\mathbf{p}_l$ to the fixed-point representation $(BW, F)^*_{l, p}$ stored in $r^I_Q$ using independent quantization for a 15-layer sequential CNN pre-trained on MNIST. Independent \texttt{OptSearchCNN} successfully terminates with $\Delta a^I_{l, p} \leq \epsilon_{l, p} = 0.003$ (0.3\%) for all layers $l$ and parameters $p$ and with bitwidths as low as low as 2 bits for weights, 1 bit for biases (pruned away), and 3 bits for activations as seen in Fig.~\ref{fig:ind_bw_seq}.

Given these optimal fixed-point representations, we can now evaluate whether the accuracy loss of the network $\Delta a$ is comparable to the individual accuracy losses $\Delta a^I_{l, p}$. 

We can execute the post-training quantization pipeline in Fig.~\ref{fig:ptq_process} to sequentially quantize parameters of the network to their fixed-point representations collected in $r^I_Q$ returned by \texttt{OptSearchCNN} and evaluate $\Delta a$ every time parameter values are quantized. Weights, biases, and activations from layers 1 to L are quantized in the respective order. In such a case, unlike Independent quantization, the preceding parameters still remain quantized as we feed the fixed-point network back into our pipeline (Fig.~\ref{fig:ptq_process}). 

Fig.~\ref{fig:ind_seq_acc_loss} shows that as the respective parameters are sequentially quantized from layers 1 to L, the $\Delta a$ of the quantized network grows quite significantly, rising higher than $\epsilon_{l, p} = 0.003$(0.3\%) that was set for each parameter $\mathbf{p}_l$ of layer $l$. After quantizing all parameters, we note a final inference accuracy loss of approximately 50\%. Although too much of the accuracy of the network is lost by only quantizing the weights (roughly 12\%), the accuracy loss of the network increases even more dramatically by the time most of the activations are quantized.

We also observe that the accuracy loss grows more rapidly when sequentially quantizing the activations as compared to weights. We reversed the order in which these parameters were quantized, namely trying activations, biases and weights instead and found that this rapid accuracy loss increase was now observed for weights instead of activations. From this we concluded that the rate at which this accuracy loss increases is not due to any one of the parameters that are being quantized. Instead, this is because once most of the layers of the third parameter type are quantized, too much of the information is lost. Therefore, quantizing the parameters of the network further results in a more significant increase in accuracy loss.


By ignoring the error due to other quantized parameters in the network during the evaluation of independent $\Delta a^I_{l, p}$ (\texttt{EvalAccLossCNN}) to find the optimal fixed-point representations $r^I_Q$, independent \texttt{OptSearchCNN} is able to quantize the respective parameters $\mathbf{p}_l$ quite harshly with extremely low bitwidths. Ignoring the effects of quantization on other parameters in the network however clearly does not work given that in Fig.~\ref{fig:ind_seq_acc_loss} we note a much larger accuracy loss of the entire quantized network. The error induced due to the quantized parameters propagates through the network and not taking that into account in the evaluation of $\Delta a_{l, p}$ when trying to find the optimal $(BW, F)^*_{l, p}$ does not give promising results. The exact relationship of this error is still unknown and requires further experimentation and study. Works in literature have investigated this by looking at the problem from the perspective of noise that is introduced when each of the parameters are changed and have found ways to minimize this noise while achieving a quantized network, one instance being \cite{lin2016fixed}. We note that the relationship between the individual accuracy losses $\Delta a_{l, p}$ and the accuracy loss $\Delta a$ of the quantized network is not simply linear or additive in nature.

From all our experiments, activations were more sensitive to quantization than weights or biases as they required equal if not larger bitwidths as observed in Fig.~\ref{fig:ind_bw_seq}. We discuss the reason for this in Section \ref{sec:discussion}.

As observed in Fig.~\ref{fig:ind_acc_loss_layer}, quantized activations for the most part had larger effects on the accuracy losses of the network than did quantized weights. We hypothesize that this may have to do with the more direct effect that activations have as compared to weights, on the feature maps created by the following layer. However, this discrepancy would require further study to concretely point out why quantization of activations have larger accuracy losses than do the quantization of weights.

By observing the accuracy losses in Fig.~\ref{fig:ind_acc_loss_layer}, \ref{fig:ind_seq_acc_loss} and the corresponding bitwidths in Fig.~\ref{fig:ind_bw_seq} with additional results in Appendix B, we note that biases were often pruned resulting in insignificant accuracy losses. From this we conclude that the biases had the least effect on the accuracy of the network. 

We further discuss the discrepancy in quantization behaviour between the three parameter types in Section \ref{sec:discussion}.
\subsection{Dependent Optimized Search}
\label{sec:dep_opts}
As discussed in Section \ref{sec:orderquant}, we perform dependent \texttt{OptSearchCNN} with the assumption that the optimal $(BW, F)^*_{l, p}$ for parameter $\mathbf{p}_l$ is dependent on how other parameters of the network have already been quantized to their optimal fixed-point representations up until that point. We now discuss schemes to control the accuracy loss of the CNN to perform dependent \texttt{OptSearchCNN} and based on the results choose the best scheme for the final method.

\begin{figure}
    \centering
    \includegraphics[width=0.46\textwidth]{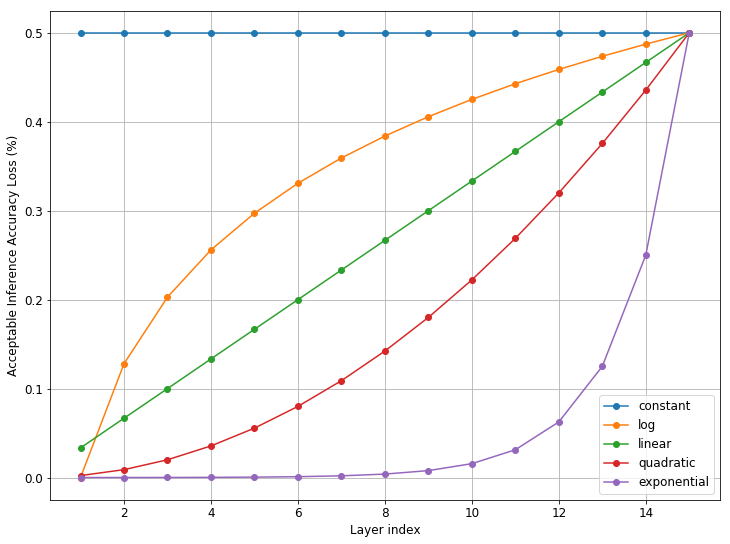}
    \caption{Allocation schemes for acceptable accuracy loss for a parameter of each layer $\epsilon^D_{l, p}$ sequentially increased to a maximum until the parameter of the last layer, for dependent optimized search. Max acceptable accuracy loss is 0.5\%}
    \label{fig:ideal_budget_alloc}
\end{figure}

\subsubsection{Acceptable accuracy loss allocation}
\label{sec:budget}
We discussed controlling the allocation of $\epsilon^D_{l, p}$ to ensure acceptable losses in inference accuracy during and after the execution of dependent \texttt{OptSearchCNN}. We may think of this as the allocation of a budget for acceptable loss when quantizing each parameter in the network, for which \texttt{OptSearchCNN} must find optimal $(BW, F)^*_{l, p}$. For layer $l$ and parameter $p$ during the execution of dependent \texttt{OptSearchCNN}, we can allocate an acceptable accuracy loss using

\begin{equation}
\label{eq:budget_alloc}
    \epsilon^D_{l, p} = \frac{\epsilon^D_{L, p}}{f(L)} \cdot f(l) 
\end{equation}
where $\epsilon^D_{L, p}$ is the final accuracy drop after quantizing all $L$ layers of parameter $p$ and the function $f(x)$ is a scheme we use to allocate the accuracy loss discussed below.

We experiment with 5 different schemes of allocating the acceptable accuracy loss as shown in Fig.~\ref{fig:ideal_budget_alloc} for multiple layers for just one parameter ($l = 1, \cdots, L$, $p \in \{ \mathbf{W}, \mathbf{B}, \mathbf{A} \}$):
\begin{enumerate}
    \item \textbf{Constant}($f(x) = 1$): $\epsilon^D_{l, p}$ is kept constant throughout the execution of \texttt{OptSearchCNN}. This allocation scheme spends the entire budget at the start of the algorithm.
    \item \textbf{Log}($f(x) = \ln(x)$): $\epsilon^D_{l, p}$ is increased using the natural logarithm. Most of the budget is allocated early, however this allocation rate is decreased later on. 
    \item \textbf{Linear}($f(x) = x$) Linear increase of $\epsilon^D_{l, p}$ resulting in steady allocation. Note that we also accept some loss when quantizing the parameter of the first layer with this approach.
    \item \textbf{Quadratic}($f(x) = x^2$): Quadratic increase in $\epsilon^D_{l, p}$. This approach is more conservative with allocating the budget.
    \item \textbf{Exponential}($f(x) = 2^x$):  Exponential increase in $\epsilon^D_{l, p}$ with base 2, using extremely conservative allocation initially.
\end{enumerate}

We also experimented these allocation schemes by reversing the order, that is $l = L, \cdots, 1$ and found no significant results. We therefore only present our results for $l = 1, \cdots, L$.

\begin{figure}
    \centering
    \includegraphics[width=0.46\textwidth]{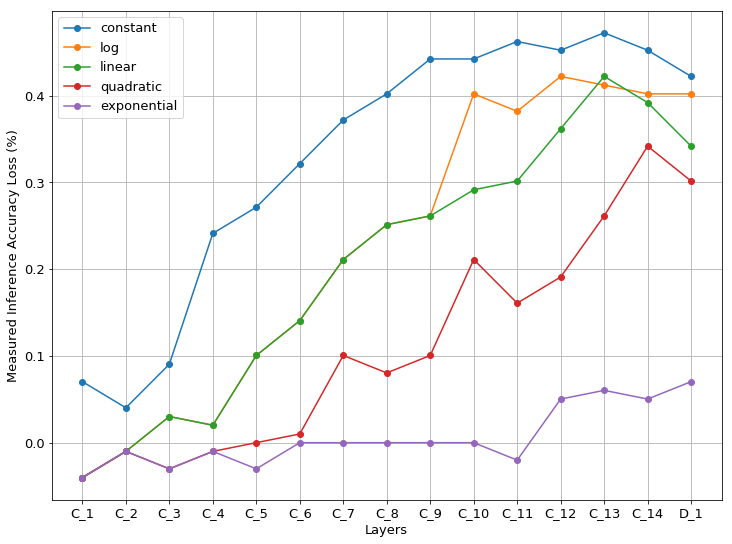}
    \caption{accuracy loss $\Delta a^D_{l, W}$ after quantizing weights of all layers to their optimal $(BW, F)^*_{l, W}$ found using Dependent Optimized Search for different acceptable loss allocation schemes. Weights of the preceding layer are also quantized to $(BW, F)^*_{k, W}$ for $k < l$.}
    \label{fig:dep_acc_loss}
\end{figure}

\begin{figure}
    \centering
    \includegraphics[width=0.48\textwidth]{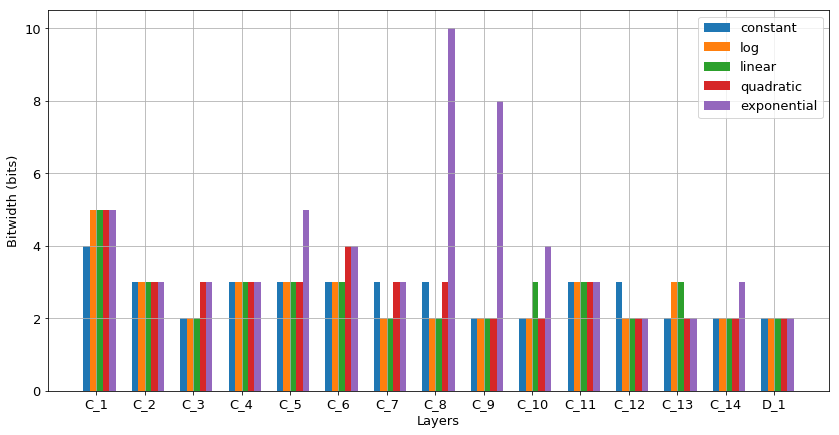}
    \caption{Optimal bitwidths $BW^*_{l, W}$ of weights of each layer found using Dependent Optimized Search for different acceptable loss allocation schemes on a 15 layer Sequential CNN (trained on MNIST). Weights of the preceding layer are also quantized to $(BW, F)^*_{k, W}$ for $k < l$.}
    \label{fig:dep_bw_seq}
\end{figure}

\texttt{OptSearchCNN} is initialized with the respective $\epsilon^D_{l, p}$ increasing up to a maximum of $\epsilon^D_{L, p} = 0.005$ (0.5\%) for parameter $p$, an initial bitwidth $BW^0_{l, p}$ of 12 bits and $\mathit{IND} = \mathit{False}$.

After successful termination of dependent \texttt{OptSearchCNN} with the five different methods of allocating $\epsilon^D_{L, p}$, \texttt{OptSearchCNN} returns optimal fixed-point representations $r_Q = r^D_Q$, where each $(BW, F)^*_{l, p}$ is found depending on how preceding parameters have been quantized up until that point.

To keep it concise, we present and discuss our results for executing dependent optimized search on the weights of one CNN while keeping activations at floating-point precision. We also investigate this on activations while keeping weights at floating-point precision, results for which can be found in Appendix C. The observations and conclusions are based on tests on all parameters of all the sequential models for the four datasets presented in Section \ref{sec:exp_setup}, the results for which can be found in Appendix C.

\begin{figure}
    \centering
    \includegraphics[width=0.48\textwidth]{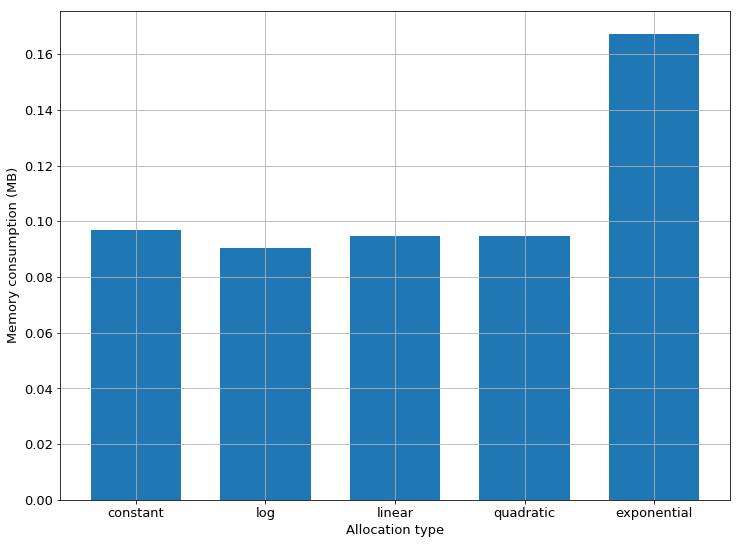}
    \caption{Total memory consumption of the quantized weights with optimal bitwidths $BW^*_{l, W}$ in Fig.~\ref{fig:dep_bw_seq} for the 15-layer Sequential CNN trained on MNIST.}
    \label{fig:dep_mem}
\end{figure}

Fig.~\ref{fig:dep_acc_loss} shows the measured accuracy loss $\Delta a^D_{l, W}$ using dependent \texttt{OptSearchCNN} and dependent \texttt{EvalAccLossCNN} for the five different schemes of allocating $\epsilon^D_{l, p}$ while the weights of a 15 layer sequential CNN (trained on MNIST) are being quantized one layer at a time. For all allocation schemes, \texttt{OptSearchCNN} is able to find optimal fixed-point representations for weights of all layers such that the accuracy loss is below the maximum acceptable accuracy loss of 0.5\%. Fig.~\ref{fig:dep_bw_seq} shows the optimal bitwidths for the quantized weights of each layer returned by dependent \texttt{OptSearchCNN} for the five different schemes of allocation, and also the respective memory consumption of the weights in Fig.~\ref{fig:dep_mem}.

\begin{figure*}[t!]
    \centering
    \begin{subfigure}[t]{0.45\textwidth}
        \centering
        \includegraphics[width=\textwidth]{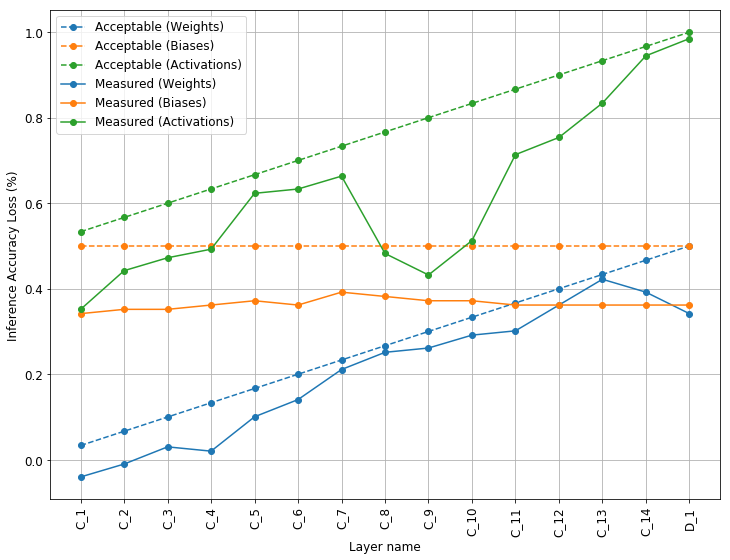}
        \caption{Acceptable and measured accuracy loss $\Delta a^D_{l, p}$ after sequentially quantizing the parameters $\mathbf{p}_l$ in the order ($\mathbf{W} \rightarrow \mathbf{B} \rightarrow \mathbf{A}$) from layers 1 to L.}
        \label{fig:opts_seq_acc_loss}
     \end{subfigure}
    \hspace{10mm}
    \begin{subfigure}[t]{0.45\textwidth}
        \centering
        \includegraphics[width=\textwidth]{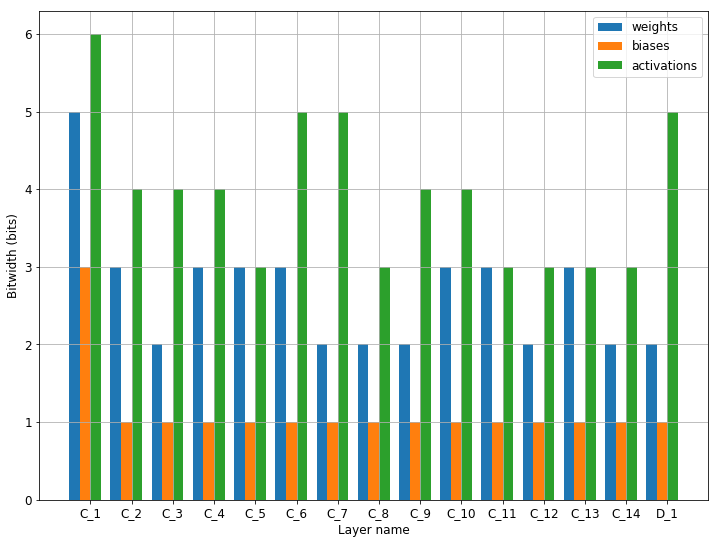}
        \caption{Optimal bitwidths $BW^*_{l, p}$ of parameters of each layer.}
        \label{fig:opts_seq_bw}
    \end{subfigure}
    \caption{Results for accuracy loss and the optimal bitwidths of the quantized CNN resulting from dependent optimized search using a linear acceptable loss allocation scheme on a 15-layer sequential CNN trained on MNIST.}
    \label{fig:opts_seq_results}
\end{figure*}
We now discuss our observations and conclusions for dependent optimized search based on the results collected for both weights and activations, here and in Appendix C. We group these based on the acceptable loss allocation schemes.
\begin{itemize}
    \item \textbf{Greedy Allocation:} The constant allocation scheme proves to be quite greedy as it results in extremely harsh quantization with low bitwidths for all layers therefore resulting in low memory consumption as observed in Fig.~\ref{fig:dep_mem}, but this automatically resulted in higher accuracy losses $\Delta a^D_{l, p}$ as compared to the other schemes as observed in Fig.~\ref{fig:dep_acc_loss}. Although this measured loss is below what is acceptable $(< 0.5\%)$, it essentially uses up most of the budget quite early in the execution of \texttt{OptSearchCNN}. For one of the cases, this scheme resulted in dependent \texttt{OptSearchCNN} using up too much of the budget early and quantizing initial layers too harshly, therefore being unable to find sufficiently low bitwidths for parameters of successive layers. 
    
    On the other hand, the log allocation for $\epsilon^D_{l, p}$ is also on the greedier side as it allocates most of the budget quite early while decreasing the rate of this allocation later. From our experiments, it resulted in similar if not larger accuracy losses than the Linear approach. However it usually performed better than the constant case as $\epsilon^D_{l, p}$ was allocated in a more controlled manner. The memory consumption and therefore the respective bitwidths were also comparable to the constant and linear schemes.
    
    \item \textbf{Linear Allocation:} The linear scheme is a simple one where the allocated acceptable accuracy loss is steadily increased for parameters of each layer. This generally resulted in lower accuracy losses than the constant and log cases while still achieving sufficiently low accuracy losses and low bitwidths as observed in Fig.~\ref{fig:dep_acc_loss}, \ref{fig:dep_bw_seq} and from the other results in Appendix C. Throughout our experiments, the linear allocation scheme mostly resulted in the lowest memory consumption for both weights and activations while having acceptable losses in inference accuracy. The simplicity of this allocation scheme proves to be quite useful for finding optimal fixed-point representations.
    
    \item \textbf{Conservative Allocation} The two conservative allocation schemes for acceptable accuracy loss, namely quadratic and exponential require dependent \texttt{OptSearchCNN} to find minimum bitwidth fixed-point representations while accepting almost no loss in inference accuracy for most of the parameters of the network as observed in Fig.~\ref{fig:ideal_budget_alloc}. This often resulted in extremely low measured accuracy losses $\Delta a^D_{l, p}$ for the entire quantized network as can be seen in Fig.~\ref{fig:dep_acc_loss}, ending with most of the acceptable budget still being available for further quantization. However, this automatically causes \texttt{OptSearchCNN} to find much larger bitwidths for parameters of each layer, as compared to the other allocation schemes resulting in a higher memory usage. Clearly, this is not optimal given that we can still quantize more aggressively, thus lowering the bitwidths of parameters of some layers, and accept a higher accuracy loss than what is currently measured.
\end{itemize}

\subsubsection{Optimal Allocation Scheme for Acceptable Accuracy Loss}
From our experiments on all the models for dependent optimized search as presented above, we find that the linear allocation of acceptable accuracy loss $\epsilon^D_{l, p}$ for dependent optimized search performs the best. Concretely, we note that the linear scheme ensures acceptable losses in inference accuracy and sufficiently low bitwidths thereby resulting in lower memory consumption for both weights and activations. We note that the linear scheme provides a reasonable balance between greedy and conservative quantization as the accuracy losses are acceptable and bitwidths found are the lowest. Additional empirical data to support this conclusion can be found in Appendix C. The final inference accuracy lost after quantizing all L layers $\Delta a^D_{L, p}$ is well below the maximum $\epsilon^D_{L, p} = 0.005$ with a small amount of the budget left to spare. 

From the observations and conclusions of Section \ref{sec:ind_opts} we noted that that the biases do not contribute much to the accuracy losses of the network. Therefore, we decided to use the constant allocation scheme for acceptable accuracy loss for quantization of biases of the CNN, with a value of $\epsilon^D_{l, B} = \epsilon^D_{L, W}$. Additionally, it helps to note that the linear approach is simple and straight forward.

\subsection{Linear Allocation Dependent Optimized Search}
From the results gathered on the allocation scheme for acceptable loss $\epsilon^D_{l, p}$ in inference accuracy when performing dependent optimized search, we noted that linear allocation worked best, giving sufficiently low accuracy losses for the resulting quantized network with low bitwidths and therefore a low memory consumption.

We now use this allocation scheme as part of our algorithm to quantize parameters of a 32-bit floating-point full precision CNN. We then evaluate the quantized network against the constraints in \eqref{eq:constraints} and based on the given metrics, compare it to a simple baseline approach that is used by a lot of commercial tools. We first present the baseline approach, then present and discuss the results to compare it against our approach.

\subsubsection{Baseline}
A simple approach to post-training quantization that is used by a number of commercial tools is that of quantizing the entire network to the same fixed-point representation. Given that hardware is restricted to 4, 8, 16, 32 bits, quantizing an entire network to the same representation is usually limited to 8 bits as lower bitwidths result in larger inference accuracy degradation.

For the fractional offset, the tools find a fractional offset given the bitwidth such that none of the parameter distributions are subjected to clipping. We used this approach as a starting point for our algorithm in \texttt{OptSearchCNN} as defined in Step 2 in Section \ref{sec:algodesign}. Concretely, we find a fractional offset $F$ for a given bitwidth $BW$ by assigning as many of the available integer bits in the bitwidth to the largest absolute valued number in the parameter distribution. This ensures that the range of values is covered avoiding any possibilities of clipping. The remaining bits are then assigned to the fractional part $F$ as is defined in \eqref{eq:noclip}.

For our baseline, we will use the fixed-bitwidth approach and quantize weights, biases and activations for all layers to a fixed-point representation with a bitwidth of 8 bits.

\subsubsection{Final method}
We now put together our final method based on the results collected earlier. From Section \ref{sec:dep_opts} we concluded that using a linear allocation scheme for the acceptable accuracy loss $\epsilon^D_{l, p}$ using dependent \texttt{OptSearchCNN} resulted in a sufficiently quantized network with bitwidths as low as 2 bits with an accuracy loss of the quantized network that was acceptable. Having tested this on weights and activations independently, keeping the other parameter at floating-point precision, we found that this method worked well in both cases. We now aim to quantize all three parameter types of all layers to sequentially obtain a CNN with all parameters quantized to their respective optimal fixed-point representations such that the final accuracy loss of the entirely quantized model is still acceptable, as is required by our constraints in \eqref{eq:constraints}.

For the acceptable accuracy loss $\epsilon$, we specify an acceptable loss of $\epsilon = 0.01$ (1\%) for the final quantized network. We use the first half of this budget for quantization of the weights of all layers $\epsilon_{L, W} = 0.005$ and the second half ($0.005 < \epsilon_{l, A} \leq 0.01$ for all $1\leq l \leq L$) for quantization of all activations of all layers. Therefore after termination of \texttt{OptSearchCNN}, $\epsilon_{L, A} = \epsilon = 0.01$ and we need that $\Delta a_{L, A} \leq \epsilon_{L, A} = \epsilon$. Using the linear allocation scheme defined earlier, we then specify the intermediate acceptable accuracy losses for weights and activations $\epsilon_{l, W}$, $\epsilon_{l, A}$ as was done previously.

For biases however, we use a constant allocation scheme given that our experiments showed quantized biases to have a minimal effect on the accuracy loss. Additionally, in Fig.~\ref{fig:dep_acc_loss} we noted that the measured accuracy loss $\Delta a^D_{L, W} < \epsilon^D_{L, W}$ after quantizing weights of all $L$ layers, and so the remainder of the budget should suffice for the biases. Hence, we specify $\epsilon_{l, B} = \epsilon_{L, W} = 0.005$ for all $1 \leq l \leq L$.

We now execute dependent \texttt{OptSearchCNN} by ordering the \texttt{for} loops on lines 2, 3 as ($\mathbf{W} \rightarrow \mathbf{B} \rightarrow \mathbf{A}$) from layers 1 to L. We use an initial bitwidth $BW^0_{l, p} = 12$ for all layers $l$ and all parameters $p$. To perform dependent optimized search we set the boolean flag $\mathit{IND} = \mathit{False}$. 

Fig.~\ref{fig:opts_seq_results} shows the result of dependent \texttt{OptSearchCNN} on a 15 layer sequential CNN trained on MNIST using the aforementioned scheme for allocating accuracy loss. Fig.~\ref{fig:opts_seq_acc_loss} presents the allocated values for acceptable accuracy loss $\epsilon_{l, p}$ and the respective measured accuracy loss $\Delta a_{l, p}$ and Fig.~\ref{fig:opts_seq_bw} shows the optimal bitwidths found for the respective weights, biases and activations.

\begin{figure}[t!]
    \centering
    \includegraphics[width=0.4\textwidth]{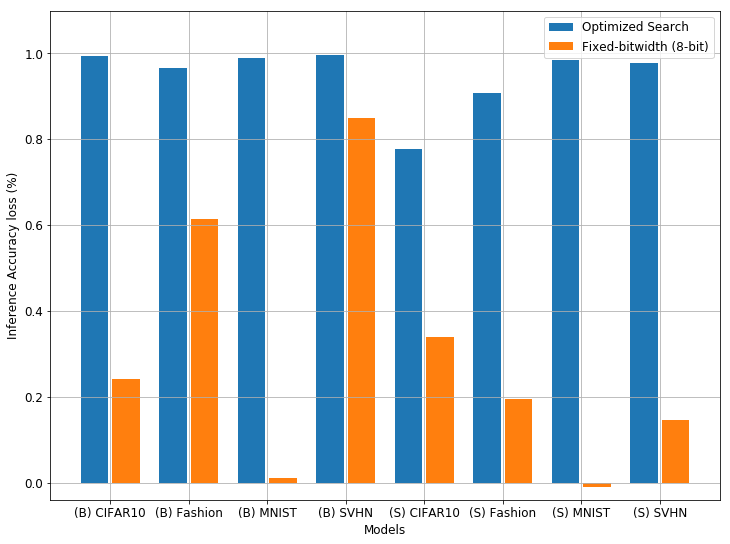}
    \caption{Comparing Dependent Optimized Search with linear acceptable loss allocation to a fixed-bitwidth approach (8-bit) on accuracy loss $\Delta a$ for (S)equential and (B)ranched CNN structures}
    \label{fig:opts_comp_acc_loss}
\end{figure}

\begin{figure*}[t!]
    \centering
    \begin{subfigure}[t]{0.45\textwidth}
        \centering
        \includegraphics[width=\textwidth]{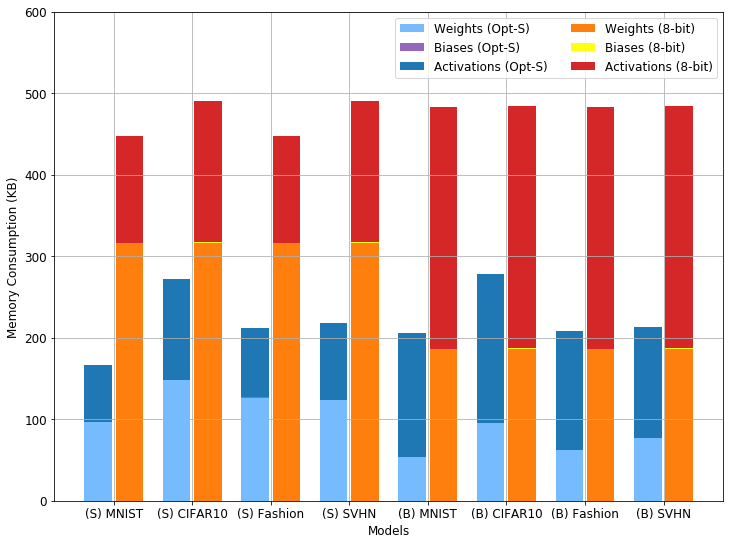}
        \caption{Total memory consumption of the quantized networks}
        \label{fig:opts_comp_mem}
    \end{subfigure}
    \hspace{10mm}
     \begin{subfigure}[t]{0.45\textwidth}
         \centering
         \includegraphics[width=\textwidth]{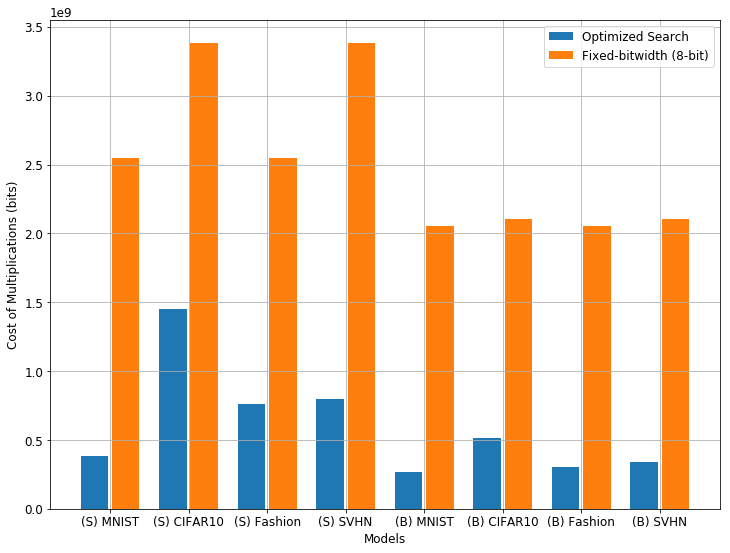}
         \caption{Cost of multiplications of the quantized networks}
         \label{fig:opts_comp_mult}
     \end{subfigure}
     \caption{Comparing Dependent Optimized Search with linear acceptable loss allocation scheme to a fixed-bitwidth approach (8 bit) on memory consumption and cost of multiplications for Sequential(S) and Branched(B) CNN structures}
     \label{fig:opts_comp_results}
\end{figure*}

We note in Fig.~\ref{fig:opts_seq_acc_loss} that $\Delta a_{l, p} \leq \epsilon_{l, p}$ up until all parameters are quantized to their optimal fixed-point representations. After quantization of all the weights, the accuracy loss of the network is approximately $\Delta a_{L, W} = 0.0036$ (0.36\%). Keeping the quantized weights, we find that sequential quantization of biases with the constant $\epsilon_{l, B}$ leaves the accuracy loss almost unchanged. Finally, with the quantized weights and biases, we find that linearly increasing $\epsilon_{l, A}$ up to $\epsilon_{L, A}$ to sequentially find fixed-point representations for the activations of all layers results in an accuracy loss that steadily rises and stays below 1\% in the end, as was originally desired.

In Fig.~\ref{fig:opts_seq_acc_loss} we also note how the accuracy drop reduces for certain parameters (examples: activations of layer C\_8 and weights of layer C\_14). In such cases, when dependent \texttt{OptSearchCNN} finds an appropriate optimal bitwidth, the $\Delta a^D_{l, p}$ ends up being lower than $\Delta a^D_{l-1, p}$. The corresponding bitwidth for activations of layer C\_8 is also quite low (3 bits). We believe that this may happen due to possible regularization effects or compensating multiplications in that layer that make up for the errors due the quantization of parameters of preceding layers.

For this acceptable loss in inference accuracy, dependent \texttt{OptSearchCNN} was able to find optimal fixed-point representations $(BW, F)^*_{l, p}$ with bitwidths of 2-5 bits for weights, 1-3 bits for biases and 3-6 bits for activations as seen in Fig.~\ref{fig:opts_seq_bw}. We also observe that the bitwidths required to represent the parameter values of each layer vary across the different layers and the parameter types. We discuss these discrepancies in Section \ref{sec:discussion}.

In Fig.~\ref{fig:opts_seq_bw} we observe how dependent optimized search is able to adapt each optimal bitwidths of each layer based on how the parameters of the network have already been quantized. When the network is quantized too harshly, the parameters of the successive layer are quantized more conservatively with larger bitwidths to compensate and so this bitwidth varies for each layer. This is solely because of the fact that the evaluation of the accuracy loss now includes the other quantized parameter values in the network.

Additional results with the other models and datasets can be found in Appendix D.

We experimented with changing the order in which \texttt{OptSearchCNN} looks for optimal fixed-point representations for the parameter types to investigate whether the order in which we performed this mattered, as discussed in Section \ref{sec:orderquant}. We noted 6 possible ways in which we could order the execution of the algorithm for the parameter types. Through our experiments we were able to limit this to two, given that the biases had a very minor effect on the accuracy loss and the bitwidths it was able to find and hence only experimented with ($\mathbf{W} \rightarrow \mathbf{B} \rightarrow \mathbf{A}$) or ($\mathbf{A} \rightarrow \mathbf{W} \rightarrow \mathbf{B}$). While the results for the former approach were presented and discussed, the dependent \texttt{OptSearchCNN} did not work with the latter approach. Concretely, when searching for $(BW, F)^*_{l, W}$ after finding all the $(BW, F)^*_{l, A}$ the \texttt{OptSearchCNN} would often reach a point where the network was quantized too much to be able to find a low bitwidth for which the accuracy loss was acceptable. 

While we could not find a concrete reason for why ($\mathbf{A} \rightarrow \mathbf{W} \rightarrow \mathbf{B}$) did not work while ($\mathbf{W} \rightarrow \mathbf{B} \rightarrow \mathbf{A}$) clearly did, we noticed that the optimal bitwidths found by dependent \texttt{OptSearchCNN} for activations varied in the two cases. The variation was not conclusive as activation bitwidths were lower for some layers while being higher for others. It is difficult to evaluate why this variation exists given the non-linearity in the network. However, we hypothesize that the multiply-accumulates might work differently in the two cases, and that may have an effect on the fixed-point representations needed to minimize deviation from of the values from their original floating-point values. From our experiments, weights needed to be quantized first.

\subsubsection{Final method vs Baseline}
We now compare the results of our aforementioned method against the baseline of an 8-bit fixed-bitwidth for all parameters in the network. The aim is to benchmark the resulting quantized model of our algorithm against that generated by common commercial tools on the metrics defined in the constraints in \eqref{eq:constraints}, namely accuracy loss, memory consumption and cost of multiplications.

Fig.~\ref{fig:opts_comp_acc_loss} shows the accuracy loss $\Delta a$ of the quantized models provided by the two approaches for each pre-trained CNN model based on the two architectures namely Sequential (S) and Branched (B) trained on the four datasets. Fig.~\ref{fig:opts_comp_results} shows the results for the calculations of memory consumption and cost of multiplications respectively. 

These results clearly show that the 8-bit quantized model have lower accuracy losses than those generated by \texttt{OptSearchCNN}. However, given that dependent \texttt{OptSearchCNN} is able to find lower bitwidths for the parameters, the quantized models resulting from our method of dependent \texttt{OptSearchCNN} consume between 42-62\% (average 53\%) lower memory than the 8-bit fixed-bitwidth approach. The quantized models resulting from dependent \texttt{OptSearchCNN} also have a 60-87\% (average 77.5\%) lower cost of multiplications as compared with the fixed-bitwidth approach. Given that the energy consumption depends on the cost of its operations, we note that the quantized models from \texttt{OptSearchCNN} would effectively consume less energy than the 8-bit bitwidth models as noted in Fig.~\ref{fig:opts_comp_mult}. 

Comparing a 32-bit floating-point precision CNN, an 8-bit implementation provides a significant reduction in memory consumption and energy consumption at the cost of some accuracy loss. However our method finds even more compressed and efficient CNN models by giving up a little more accuracy loss. Compared to the 32-bit floating-point CNN, the quantized models from dependent \texttt{OptSearchCNN} consume an average of 88.4\% (8.6x) less memory. 

With regards to our approach of \texttt{OptSearchCNN}, for these final tests, we noted that in the branched CNNs \texttt{OptSearchCNN} was able to reduce the bitwidths of parameters corresponding to certain branches down to 1 bit, which in our paradigm translates to pruning of the values of a layer. We noticed this to be the case for the branched CNN trained on CIFAR10, SVHN and Fashion-MNIST. Empirical data for the same may be found in Appendix D. We noted that the branches that were pruned were usually of the same type, namely the branch with a pooling layer followed by the convolutional layer in it. This pruning effect had minimal effect on inference accuracy and \texttt{OptSearchCNN} was able to therefore remove redundancies in the network and only keep layers and branches that produced important features.

It is clear that while the classification accuracy of the 8-bit implementation is close to that of the full-precision model, it trades-off with having larger bitwidths, which dependent \texttt{OptSearchCNN} is able to optimize for. The quantized models resulting from dependent \texttt{OptSearch} using a linear allocation scheme for acceptable loss results in significantly compressed models with accuracy losses just under 1\%.

\subsection{Discussion}
\label{sec:discussion}
We highlight and potentially explain some general observations presented from our experiments in this paper.

\subsubsection{Layer-wise quantization}
Fig.~\ref{fig:opts_seq_bw} clearly shows the advantage of finding optimal fixed-point representations for each parameter in the CNN layer-wise. In comparison, while the 8-bit fixed-bitwidth approach resulted in low inference accuracy losses, their quantized models consumed more memory. We also tried using a lower bitwidth than 8-bits for the fixed-bitwidth approach to quantize the models tested in Fig.~\ref{fig:opts_comp_results}, however this generally led to a large degradation in inference accuracy ($> 40\%$). One reason for this can be seen more closely in Fig.~\ref{fig:opts_seq_bw} and in other similar figures found in Appendix D, that clearly some layers require larger bitwidths than others in order to minimize the accuracy loss. The weights of the first layer always required a larger bitwidth than the successive layers thereafter. For activations however, this varied as the first and last layers generally required larger bitwidths. Additionally, biases were quantized to low bitwidths for only a few of the layers in the network.

Given that the first layer has the least number of parameters, therefore fewer redundancies, and also is the layer that generates the initial feature maps, we expected it to require more bits to sufficiently represent the respective parameters. Additionally, given that we accept a very low accuracy loss for the weights and activations of the first layer, dependent \texttt{OptSearchCNN} quantizes these parameters in the first layer very conservatively. This observation and conclusion is also widely supported in literature as was discussed in Section \ref{sec:relatedwork}, that the first layers need to be quantized conservatively to retain the inference accuracy of the network. From our results in Appendix D and in Fig.~\ref{fig:opts_seq_bw}, we also often noticed that activations for the last layer required more bits than for the other layers in the network. This data partially supports the work of \cite{choi2018pact} that notes that conservative quantization is required for the first and last layer to minimize the inference accuracy lost. We reason that since the activations of the last layer classify the image into one of the 10 classes, that quantizing this layer too harshly may result in images being wrongly classified as activations are switched one way or another.

Dependent \texttt{OptSearchCNN} using the linear allocation scheme for acceptable inference accuracy loss was successfully able to quantize the initial and last layer conservatively as needed while varying the bitwidths for the parameters per layer, and also compensating for harsh quantization if needed.

\subsubsection{Parameter types discrepancies}
From all our results, whether they pertained to the brute-force plots in Section \ref{sec:bf_analysis} or the results of Optimized Search, we noticed that activations generally required equal or larger bitwidths than those needed for weights. From this we concluded that activations were generally more sensitive to quantization than weights, and generally required larger bitwidths to minimize accuracy degradation of the CNN. We found this to be consistent with works in literature, where the quantized models resulting from their approaches also required larger bitwidths for activations than for weights. 

To investigate this discrepancy, we studied the the distribution of the weights and activations of all layers of the network, similar to Fig.~\ref{fig:quant_dist}. We observed that the distribution of activations had much longer tails than did the distribution of weights. Concretely, activations covered a much larger range of numbers than did weights. Often there were many outlying activations at values much greater than 20 (a few activations with values of 80). On the other hand, the distribution of weights usually followed a bell-shaped curve centered around zero and had values that at most reached 5. Given that the size of our bitwidth gives an indication of the range of values it can cover as noted in \eqref{eq:max_val}, clearly larger bitwidths are required to sufficiently represent the entire distribution of activations. As we reduce the bitwidth and fractional offset together, we increase the value of the LSB and therefore lose precision as intermediate values are rounded to discrete values.

From our experiments this loss of precision for activations seems to affect the accuracy of the network much more than it does for weights. This was observed in Fig.~\ref{fig:bf_results}, \ref{fig:ind_acc_loss_layer}. Since there were a very small fraction (0.05\%) of outlying activations, we also noticed that these could be clipped to a slightly lower bitwidth without much accuracy loss (reduction in 1-2 bits roughly). Any further reduction in the bitwidth and potentially also the fractional offset for the quantization of activations would result in too large of a loss of precision therefore leading to more significant accuracy degradation.

Since the distribution of weights were more clustered around zero, and the distribution itself covered a very small range of values, smaller bitwidths could be used to represent them with lower accuracy losses.

We recommend further study into the reasons for which quantization of activations have a larger effect on the accuracy of the network do the quantization of weights.

Additionally, biases were often pruned (1-bit) which is counter intuitive considering the fact that the biases would have a major contribution to the activations. However, we did not observe this to be the case. From our experiments, we conclude that most of the biases could effectively be removed without any effect on the accuracy of the network.

\subsubsection{Advantage of Arbitrary Bitwidths}
From the results in Fig.~\ref{fig:opts_seq_bw} and the results in Appendix D, we observed that \texttt{OptSearchCNN} often suggested optimal bitwidths like 2, 3 or 5 bits, essentially bitwidths that do not conform to those used in hardware. While it is quite difficult to create dedicated hardware to support these arbitrary bitwidths, our experiments clearly show that there is an incentive to work towards developing hardware with arbitrary bitwidths as they can allow networks to be quantized and compressed more efficiently, especially with the layer-wise quantization method.

\subsubsection{Dependent Optimized Search vs K-means approaches}
As highlighted in Section \ref{sec:relatedwork}, approaches that utilize K-means clustering to quantize parameters of networks cannot control performance loss due to quantization. Dependent Optimized Search on the other hand finds minimum bitwidths based on a user defined acceptable loss in inference accuracy. This makes the problem more controllable. 

Another disadvantage is that the K-means approach is only applied to reduce memory consumption for the weights of the model. The RAM used due to the activations is not considered in this process. Using Dependent Optimized Search, we also find low bitwidths to quantize activations therefore reducing the respective RAM requirements. 

Additionally, while K-means based approaches are able to reduce memory consumption by representing fewer weights, the centroids themselves are 32-bit floating point values therefore still relying on floating-point operations. With Optimized Search, we eliminate the need for floating-point precision entirely by reducing the precision of all parameters to low bitwidth fixed-point representations.

\subsubsection{Implementation concerns}
As mentioned earlier in this paper, we noted that our work was performed in simulation on a CPU and GPU using the Keras API to handle CNN models easily and using Python to perform experiments on the CNN models. The main reason for our work in simulation is the fact that Keras does not allow for the use of integer precision. We therefore needed to scale the values back down to floating-point using $2^F$ in \eqref{eq:quantfunc}. 

With the quantization function in \eqref{eq:quantfunc}, we simulated quantization by creating a map between floating-point numbers and the fixed-point equivalent values in floating-point. Therefore, the quantized parameters of the CNN resulting from \texttt{OptSearchCNN} still uses floating-point precision to represent the fixed-point numbers that we have mapped the original parameters to. All the calculations involving low-precision fixed-point numbers are still in floating-point and therefore occur on floating-point ALUs in the CPU and GPU. Therefore we cannot directly measure the benefit of the decrease in computational cost and memory consumption using our setup. 

To truly evaluate the performance benefit of our quantized model, inference of the quantized CNN would have to be implemented in a lower level language like C and the parameters would have to be represented as integers (simply removing division by $2^F$ in \eqref{eq:quantfunc}). Testing the resulting C program on a CPU would show the true benefit with respect to computational cost and memory consumption as CPUs will rely on integer ALUs if all the numbers in the network including the input images are represented as integers. The quantized model can also then be implemented on more constrained hardware like micro-controllers and FPGAs, possibly requiring some additional lower-level optimizations for memory and computations.

However, evaluating the performance benefit would be quite difficult considering the fact that our low-precision model relies on arbitrary bitwidths for the parameters which are not supported by CPUs and other common hardware today. Therefore we would have to round up the optimal bitwidths that we found to bitwidths of 4, 8 or 16 bits. In such a case, the memory consumption and computational cost would be higher than what we found through our experimentation. However, the model would still be optimized using layer-wise quantization. This is where the baseline approach and papers such as \cite{banner2018posttraining} have an advantage, in that their quantized models use bitwidths of 8 and 4 bits respectively and can therefore be implemented and tested on hardware such as CPUs, micro-controllers etc.

In any case, we believe that our work should provide some incentive into developing hardware that supports arbitrary bitwidths, especially for neural networks as we clearly observe benefits from it from our experiments.

\subsubsection{Applicability and Limitations}
Our work and the resulting conclusions were drawn based on the experiments on two types of architectures trained on four common datasets. However a lot of the conclusions we drew supported the observations and conclusions seen in literature. For example, the conclusion that the first and last layers required larger bitwidths and more conservative quantization is also supported by works in the literature. We also expect the conclusions drawn from brute-force analysis to be applicable to quantization of parameters on other CNNs trained on the other datasets therefore allowing our algorithm to work on these other CNNs and datasets as well. 

To further validate the conclusions we have drawn, we recommend testing the algorithm on more datasets like ImageNet, and on much larger and more complex models like the ones commonly tested in the literature. Due to time constraints, we chose to restrict ourselves to models that could quickly be trained on a desktop/laptop GPU within an hour to allows us to collect more results. Additionally, the more complex models like AlexNet and ResNet are quite large in size and would not be ones that would be ported to constrained hardware such as micro-controllers, given the small amount of memory on-chip. 

We do also recommend testing our algorithm on CNNs for regression tasks, image denoising tasks, segmentation tasks, CNNs with skip connections (ResNet) and for some other common tasks involving CNNs with other varied architectures.

Our algorithm should generally apply to the other aforementioned cases given that we simply look at how to minimize the bitwidth of a group of numbers such that the accuracy loss (an evaluation metric) is within acceptable bounds. We can expect to find some new observations and draw some new conclusions for those cases. We limited ourselves to CNNs for image classification to restrict scope and also because it is a common application.

\section{Conclusions and further work}
\label{sec:conclusion}
In this paper, we designed a pipeline to quantize the parameters of each layer of a pre-trained CNN based on a quantization function that converted floating-point numbers to a fixed-point representation characterized by a bitwidth and fractional offset. Using this pipeline, we analyzed the effects of changing the fixed-point representations used to quantize the respective parameters on the CNN on the inference accuracy. The predictable pattern observed was the basis for the design of a method (\texttt{OptSearchCNN}) to efficiently search for the optimal bitwidths and fractional offsets for all parameters for each layer of a given pre-trained CNN. We also investigated two approaches (independent and dependent) to using this method that was based on whether we took the errors due to quantization of other parameter values into account. Results showed that sequentially quantizing the network in a controlled manner by controlling the acceptable accuracy loss at each stage proves to give the best results with respect to accuracy loss, memory consumption and cost of multiplications, at least for the two architectures trained on the four datasets that we tested our work on. Ignoring the effects of quantization from other parameters in the network proves to be error prone resulting in a CNN with large losses in inference accuracy. Our resulting method, dependent optimized search was then compared against a common baseline approach proving to be advantageous due to the lower bitwidths that \texttt{OptSearchCNN} was able to find.

Following are our recommendations for further work:
\begin{itemize}
    \item Swapping the order of the for-loops in \texttt{OptSearchCNN} thereby interleaving the search for fixed-point representations and searching across all parameters sequentially per layer, as compared to our current approach where we execute optimized search one parameter at a time for all layers. 
    \item Using a more fine-grained approach by quantizing kernels differently rather than the layer-wise granularity used in this paper
    \item Investigating whether the optimal bitwidths found by \texttt{OptSearchCNN} are over-fit to the dataset used and if the inference accuracy would vary when testing this quantized model on new data
    \item Implementing the resulting quantized model with the bitwidths suggested by \texttt{OptSearchCNN} on a hardware platform to understand how close our simulated results measure up in reality.
    \item Execute \texttt{OptSearchCNN} on larger and more complex models (AlexNet, ResNet etc.) with more complex datasets like ImageNet.
    \item Investigating why certain weights/activations distributions can be quantized to a lower precision than others and if this has something to do with the properties of the CNN.
    \item Using this approach on other types of Neural networks for other types of problems (Examples: Regression, Segmentation, RNNs, Image Denoising).
    \item Development of hardware to support arbitrary bitwidths
    \item Investigating why the accuracy losses due to quantized activations are generally larger than the accuracy losses due to quantized weights
    \item Investigating the reasons why the order in which parameters are quantized matters. 
    \item Testing \texttt{OptSearchCNN} with varied acceptable accuracy losses to determine the least acceptable accuracy loss possible for the given CNN trained on a dataset.
    \item Investigating the relationship between the quantization error after quantizing a certain distribution of parameters and the accuracy loss of the network. 
    \item Development of a method to quantize parameters of each layer independently of other parameters in the CNN. This would then allow for parameters in the network to be quantized in a random order. 
\end{itemize}

\bibliography{references.bib}{}

\begin{thebibliography}{10}
\providecommand{\url}[1]{#1}
\csname url@samestyle\endcsname
\providecommand{\newblock}{\relax}
\providecommand{\bibinfo}[2]{#2}
\providecommand{\BIBentrySTDinterwordspacing}{\spaceskip=0pt\relax}
\providecommand{\BIBentryALTinterwordstretchfactor}{4}
\providecommand{\BIBentryALTinterwordspacing}{\spaceskip=\fontdimen2\font plus
\BIBentryALTinterwordstretchfactor\fontdimen3\font minus
  \fontdimen4\font\relax}
\providecommand{\BIBforeignlanguage}[2]{{%
\expandafter\ifx\csname l@#1\endcsname\relax
\typeout{** WARNING: IEEEtranS.bst: No hyphenation pattern has been}%
\typeout{** loaded for the language `#1'. Using the pattern for}%
\typeout{** the default language instead.}%
\else
\language=\csname l@#1\endcsname
\fi
#2}}
\providecommand{\BIBdecl}{\relax}
\BIBdecl

\bibitem{anwar2015fixed}
S.~Anwar, K.~Hwang, and W.~Sung, ``Fixed point optimization of deep
  convolutional neural networks for object recognition,'' in \emph{2015 IEEE
  International Conference on Acoustics, Speech and Signal Processing
  (ICASSP)}.\hskip 1em plus 0.5em minus 0.4em\relax IEEE, 2015, pp. 1131--1135.

\bibitem{banner2018posttraining}
R.~Banner, Y.~Nahshan, and D.~Soudry, ``Post training 4-bit quantization of
  convolutional networks for rapid-deployment,'' in \emph{Advances in Neural
  Information Processing Systems}, H.~Wallach, H.~Larochelle, A.~Beygelzimer,
  F.~d\textquotesingle Alch\'{e}-Buc, E.~Fox, and R.~Garnett, Eds.,
  vol.~32.\hskip 1em plus 0.5em minus 0.4em\relax Curran Associates, Inc.,
  2019, pp. 7950--7958.

\bibitem{cheng2018differentiable}
H.-P. Cheng, Y.~Huang, X.~Guo, Y.~Huang, F.~Yan, H.~Li, and Y.~Chen,
  ``Differentiable fine-grained quantization for deep neural network
  compression,'' \emph{arXiv preprint arXiv:1810.10351}, 2018.

\bibitem{choi2018pact}
J.~Choi, Z.~Wang, S.~Venkataramani, P.~I.-J. Chuang, V.~Srinivasan, and
  K.~Gopalakrishnan, ``Pact: Parameterized clipping activation for quantized
  neural networks,'' \emph{arXiv preprint arXiv:1805.06085}, 2018.

\bibitem{choi2016towards}
Y.~Choi, M.~El-Khamy, and J.~Lee, ``Towards the limit of network
  quantization,'' \emph{arXiv preprint arXiv:1612.01543}, 2016.

\bibitem{choukroun2019low}
Y.~{Choukroun}, E.~{Kravchik}, F.~{Yang}, and P.~{Kisilev}, ``Low-bit
  quantization of neural networks for efficient inference,'' in \emph{2019
  IEEE/CVF International Conference on Computer Vision Workshop (ICCVW)}, 2019,
  pp. 3009--3018.

\bibitem{courbariaux2014training}
M.~Courbariaux, Y.~Bengio, and J.-P. David, ``Training deep neural networks
  with low precision multiplications,'' \emph{arXiv preprint arXiv:1412.7024},
  2014.

\bibitem{courbariaux2015binaryconnect}
------, ``Binaryconnect: Training deep neural networks with binary weights
  during propagations,'' in \emph{Advances in Neural Information Processing
  Systems}, C.~Cortes, N.~Lawrence, D.~Lee, M.~Sugiyama, and R.~Garnett, Eds.,
  vol.~28.\hskip 1em plus 0.5em minus 0.4em\relax Curran Associates, Inc.,
  2015, pp. 3123--3131.

\bibitem{courbariaux2016binarized}
M.~Courbariaux, I.~Hubara, D.~Soudry, R.~El-Yaniv, and Y.~Bengio, ``Binarized
  neural networks: Training deep neural networks with weights and activations
  constrained to +1 or -1,'' \emph{arXiv preprint arXiv:1602.02830}, 2016.

\bibitem{gong2014compressing}
Y.~Gong, L.~Liu, M.~Yang, and L.~Bourdev, ``Compressing deep convolutional
  networks using vector quantization,'' \emph{arXiv preprint arXiv:1412.6115},
  2014.

\bibitem{guo2018survey}
Y.~Guo, ``A survey on methods and theories of quantized neural networks,''
  \emph{arXiv preprint arXiv:1808.04752}, 2018.

\bibitem{han2015deep}
S.~Han, H.~Mao, and W.~J. Dally, ``Deep compression: Compressing deep neural
  networks with pruning, trained quantization and huffman coding,'' \emph{arXiv
  preprint arXiv:1510.00149}, 2015.

\bibitem{jacob2018quantization}
B.~Jacob, S.~Kligys, B.~Chen, M.~Zhu, M.~Tang, A.~Howard, H.~Adam, and
  D.~Kalenichenko, ``Quantization and training of neural networks for efficient
  integer-arithmetic-only inference,'' in \emph{Proceedings of the IEEE
  Conference on Computer Vision and Pattern Recognition}, 2018, pp. 2704--2713.

\bibitem{krishnamoorthi2018quantizing}
R.~Krishnamoorthi, ``Quantizing deep convolutional networks for efficient
  inference: A whitepaper,'' \emph{arXiv preprint arXiv:1806.08342}, 2018.

\bibitem{nic:speedops}
\BIBentryALTinterwordspacing
N.~Limare, ``Integer and floating-point arithmetic speed vs precision,'' 2015,
  [accessed 5-October-2019]. [Online]. Available:
  \url{http://nicolas.limare.net/pro/notes/2014/12/12\_arit\_speed/}
\BIBentrySTDinterwordspacing

\bibitem{lin2016fixed}
D.~Lin, S.~Talathi, and S.~Annapureddy, ``Fixed point quantization of deep
  convolutional networks,'' in \emph{International Conference on Machine
  Learning}, 2016, pp. 2849--2858.

\bibitem{lin2016overcoming}
D.~D. Lin and S.~S. Talathi, ``Overcoming challenges in fixed point training of
  deep convolutional networks,'' \emph{arXiv preprint arXiv:1607.02241}, 2016.

\bibitem{rastegari2016xnor}
M.~Rastegari, V.~Ordonez, J.~Redmon, and A.~Farhadi, ``Xnor-net: Imagenet
  classification using binary convolutional neural networks,'' in
  \emph{European Conference on Computer Vision}.\hskip 1em plus 0.5em minus
  0.4em\relax Springer, 2016, pp. 525--542.

\bibitem{shin2017fixed}
S.~Shin, Y.~Boo, and W.~Sung, ``Fixed-point optimization of deep neural
  networks with adaptive step size retraining,'' in \emph{2017 IEEE
  International conference on acoustics, speech and signal processing
  (ICASSP)}.\hskip 1em plus 0.5em minus 0.4em\relax IEEE, 2017, pp. 1203--1207.

\bibitem{stock2019and}
P.~Stock, A.~Joulin, R.~Gribonval, B.~Graham, and H.~J{\'e}gou, ``And the bit
  goes down: Revisiting the quantization of neural networks,'' \emph{arXiv
  preprint arXiv:1907.05686}, 2019.

\bibitem{szegedy2016rethinking}
C.~Szegedy, V.~Vanhoucke, S.~Ioffe, J.~Shlens, and Z.~Wojna, ``Rethinking the
  inception architecture for computer vision,'' in \emph{Proceedings of the
  IEEE conference on computer vision and pattern recognition}, 2016, pp.
  2818--2826.

\bibitem{wu2018training}
\BIBentryALTinterwordspacing
S.~Wu, G.~Li, F.~Chen, and L.~Shi, ``Training and inference with integers in
  deep neural networks,'' in \emph{International Conference on Learning
  Representations}, 2018. [Online]. Available:
  \url{https://openreview.net/forum?id=HJGXzmspb}
\BIBentrySTDinterwordspacing

\bibitem{zhao2019improving}
R.~Zhao, Y.~Hu, J.~Dotzel, C.~De~Sa, and Z.~Zhang, ``Improving neural network
  quantization without retraining using outlier channel splitting,'' in
  \emph{International Conference on Machine Learning}, 2019, pp. 7543--7552.

\bibitem{zhou2016dorefa}
S.~Zhou, Y.~Wu, Z.~Ni, X.~Zhou, H.~Wen, and Y.~Zou, ``Dorefa-net: Training low
  bitwidth convolutional neural networks with low bitwidth gradients,''
  \emph{arXiv preprint arXiv:1606.06160}, 2016.

\end{thebibliography}
\bibliographystyle{IEEEtranS}

\newpage
\section*{Supplementary Material}
\subsection{Brute Force analysis on CNNs}
\begin{figure*}
    \centering
    \includegraphics[width=0.7\textwidth]{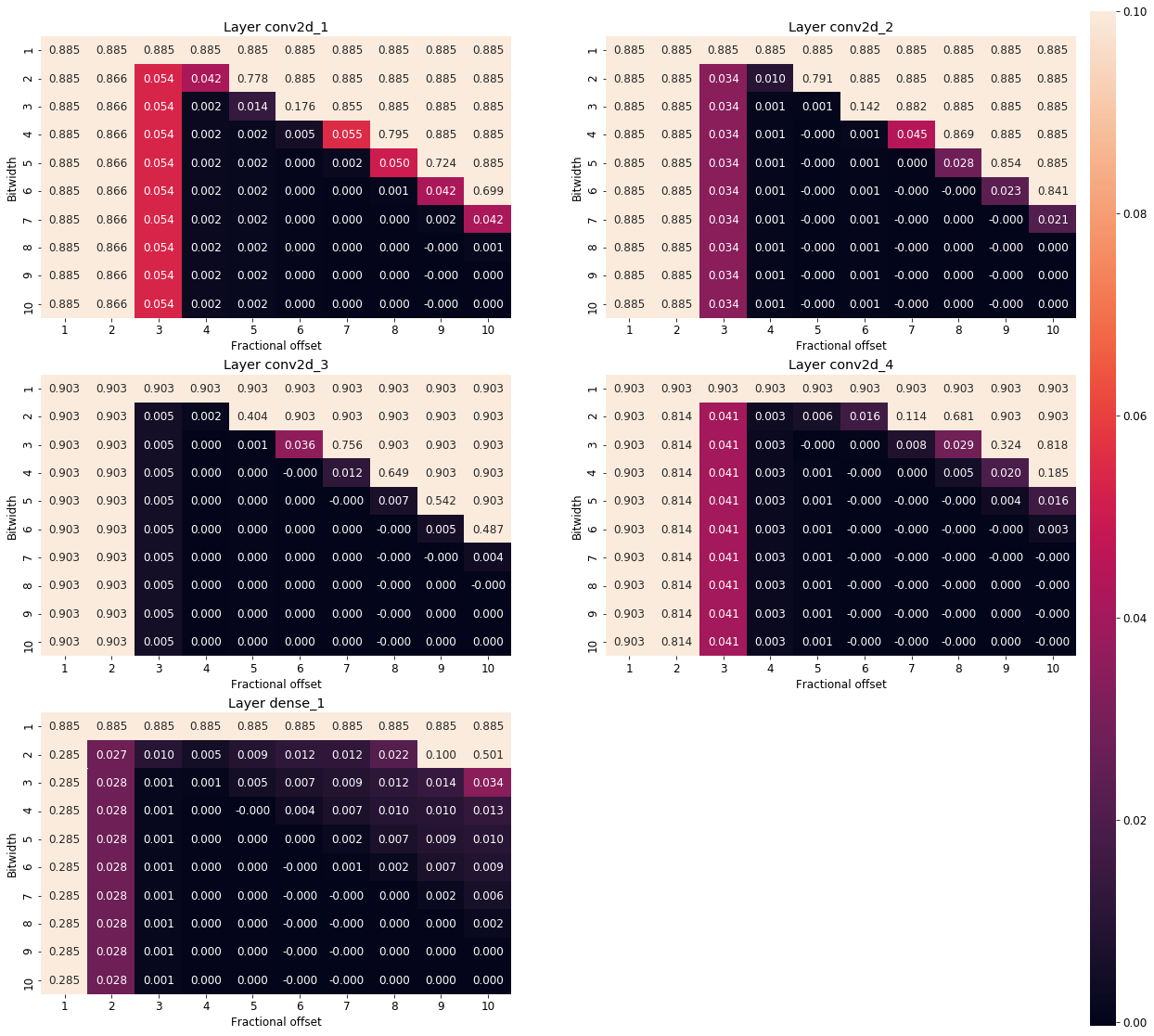}
    \caption{Brute force analysis of bitwidth and fractional offset for quantization of weights of a 5 layer CNN model trained on MNIST}
    \label{fig:bf_mnist_w}
    \centering
    \includegraphics[width=0.7\textwidth]{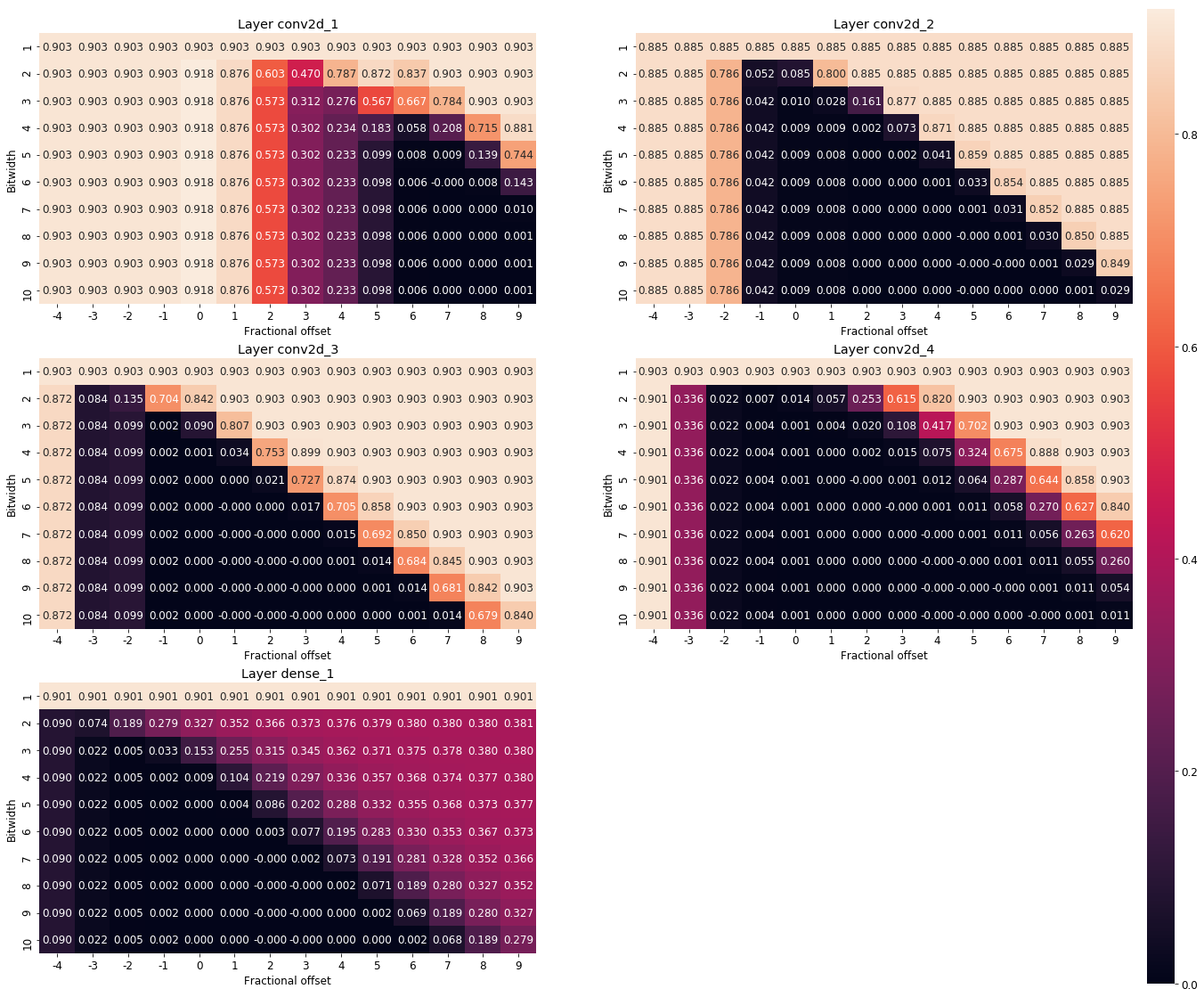}
    \caption{Brute force analysis of bitwidth and fractional offset for quantization of activations of a 5 layer CNN model trained on MNIST}
    \label{fig:bf_mnist_a}
\end{figure*}

\begin{figure*}
    \centering
    \includegraphics[width=0.7\textwidth]{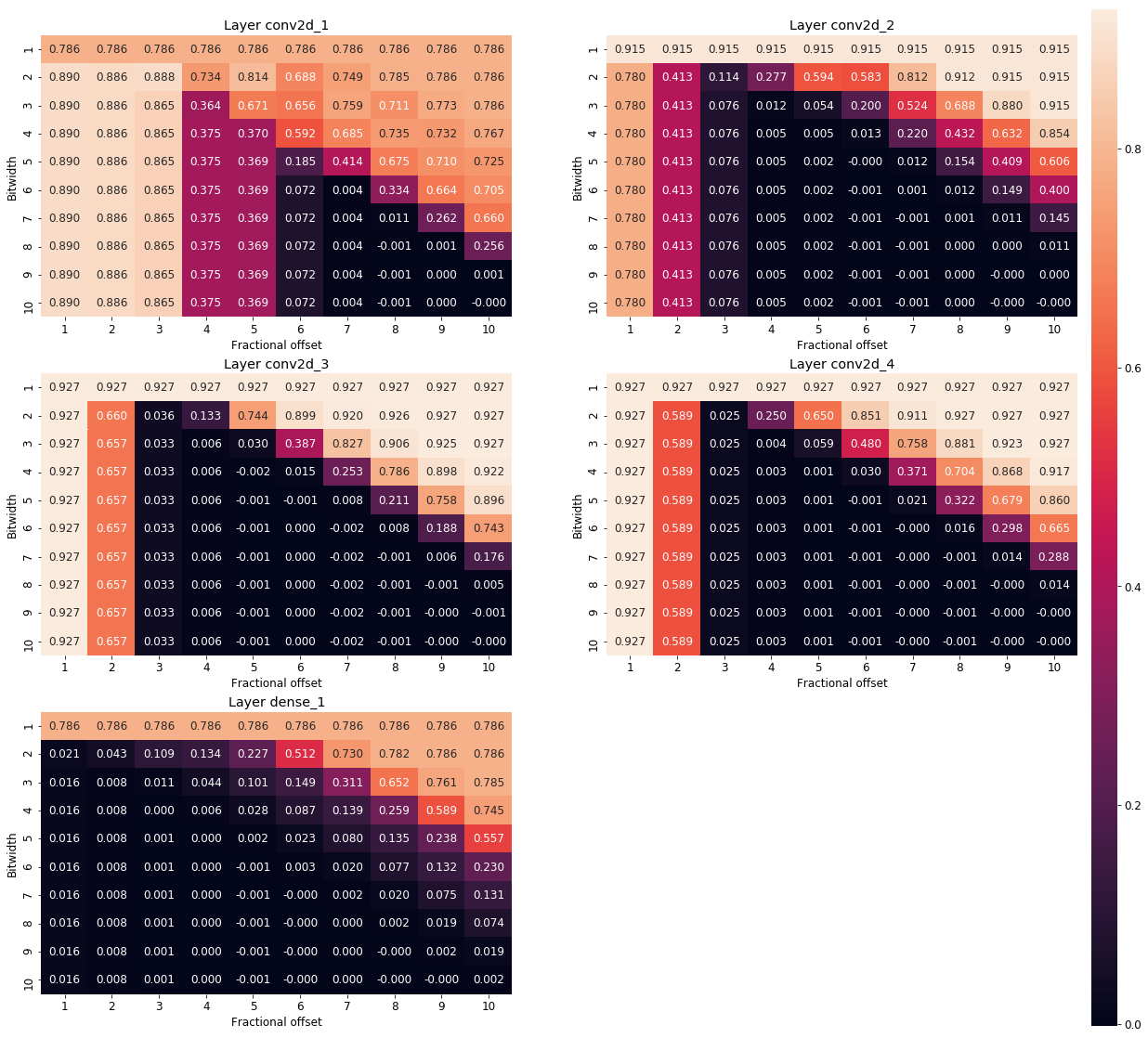}
    \caption{Brute force analysis of bitwidth and fractional offset for quantization of weights of a 5 layer CNN model trained on SVHN}
    \label{fig:bf_svhn_w}
    \centering
    \includegraphics[width=0.7\textwidth]{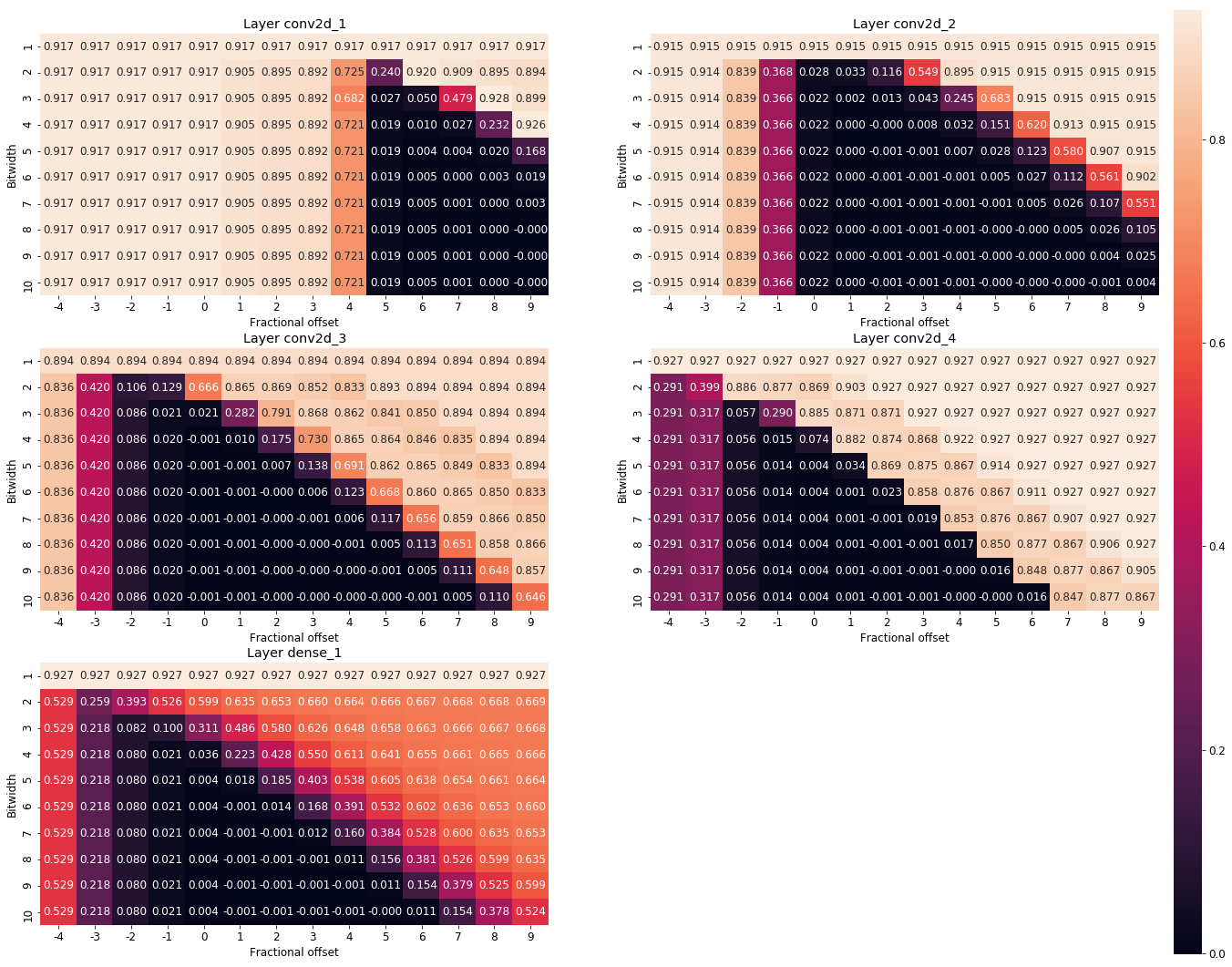}
    \caption{Brute force analysis of bitwidth and fractional offset for quantization of activations of a 5 layer CNN model trained on SVHN}
    \label{fig:bf_svhn_a}
\end{figure*}

\begin{figure*}
    \centering
    \includegraphics[width=0.7\textwidth]{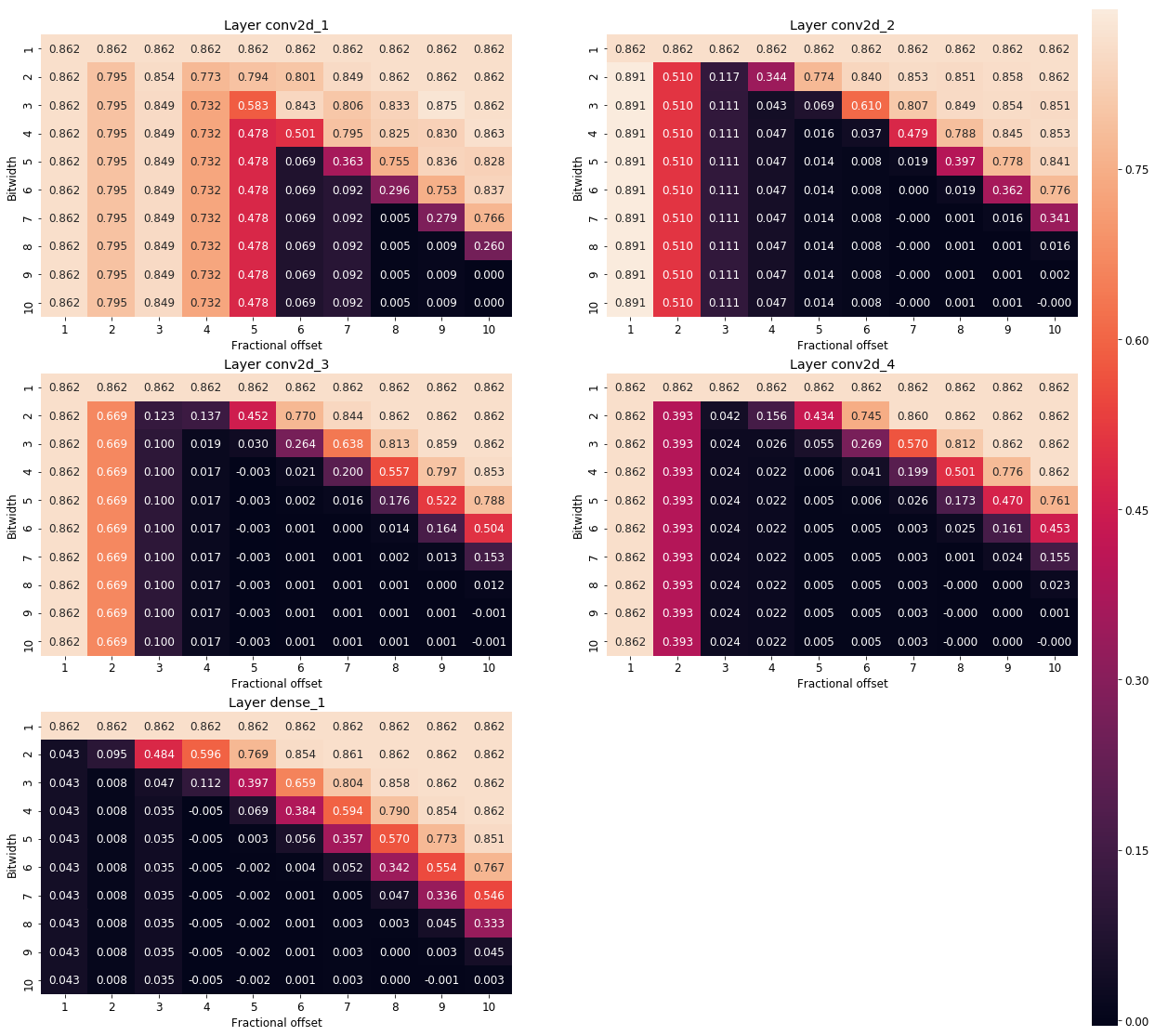}
    \caption{Brute force analysis of bitwidth and fractional offset for quantization of weights of a 5 layer CNN model trained on CIFAR10}
    \label{fig:bf_cifar10_w}
    \centering
    \includegraphics[width=0.7\textwidth]{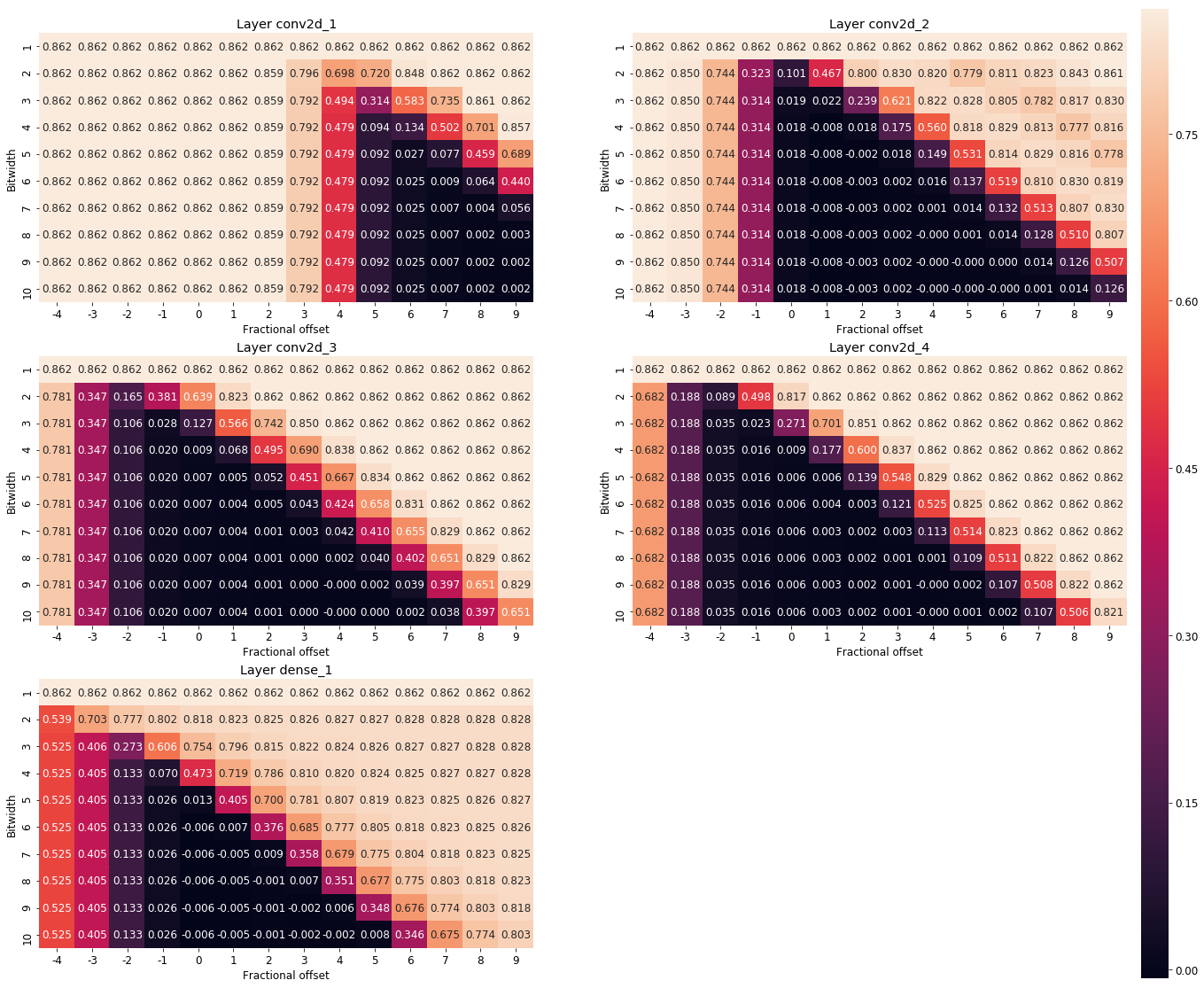}
    \caption{Brute force analysis of bitwidth and fractional offset for quantization of activations of a 5 layer CNN model trained on CIFAR10}
    \label{fig:bf_cifar10_a}
\end{figure*}

Fig.~\ref{fig:bf_mnist_w} - Fig.~\ref{fig:bf_cifar10_a} present the additional results from Brute Force analysis on CNNs. While we ran Brute Force analysis on all the models and datasets we had available for experimentation, we only present a few here as the other results are similar.

\subsection{Additional results for independent quantization}
\begin{figure*}[!ht]
    \centering
     \begin{subfigure}[t]{0.43\textwidth}
         \centering
         \includegraphics[width=\textwidth]{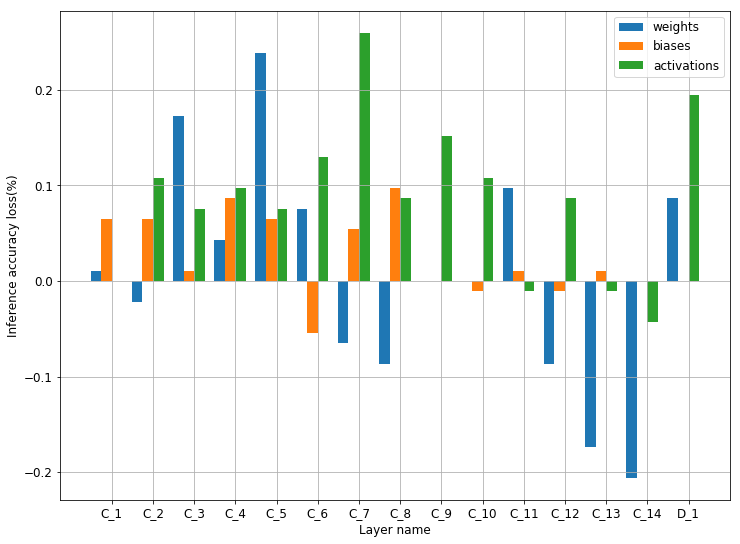}
         \caption{Inference accuracy loss $\Delta a^I_{l, p}$ for parameters of each layer after finding its optimal $(BW, F)_{l, p}$ using Independent Optimized Search while keeping other parameters at full precision.}
         \label{fig:ind_acc_loss_layer_keras_fashion}
     \end{subfigure}
    \hspace{20mm}
     \begin{subfigure}[t]{0.42\textwidth}
         \centering
         \includegraphics[width=\textwidth]{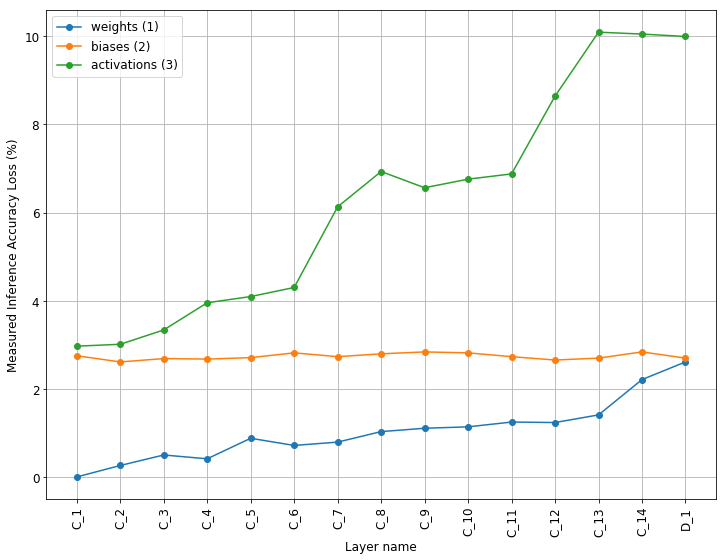}
         \caption{Inference accuracy loss of the CNN $\Delta a$ every time a parameter of a layer is quantized to the respective $(BW, F)^*_{l, p}$ found using Independent Optimized Search. Parameters of the network are quantized sequentially in the order of weights, biases followed by activations from layers 1 to L.}
         \label{fig:ind_seq_acc_loss_keras_fashion}
     \end{subfigure}
     \caption{Inference accuracy loss measured using two ways after quantizing parameters to their fixed-point representations of the pre-trained 15-layer Sequential CNN trained on MNIST using independent Optimized Search. Acceptable loss $\epsilon_{l, p} = 0.3\%$.}
     \label{fig:ind_quant_results_keras_fashion}
     
     \centering
     \begin{subfigure}[t]{0.43\textwidth}
         \centering
         \includegraphics[width=\textwidth]{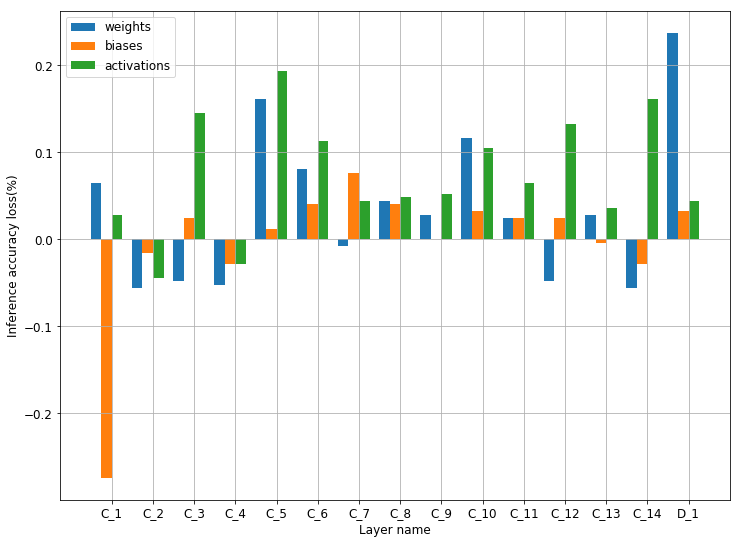}
         \caption{Inference accuracy loss $\Delta a^I_{l, p}$ for parameters of each layer after finding its optimal $(BW, F)_{l, p}$ using Independent Optimized Search while keeping other parameters at full precision.}
         \label{fig:ind_acc_loss_layer_keras_svhn}
     \end{subfigure}
    \hspace{20mm}
     \begin{subfigure}[t]{0.42\textwidth}
         \centering
         \includegraphics[width=\textwidth]{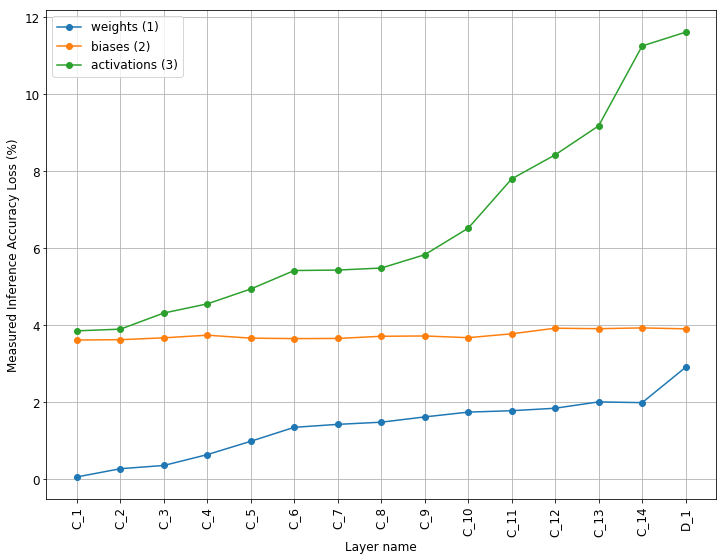}
         \caption{Inference accuracy loss of the CNN $\Delta a$ every time a parameter of a layer is quantized to the respective $(BW, F)^*_{l, p}$ found using Independent Optimized Search. Parameters of the network are quantized sequentially in the order of weights, biases followed by activations from layers 1 to L.}
         \label{fig:ind_seq_acc_loss_keras_svhn}
     \end{subfigure}
      
     \caption{Inference accuracy loss measured using two ways after quantizing parameters to their fixed-point representations of the pre-trained 15-layer Sequential CNN trained on SVHN using independent Optimized Search. Acceptable loss $\epsilon_{l, p} = 0.3\%$.}
     \label{fig:ind_quant_results_keras_svhn}
\end{figure*}

\begin{figure*}
     \centering
     \begin{subfigure}[t]{0.43\textwidth}
         \centering
         \includegraphics[width=\textwidth]{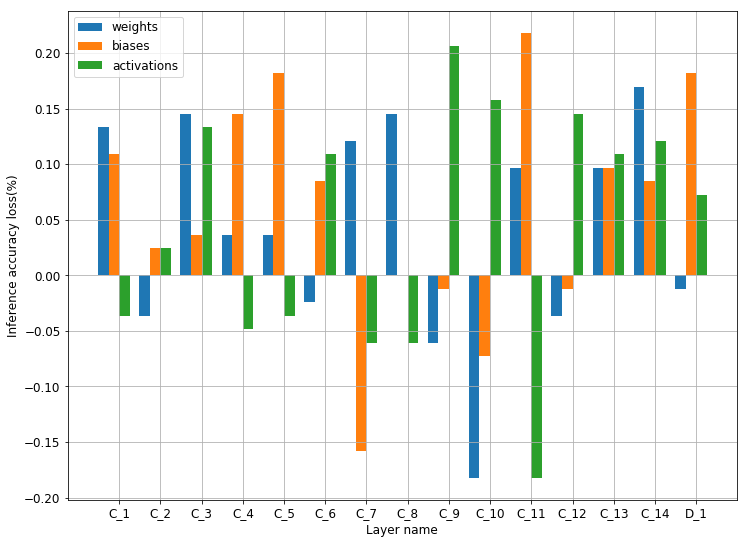}
         \caption{Inference accuracy loss $\Delta a^I_{l, p}$ for parameters of each layer after finding its optimal $(BW, F)_{l, p}$ using Independent Optimized Search while keeping other parameters at full precision.}
         \label{fig:ind_acc_loss_layer_keras_cifar10}
     \end{subfigure}
    \hspace{20mm}
     \begin{subfigure}[t]{0.42\textwidth}
         \centering
         \includegraphics[width=\textwidth]{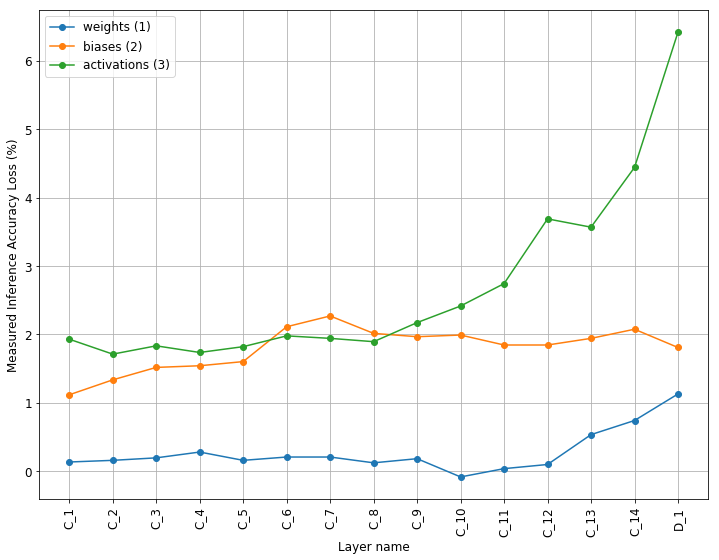}
         \caption{Inference accuracy loss of the CNN $\Delta a$ every time a parameter of a layer is quantized to the respective $(BW, F)^*_{l, p}$ found using Independent Optimized Search. Parameters of the network are quantized sequentially in the order of weights, biases followed by activations from layers 1 to L.}
         \label{fig:ind_seq_acc_loss_keras_cifar10}
     \end{subfigure}
     \caption{Inference accuracy loss measured using two ways after quantizing parameters to their fixed-point representations of the pre-trained 15-layer Sequential CNN trained on CIFAR10 using independent Optimized Search. Acceptable loss $\epsilon_{l, p} = 0.3\%$.}
     \label{fig:ind_quant_results_keras_cifar10}

    \centering
     \begin{subfigure}[t]{0.42\textwidth}
        \centering
        \includegraphics[width=\textwidth]{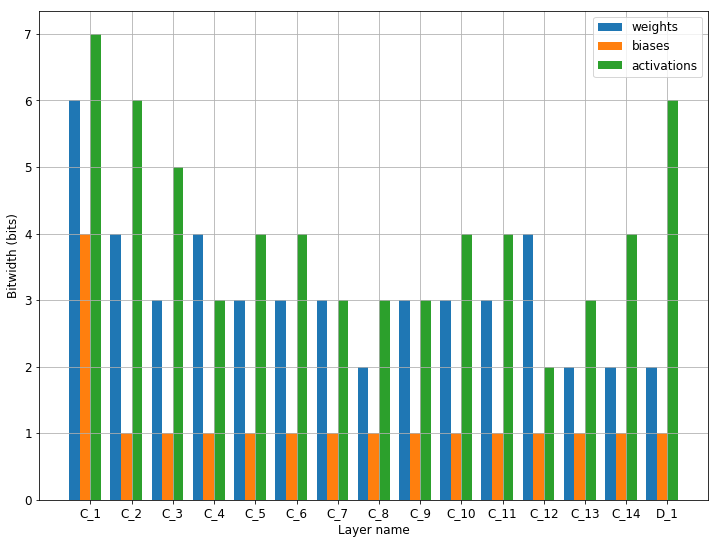}
        \caption{15-layer sequential - Fashion-MNIST}
        \label{fig:ind_bw_seq_fashion}
     \end{subfigure}
    \hspace{20mm}
    \begin{subfigure}[t]{0.42\textwidth}
        \centering
        \includegraphics[width=\textwidth]{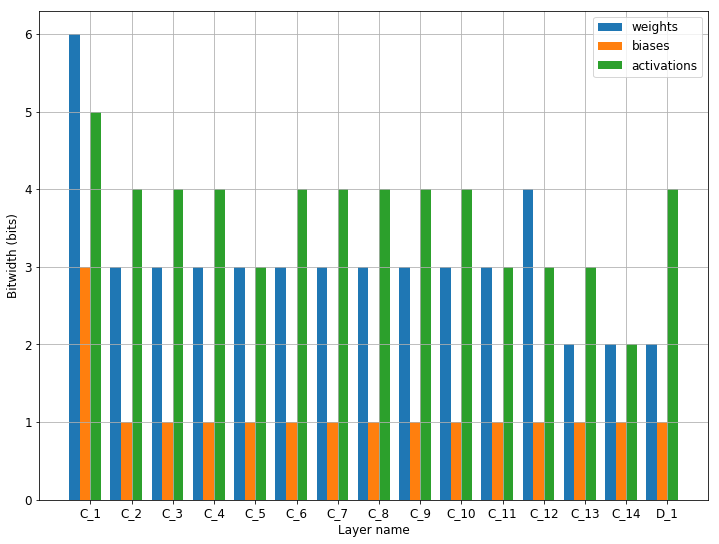}
        \caption{15-layer sequential - SVHN}
        \label{fig:ind_bw_seq_svhn}
     \end{subfigure}
     \hspace{20mm}
    \begin{subfigure}[t]{0.42\textwidth}
        \centering
        \includegraphics[width=\textwidth]{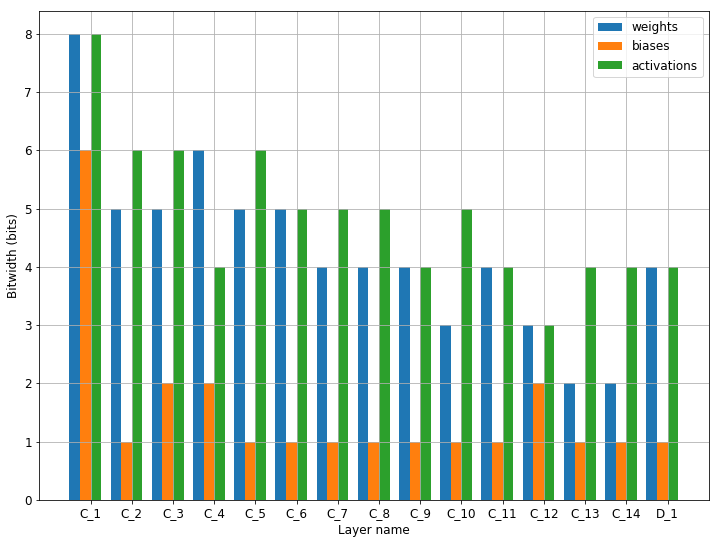}
        \caption{15-layer sequential - CIFAR10}
        \label{fig:ind_bw_seq_cifar10}
     \end{subfigure}
     \caption{Optimal bitwidths $BW^*_{l, p}$ of parameters of each layer of the pre-trained 15-layer Sequential CNN trained on CIFAR10 found using independent Optimized Search. Acceptable loss $\epsilon_{l, p} = 0.3\%$.}
     \label{fig:ind_quant_bw_all}
\end{figure*}

\begin{figure*}[t!]
    \centering
     \begin{subfigure}[t]{0.43\textwidth}
         \centering
         \includegraphics[width=\textwidth]{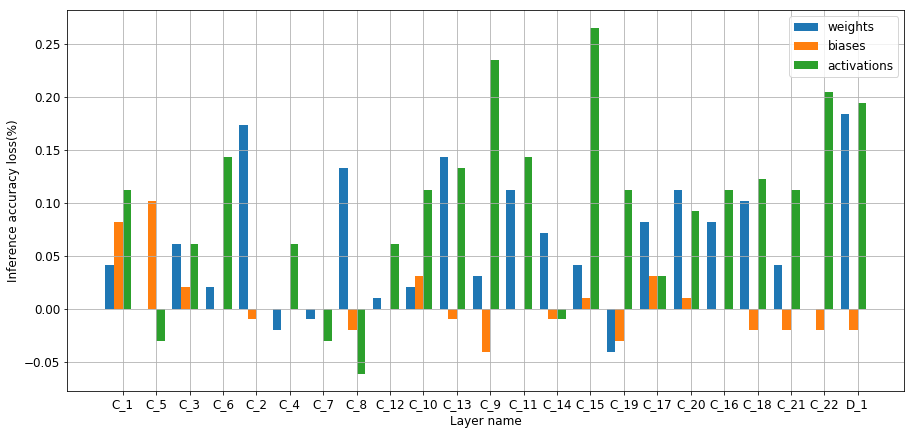}
         \caption{Inference accuracy loss $\Delta a^I_{l, p}$ for parameters of each layer after finding its optimal $(BW, F)_{l, p}$ using Independent Optimized Search while keeping other parameters at full precision.}
         \label{fig:ind_acc_loss_layer_df_mnist}
     \end{subfigure}
    \hspace{20mm}
     \begin{subfigure}[t]{0.42\textwidth}
         \centering
         \includegraphics[width=\textwidth]{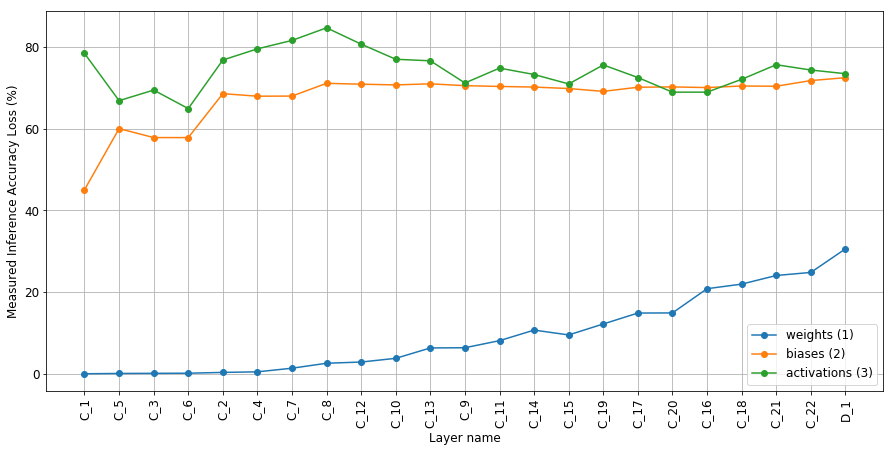}
         \caption{Inference accuracy loss of the CNN $\Delta a$ every time a parameter of a layer is quantized to the respective $(BW, F)^*_{l, p}$ found using Independent Optimized Search. Parameters of the network are quantized sequentially in the order of weights, biases followed by activations from layers 1 to L.}
         \label{fig:ind_seq_acc_loss_df_mnist}
     \end{subfigure}
      \begin{subfigure}[t]{0.42\textwidth}
        \centering
        \includegraphics[width=\textwidth]{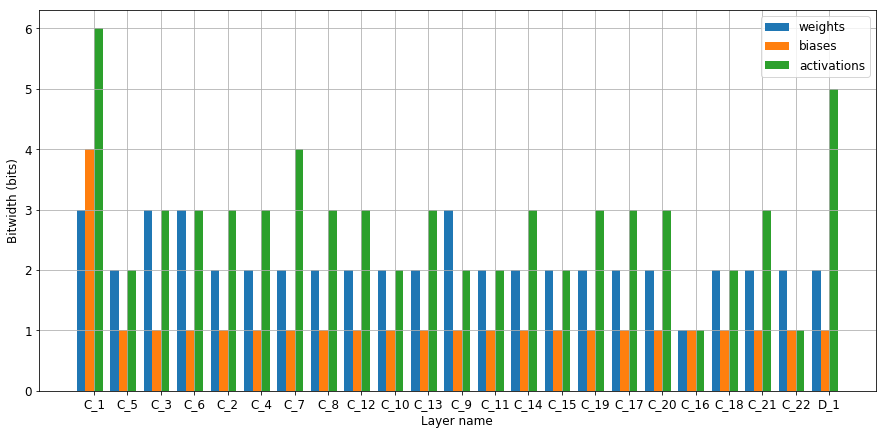}
        \caption{Optimal bitwidths $BW^*_{l, p}$ of parameters of each layer}
        \label{fig:ind_bw_bran_mnist}
     \end{subfigure}
     \caption{Inference accuracy loss measured using two ways after quantizing parameters to their fixed-point representations of the pre-trained 23-layer Branched CNN trained on MNIST using independent Optimized Search. Acceptable loss $\epsilon_{l, p} = 0.3\%$.}
     \label{fig:ind_quant_results_df_mnist}
     
     \centering
     \begin{subfigure}[t]{0.43\textwidth}
         \centering
         \includegraphics[width=\textwidth]{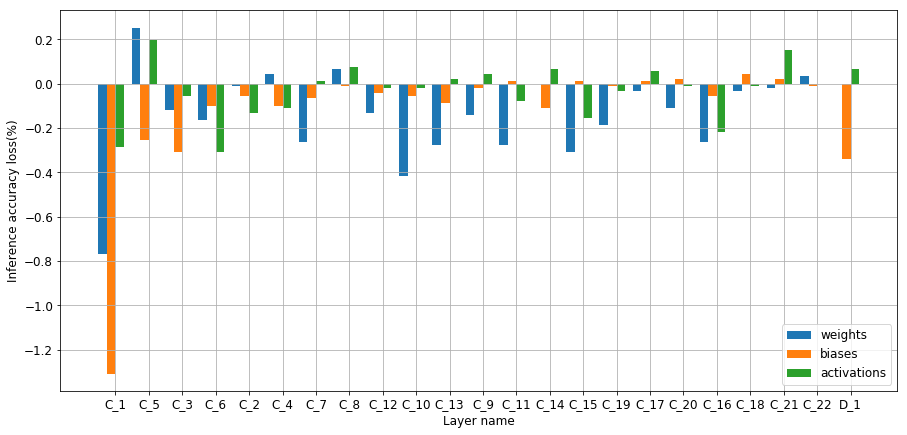}
         \caption{Inference accuracy loss $\Delta a^I_{l, p}$ for parameters of each layer after finding its optimal $(BW, F)_{l, p}$ using Independent Optimized Search while keeping other parameters at full precision.}
         \label{fig:ind_acc_loss_layer_df_fashion}
     \end{subfigure}
    \hspace{20mm}
     \begin{subfigure}[t]{0.42\textwidth}
         \centering
         \includegraphics[width=\textwidth]{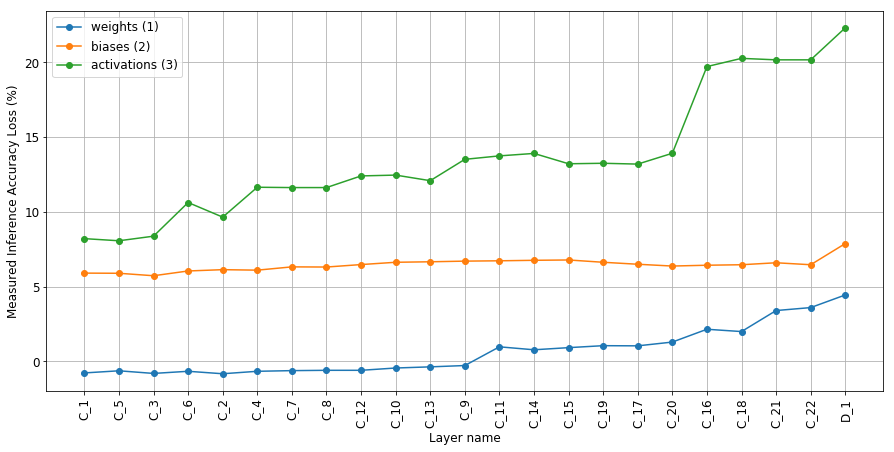}
         \caption{Inference accuracy loss of the CNN $\Delta a$ every time a parameter of a layer is quantized to the respective $(BW, F)^*_{l, p}$ found using Independent Optimized Search. Parameters of the network are quantized sequentially in the order of weights, biases followed by activations from layers 1 to L.}
         \label{fig:ind_seq_acc_loss_df_fashion}
     \end{subfigure}
      \begin{subfigure}[t]{0.42\textwidth}
        \centering
        \includegraphics[width=\textwidth]{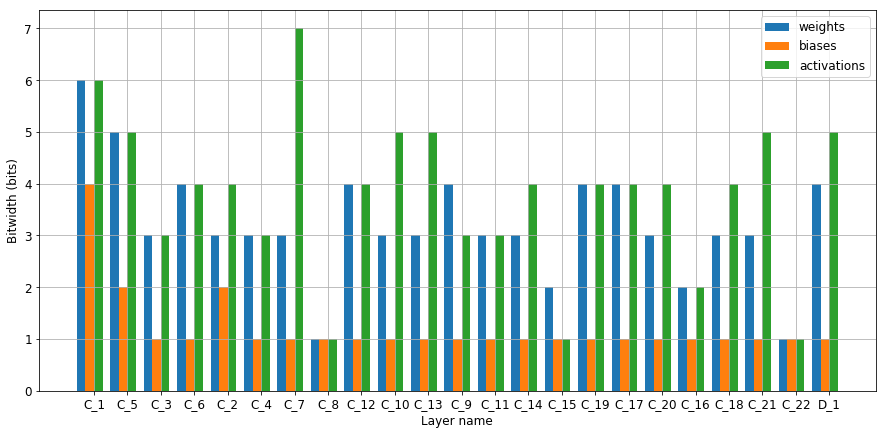}
        \caption{Optimal bitwidths $BW^*_{l, p}$ of parameters of each layer}
        \label{fig:ind_bw_bran_fashion}
     \end{subfigure}
     \caption{Inference accuracy loss measured using two ways after quantizing parameters to their fixed-point representations of the pre-trained 23-layer Branched CNN trained on Fashion-MNIST using independent Optimized Search. Acceptable loss $\epsilon_{l, p} = 0.3\%$.}
     \label{fig:ind_quant_results_df_fashion}
\end{figure*}

\begin{figure*}[t!]
    \centering
     \begin{subfigure}[t]{0.43\textwidth}
         \centering
         \includegraphics[width=\textwidth]{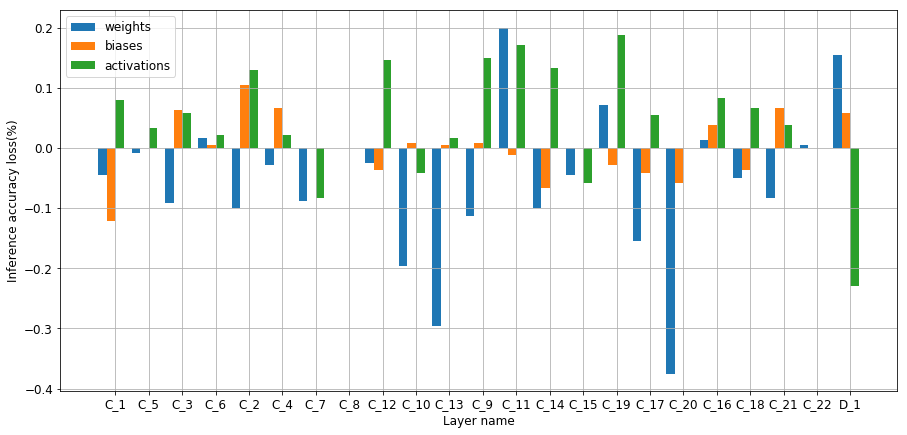}
         \caption{Inference accuracy loss $\Delta a^I_{l, p}$ for parameters of each layer after finding its optimal $(BW, F)_{l, p}$ using Independent Optimized Search while keeping other parameters at full precision.}
         \label{fig:ind_acc_loss_layer_df_svhn}
     \end{subfigure}
    \hspace{20mm}
     \begin{subfigure}[t]{0.42\textwidth}
         \centering
         \includegraphics[width=\textwidth]{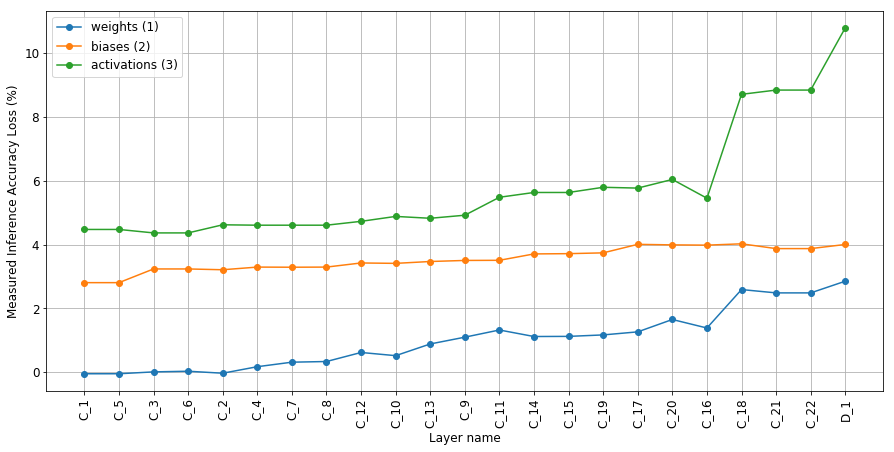}
         \caption{Inference accuracy loss of the CNN $\Delta a$ every time a parameter of a layer is quantized to the respective $(BW, F)^*_{l, p}$ found using Independent Optimized Search. Parameters of the network are quantized sequentially in the order of weights, biases followed by activations from layers 1 to L.}
         \label{fig:ind_seq_acc_loss_df_svhn}
     \end{subfigure}
      \begin{subfigure}[t]{0.42\textwidth}
        \centering
        \includegraphics[width=\textwidth]{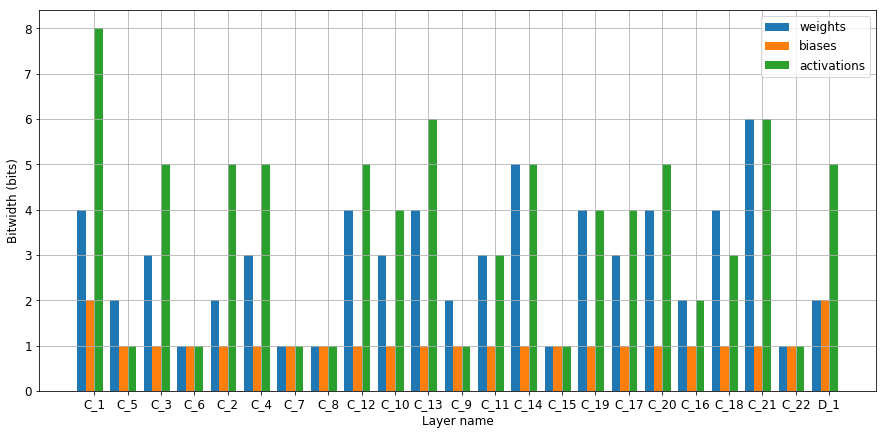}
        \caption{Optimal bitwidths $BW^*_{l, p}$ of parameters of each layer}
        \label{fig:ind_bw_bran_svhn}
     \end{subfigure}
     \caption{Inference accuracy loss measured using two ways after quantizing parameters to their fixed-point representations of the pre-trained 23-layer Branched CNN trained on SVHN using independent Optimized Search. Acceptable loss $\epsilon_{l, p} = 0.3\%$.}
     \label{fig:ind_quant_results_df_svhn}
     
     \centering
     \begin{subfigure}[t]{0.43\textwidth}
         \centering
         \includegraphics[width=\textwidth]{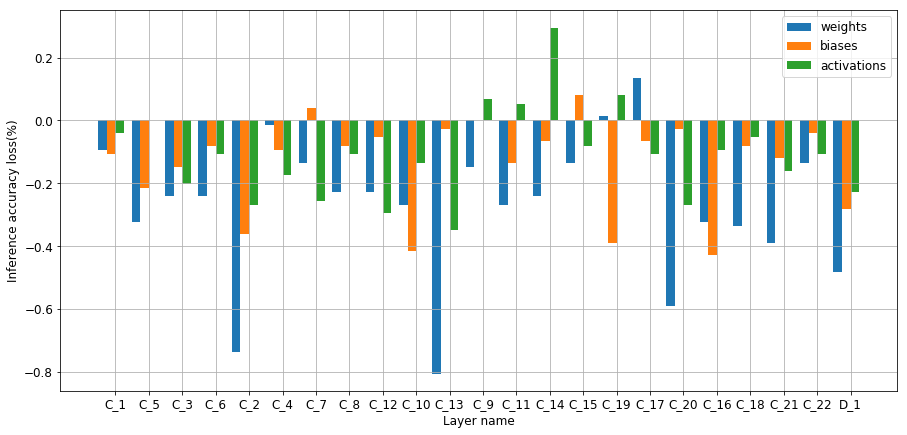}
         \caption{Inference accuracy loss $\Delta a^I_{l, p}$ for parameters of each layer after finding its optimal $(BW, F)_{l, p}$ using Independent Optimized Search while keeping other parameters at full precision.}
         \label{fig:ind_acc_loss_layer_df_cifar10}
     \end{subfigure}
    \hspace{20mm}
     \begin{subfigure}[t]{0.42\textwidth}
         \centering
         \includegraphics[width=\textwidth]{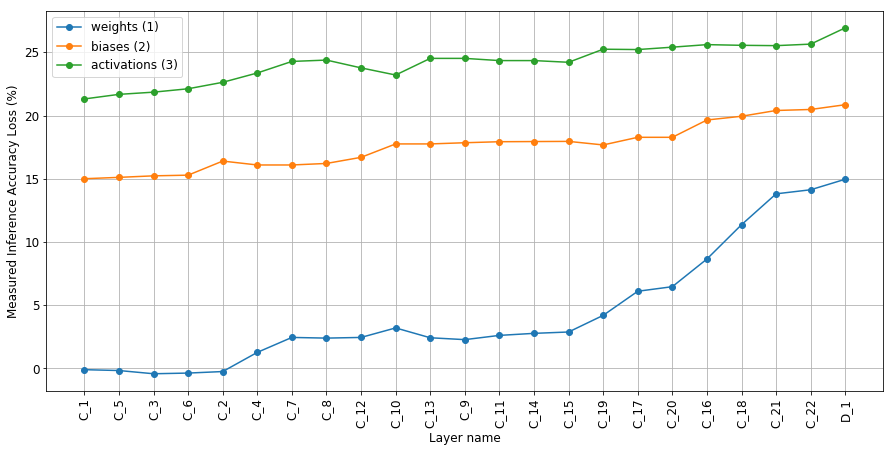}
         \caption{Inference accuracy loss of the CNN $\Delta a$ every time a parameter of a layer is quantized to the respective $(BW, F)^*_{l, p}$ found using Independent Optimized Search. Parameters of the network are quantized sequentially in the order of weights, biases followed by activations from layers 1 to L.}
         \label{fig:ind_seq_acc_loss_df_cifar10}
     \end{subfigure}
      \begin{subfigure}[t]{0.42\textwidth}
        \centering
        \includegraphics[width=\textwidth]{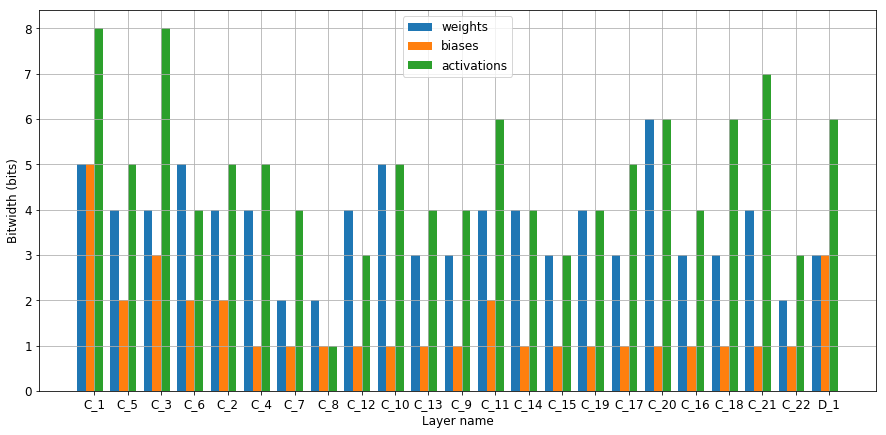}
        \caption{Optimal bitwidths $BW^*_{l, p}$ of parameters of each layer}
        \label{fig:ind_bw_bran_cifar10}
     \end{subfigure}
     \caption{Inference accuracy loss measured using two ways after quantizing parameters to their fixed-point representations of the pre-trained 23-layer Branched CNN trained on CIFAR10 using independent Optimized Search. Acceptable loss $\epsilon_{l, p} = 0.3\%$.}
     \label{fig:ind_quant_results_df_cifar10}
\end{figure*}

Fig.~\ref{fig:ind_quant_results_keras_fashion} - Fig.~\ref{fig:ind_quant_results_df_cifar10} presents additional results for independent quantization. Work of Section \ref{sec:ind_opts}.

\subsection{Additional results for dependent quantization}
\begin{figure*}
    \centering 
    \begin{subfigure}[t]{0.42\textwidth}
    \centering
    \includegraphics[width=\textwidth]{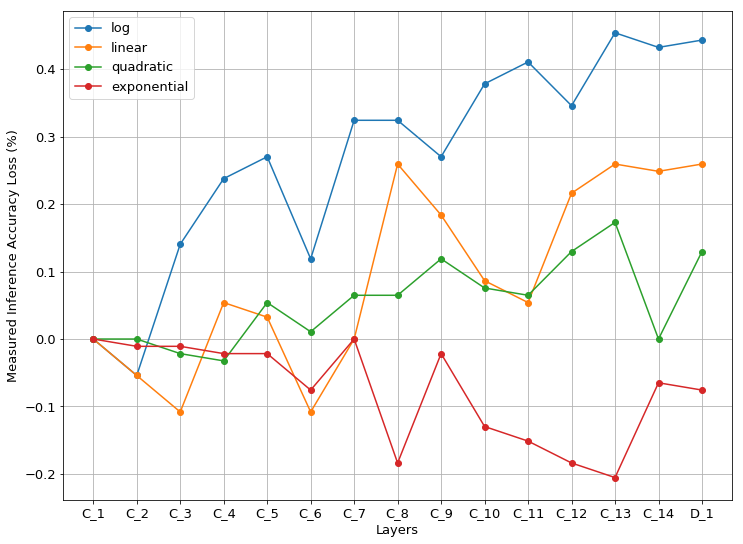}
    \caption{15 layer Sequential Fashion-MNIST Weights}
    \label{fig:dep_acc_loss_fashion}
    \end{subfigure}
    \hspace{10mm}
    \begin{subfigure}[t]{0.42\textwidth}
    \centering
    \includegraphics[width=\textwidth]{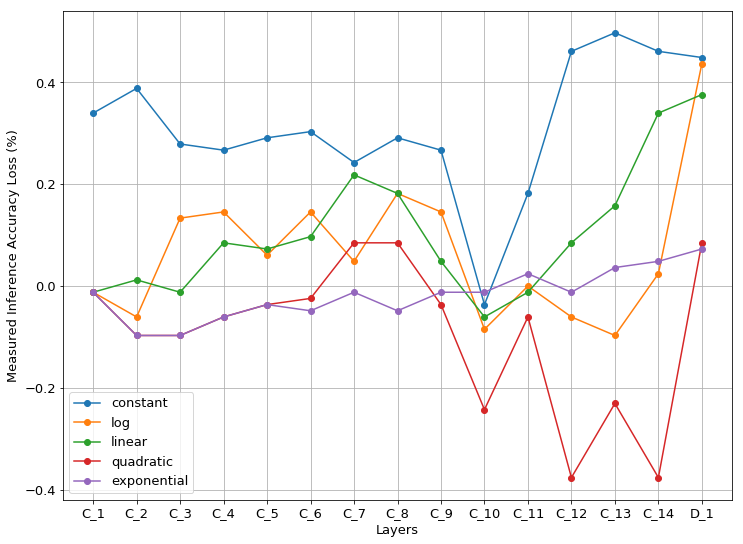}
    \caption{15 layer Sequential CIFAR10 Weights}
    \label{fig:dep_acc_loss_cifar10}
    \end{subfigure}
    \hspace{10mm}
    \begin{subfigure}[t]{0.42\textwidth}
    \centering
    \includegraphics[width=\textwidth]{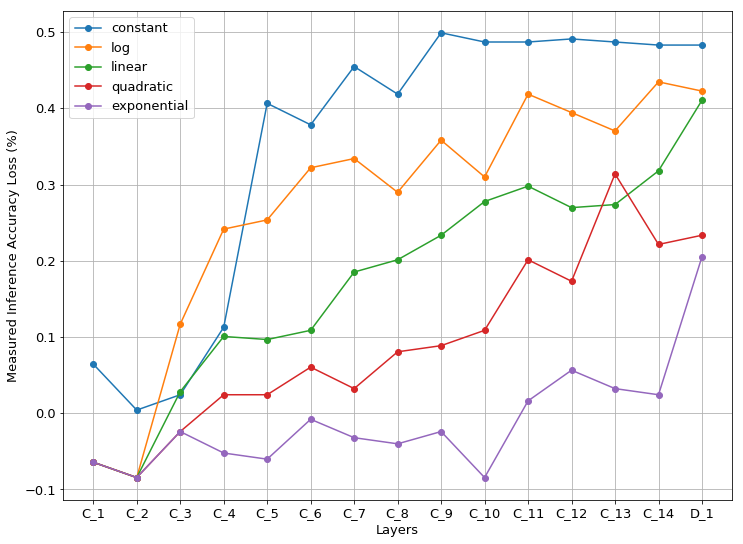}
    \caption{15 layer Sequential SVHN Weights}
    \label{fig:dep_acc_loss_svhn}
    \end{subfigure}
    \caption{Inference accuracy loss $\Delta a^D_{l, W}$ after quantizing Weights of all layers to their optimal $(BW, F)^*_{l, W}$ found using Dependent Optimized Search for different acceptable loss allocation schemes. Weights of the preceding layer are also quantized to $(BW, F)^*_{k, W}$ for $k < l$.}
    \label{fig:dep_acc_loss_w}
    
    \centering 
    \begin{subfigure}[t]{0.44\textwidth}
    \centering
    \includegraphics[width=\textwidth]{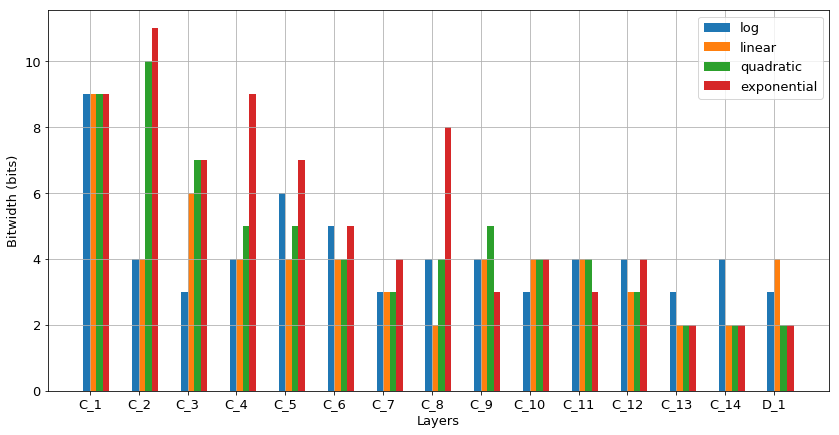}
    \caption{15 layer Sequential Fashion-MNIST Weights}
    \label{fig:dep_bw_fashion}
    \end{subfigure}
    \hspace{10mm}
    \begin{subfigure}[t]{0.44\textwidth}
    \centering
    \includegraphics[width=\textwidth]{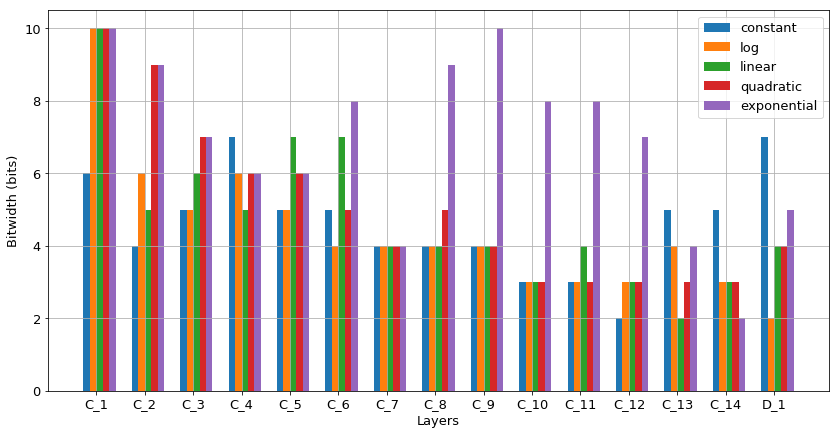}
    \caption{15 layer Sequential CIFAR10 Weights}
    \label{fig:dep_bw_cifar10}
    \end{subfigure}
    \hspace{10mm}
    \begin{subfigure}[t]{0.45\textwidth}
    \centering
    \includegraphics[width=\textwidth]{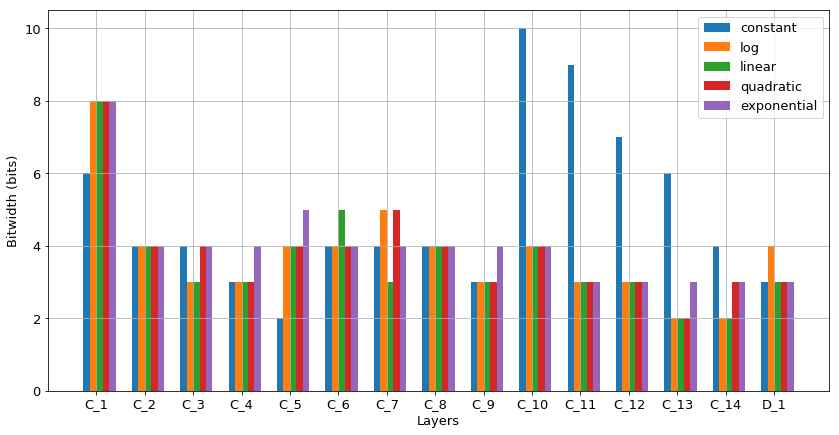}
    \caption{15 layer Sequential SVHN Weights}
    \label{fig:dep_bw_svhn}
    \end{subfigure}
    \caption{Optimal bitwidths $BW^*_{l, W}$ of Weights of each layer found using Dependent Optimized Search for different acceptable loss allocation schemes on a 15 layer Sequential CNN for the other three datasets. Weights of the preceding layer are also quantized to $(BW, F)^*_{k, W}$ for $k < l$.}
    \label{fig:dep_bw_w}
\end{figure*}

\begin{figure*}
    \centering 
    \begin{subfigure}[t]{0.42\textwidth}
    \centering
    \includegraphics[width=\textwidth]{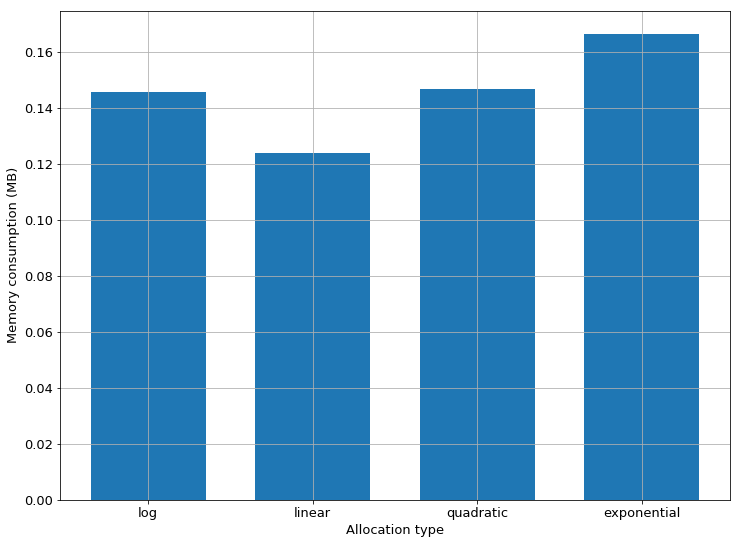}
    \caption{15 layer Sequential Fashion-MNIST Weights}
    \label{fig:dep_mem_fashion}
    \end{subfigure}
    \hspace{10mm}
    \begin{subfigure}[t]{0.42\textwidth}
    \centering
    \includegraphics[width=\textwidth]{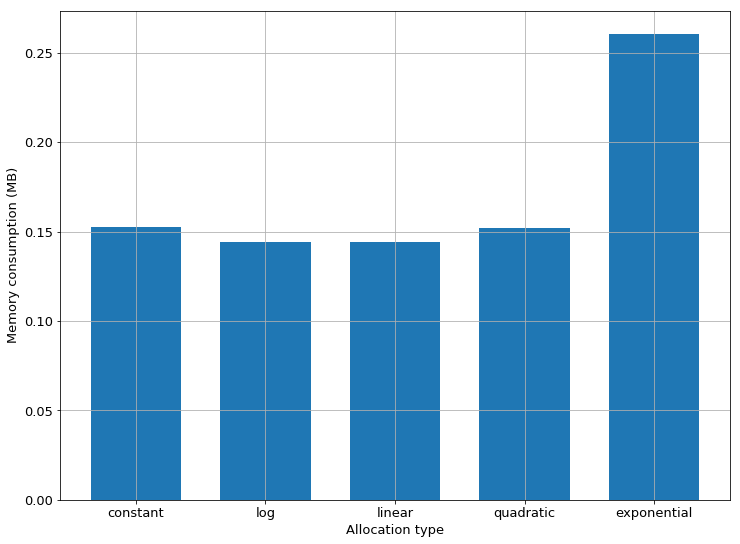}
    \caption{15 layer Sequential CIFAR10 Weights}
    \label{fig:dep_mem_cifar10}
    \end{subfigure}
    \hspace{10mm}
    \begin{subfigure}[t]{0.42\textwidth}
    \centering
    \includegraphics[width=\textwidth]{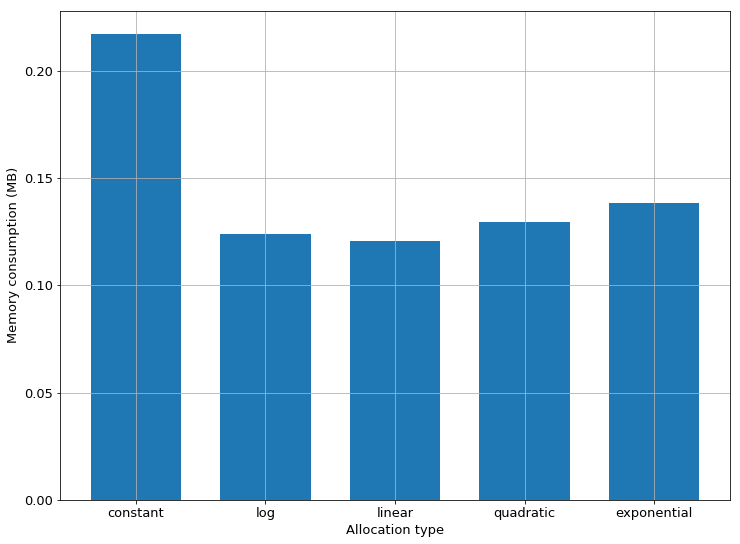}
    \caption{15 layer Sequential SVHN Weights}
    \label{fig:dep_mem_svhn}
    \end{subfigure}
    \caption{Total memory consumption of the quantized weights with optimal bitwidths $BW^*_{l, W}$ in Fig.~\ref{fig:dep_bw_w}.}
    \label{fig:dep_mem_w}
\end{figure*}

\begin{figure*}
    \centering 
    \begin{subfigure}[t]{0.43\textwidth}
    \centering
    \includegraphics[width=\textwidth]{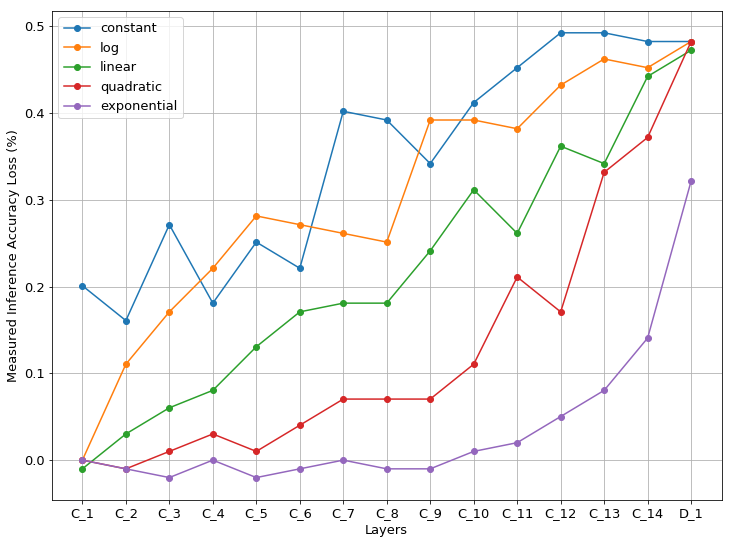}
    \caption{15 layer Sequential MNIST Activations}
    \label{fig:dep_acc_loss_mnist_a}
    \end{subfigure}
    \hspace{10mm}
    \begin{subfigure}[t]{0.43\textwidth}
    \centering
    \includegraphics[width=\textwidth]{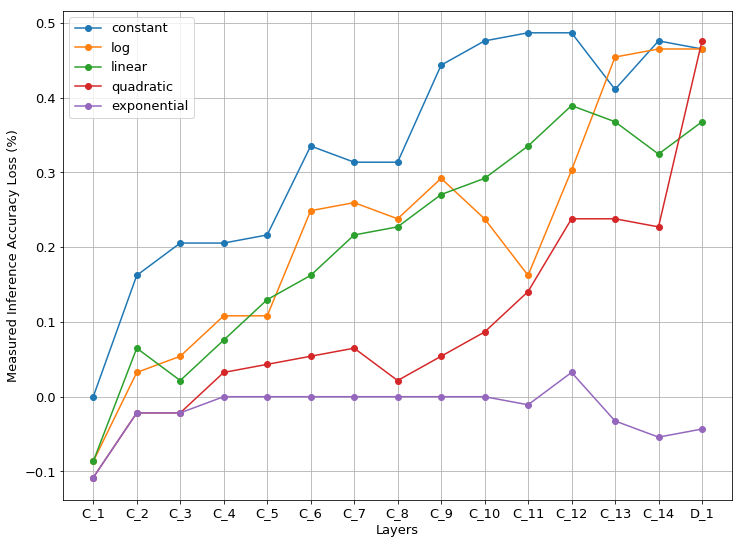}
    \caption{15 layer Sequential Fashion-MNIST Activations}
    \label{fig:dep_acc_loss_fashion_a}
    \end{subfigure}
    \hspace{10mm}
    \begin{subfigure}[t]{0.43\textwidth}
    \centering
    \includegraphics[width=\textwidth]{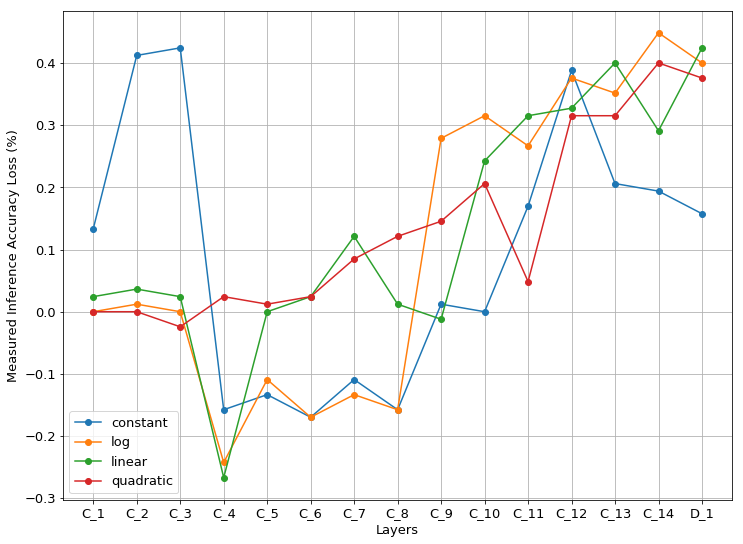}
    \caption{15 layer Sequential CIFAR10 Activations}
    \label{fig:dep_acc_loss_cifar10_a}
    \end{subfigure}
    \hspace{10mm}
    \begin{subfigure}[t]{0.43\textwidth}
    \centering
    \includegraphics[width=\textwidth]{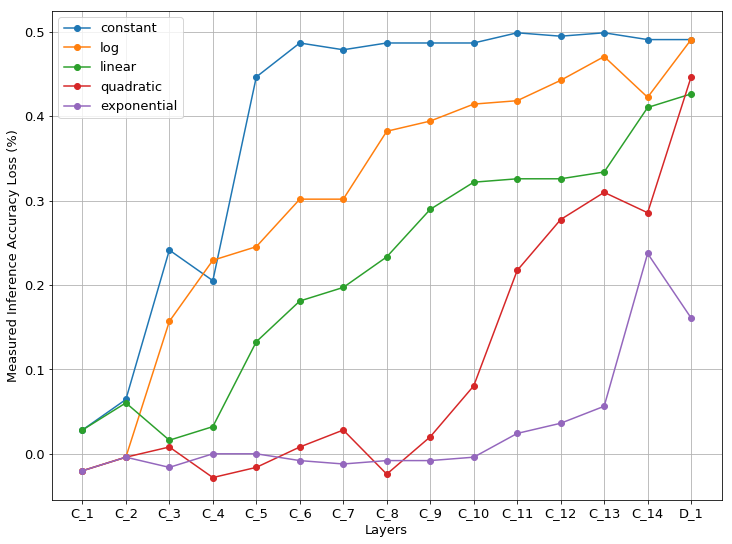}
    \caption{15 layer Sequential SVHN Activations}
    \label{fig:dep_acc_loss_svhn_a}
    \end{subfigure}
    \caption{Inference accuracy loss $\Delta a^D_{l, A}$ after quantizing activations of all layers to their optimal $(BW, F)^*_{l, A}$ found using Dependent Optimized Search for different acceptable loss allocation schemes. Activations of the preceding layer are also quantized to $(BW, F)^*_{k, A}$ for $k < l$.}
    \label{fig:dep_acc_loss_a}
    
\end{figure*}

\begin{figure*}
    \centering 
    \begin{subfigure}[t]{0.46\textwidth}
    \centering
    \includegraphics[width=\textwidth]{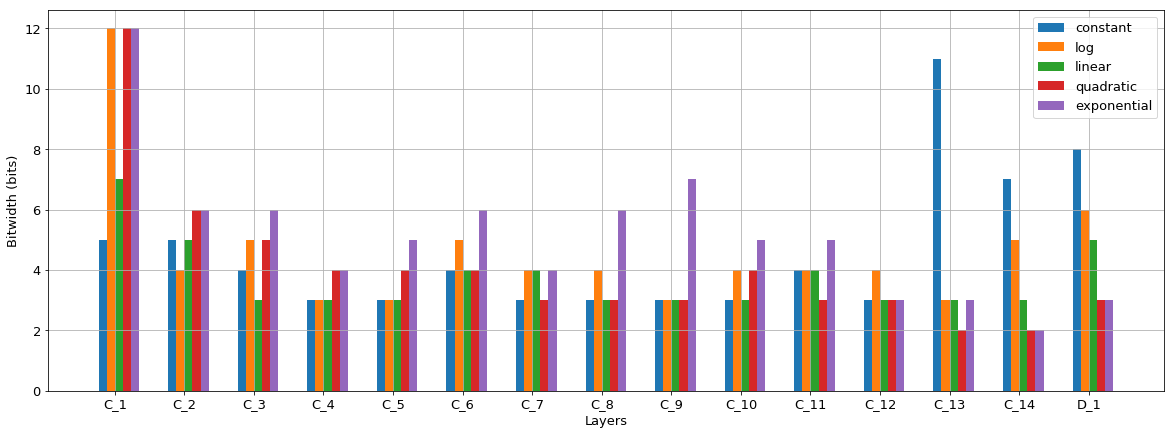}
    \caption{15 layer Sequential MNIST Activations}
    \label{fig:dep_bw_mnist_a}
    \end{subfigure}
    \hspace{7mm}
    \begin{subfigure}[t]{0.46\textwidth}
    \centering
    \includegraphics[width=\textwidth]{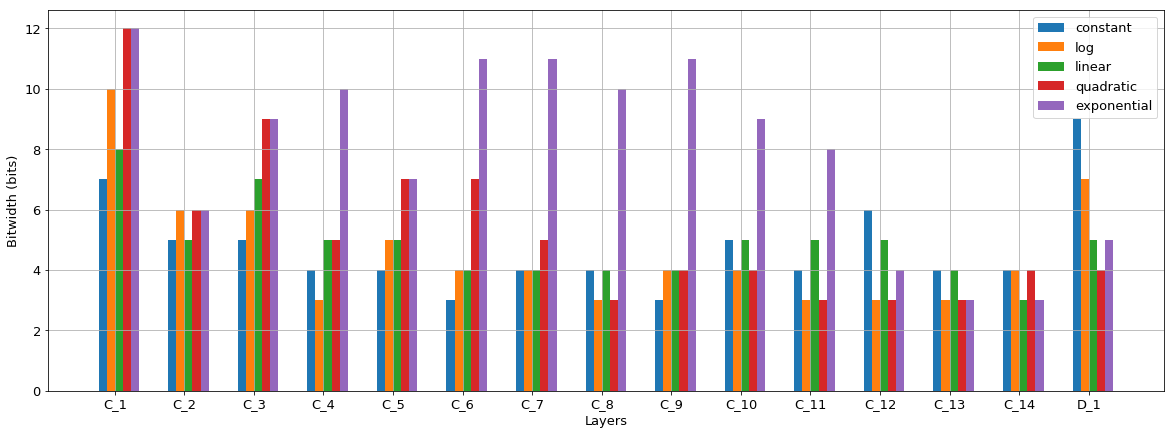}
    \caption{15 layer Sequential Fashion-MNIST Activations}
    \label{fig:dep_bw_fashion_a}
    \end{subfigure}
    \hspace{7mm}
    \begin{subfigure}[t]{0.46\textwidth}
    \centering
    \includegraphics[width=\textwidth]{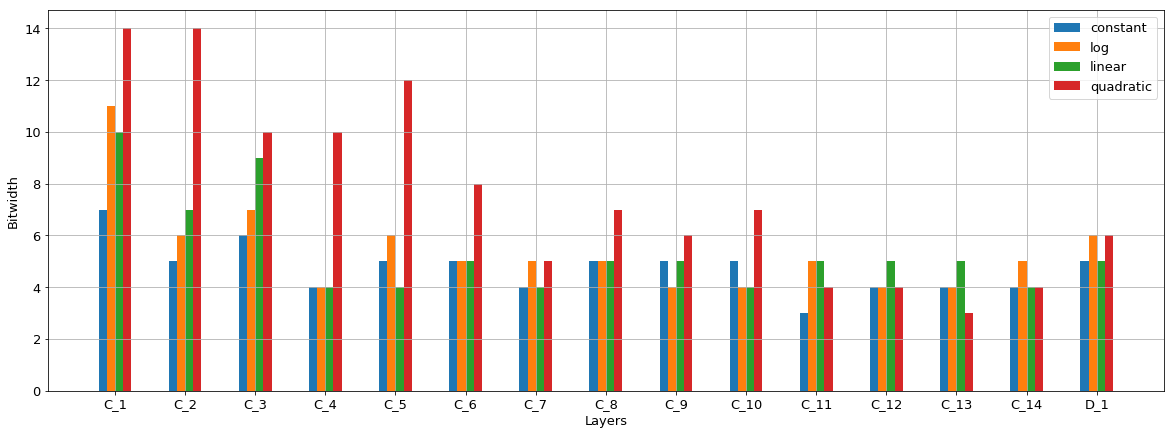}
    \caption{15 layer Sequential CIFAR10 Activations}
    \label{fig:dep_bw_cifar10_a}
    \end{subfigure}
    \hspace{7mm}
    \begin{subfigure}[t]{0.46\textwidth}
    \centering
    \includegraphics[width=\textwidth]{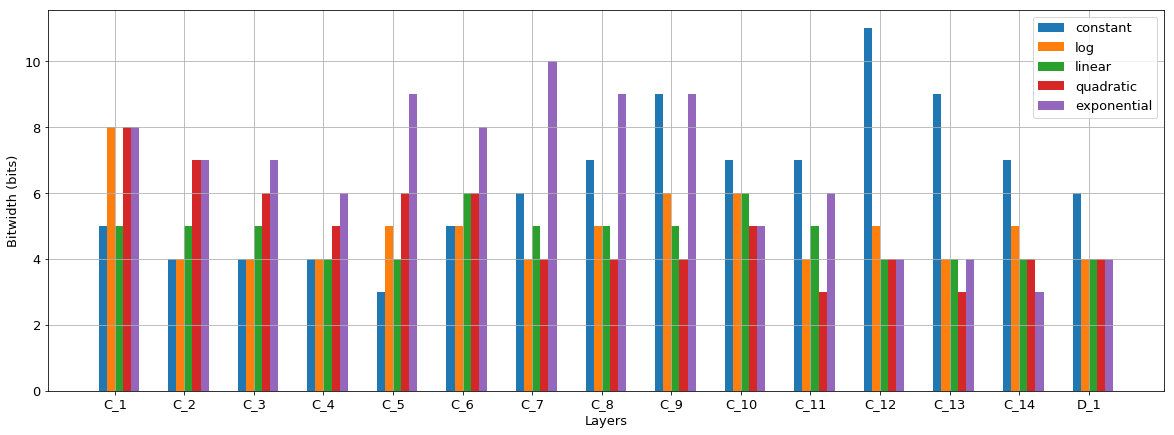}
    \caption{15 layer Sequential SVHN Activations}
    \label{fig:dep_bw_svhn_a}
    \end{subfigure}
    \caption{Optimal bitwidths $BW^*_{l, A}$ of Activations of each layer found using Dependent Optimized Search for different acceptable loss allocation schemes on a 15 layer Sequential CNN for the other three datasets. Activations of the preceding layer are also quantized to $(BW, F)^*_{k, A}$ for $k < l$.}
    \label{fig:dep_bw_a}
    
\end{figure*}

Fig.~\ref{fig:dep_acc_loss_w} - Fig.~\ref{fig:dep_bw_a} presents additional results for dependent quantization. Work of Section \ref{sec:dep_opts}.

\subsection{Additional results for the final method}

\begin{figure*}[t!]
    \centering
    \begin{subfigure}[t]{0.45\textwidth}
        \centering
        \includegraphics[width=\textwidth]{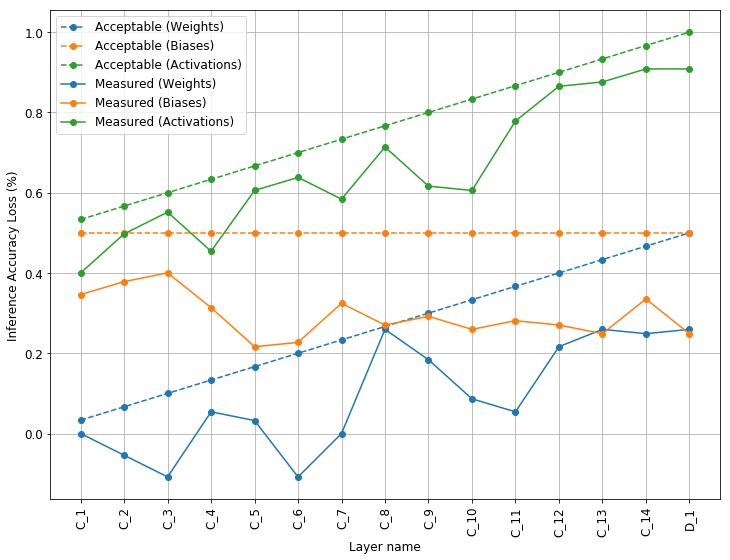}
        \caption{Acceptable and measured inference accuracy loss $\Delta a^D_{l, p}$ after sequentially quantizing the parameters $\mathbf{p}_l$ in the order ($\mathbf{W} \rightarrow \mathbf{B} \rightarrow \mathbf{A}$) from layers 1 to L.}
        \label{fig:opts_seq_fashion_acc_loss}
     \end{subfigure}
    \hspace{7mm}
    \begin{subfigure}[t]{0.45\textwidth}
        \centering
        \includegraphics[width=\textwidth]{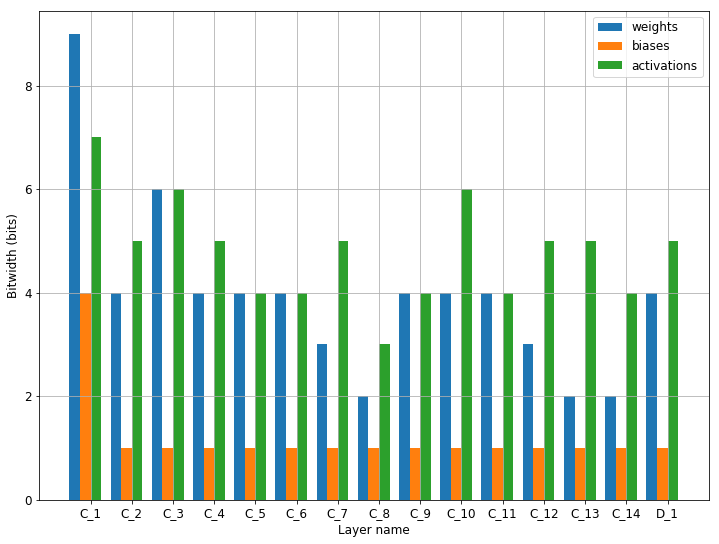}
        \caption{Optimal bitwidths $BW^*_{l, p}$ of parameters of each layer.}
        \label{fig:opts_seq_fashion_bw}
    \end{subfigure}
    \caption{Results for acceptable and measured inference accuracy loss and the optimal bitwidths of the quantized CNN resulting from dependent optimized search using a linear acceptable loss allocation scheme on a 15-layer sequential CNN trained on Fashion-MNIST.}
    \label{fig:opts_results_fashion_seq}
    
    \centering
    \begin{subfigure}[t]{0.45\textwidth}
        \centering
        \includegraphics[width=\textwidth]{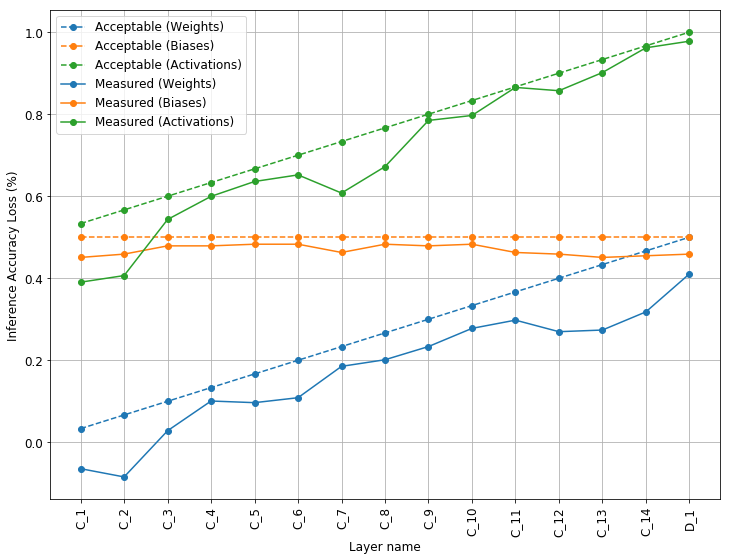}
        \caption{Acceptable and measured inference accuracy loss $\Delta a^D_{l, p}$ after sequentially quantizing the parameters $\mathbf{p}_l$ in the order ($\mathbf{W} \rightarrow \mathbf{B} \rightarrow \mathbf{A}$) from layers 1 to L.}
        \label{fig:opts_seq_svhn_acc_loss}
     \end{subfigure}
    \hspace{7mm}
    \begin{subfigure}[t]{0.45\textwidth}
        \centering
        \includegraphics[width=\textwidth]{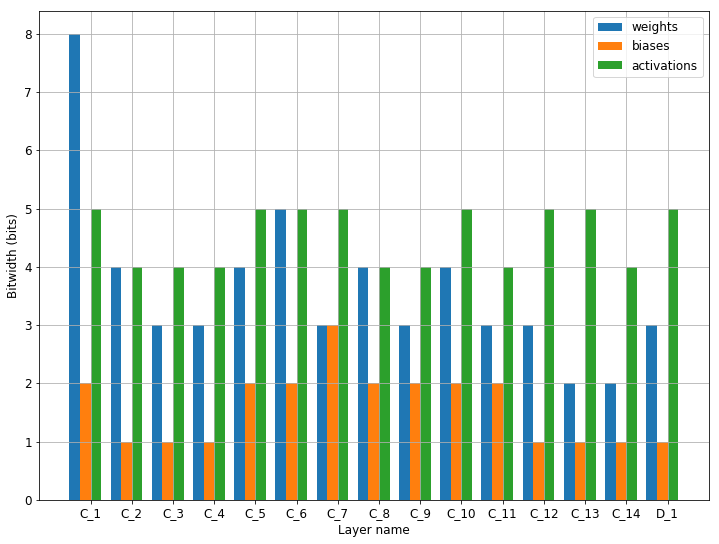}
        \caption{Optimal bitwidths $BW^*_{l, p}$ of parameters of each layer.}
        \label{fig:opts_seq_svhn_bw}
    \end{subfigure}
    \caption{Results for acceptable and measured inference accuracy loss and the optimal bitwidths of the quantized CNN resulting from dependent optimized search using a linear acceptable loss allocation scheme on a 15-layer sequential CNN trained on SVHN.}
    \label{fig:opts_results_svhn_seq}
    
\end{figure*}

\begin{figure*}
    \centering
    \begin{subfigure}[t]{0.45\textwidth}
        \centering
        \includegraphics[width=\textwidth]{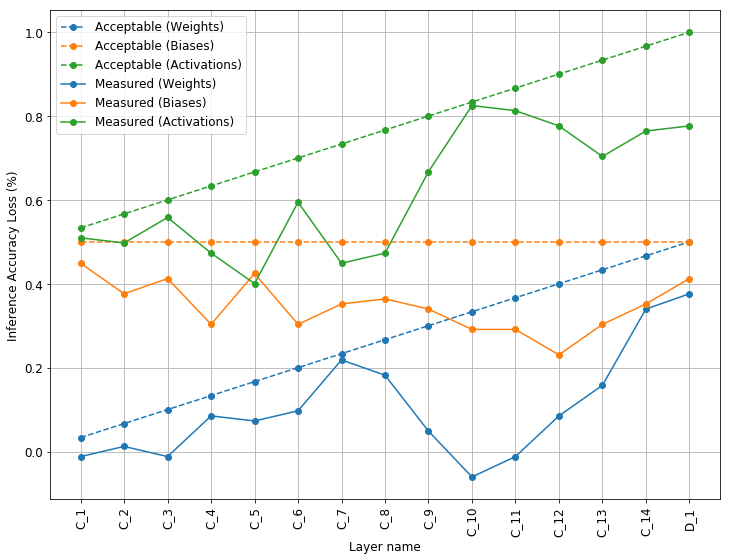}
        \caption{Acceptable and measured inference accuracy loss $\Delta a^D_{l, p}$ after sequentially quantizing the parameters $\mathbf{p}_l$ in the order ($\mathbf{W} \rightarrow \mathbf{B} \rightarrow \mathbf{A}$) from layers 1 to L.}
        \label{fig:opts_seq_cifar10_acc_loss}
     \end{subfigure}
    \hspace{7mm}
    \begin{subfigure}[t]{0.45\textwidth}
        \centering
        \includegraphics[width=\textwidth]{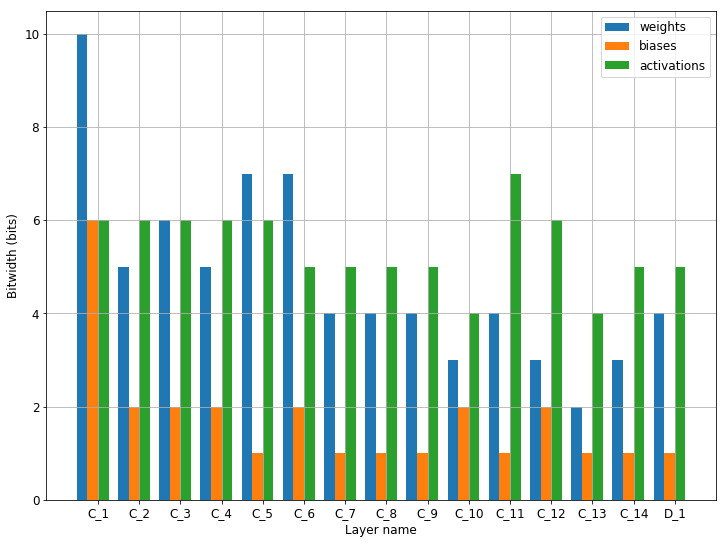}
        \caption{Optimal bitwidths $BW^*_{l, p}$ of parameters of each layer.}
        \label{fig:opts_seq_cifar10_bw}
    \end{subfigure}
    \caption{Results for acceptable and measured inference accuracy loss and the optimal bitwidths of the quantized CNN resulting from dependent optimized search using a linear acceptable loss allocation scheme on a 15-layer sequential CNN trained on CIFAR10.}
    \label{fig:opts_results_cifar10_seq}
    
    \centering
    \begin{subfigure}[t]{0.45\textwidth}
        \centering
        \includegraphics[width=\textwidth]{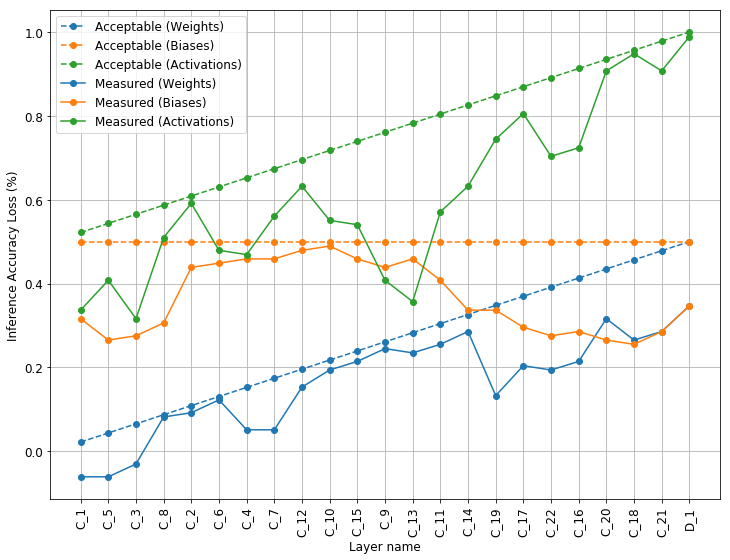}
        \caption{Acceptable and measured inference accuracy loss $\Delta a^D_{l, p}$ after sequentially quantizing the parameters $\mathbf{p}_l$ in the order ($\mathbf{W} \rightarrow \mathbf{B} \rightarrow \mathbf{A}$) from layers 1 to L.}
        \label{fig:opts_df_mnist_acc_loss}
     \end{subfigure}
    \hspace{4mm}
    \begin{subfigure}[t]{0.47\textwidth}
        \centering
        \includegraphics[width=\textwidth]{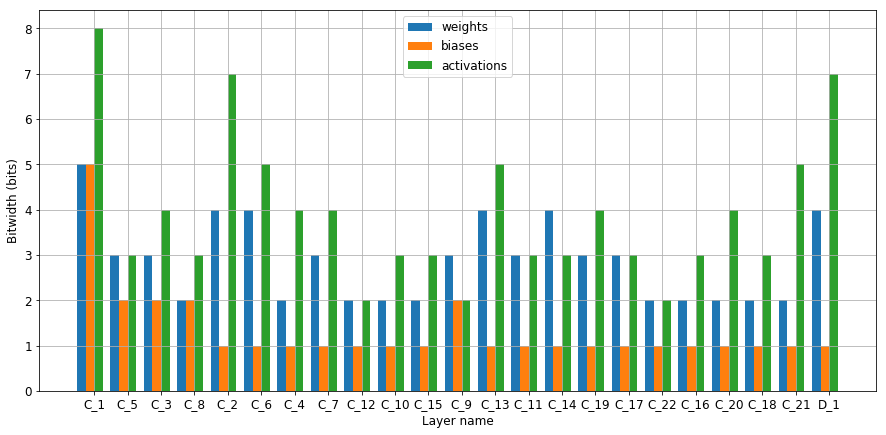}
        \caption{Optimal bitwidths $BW^*_{l, p}$ of parameters of each layer.}
        \label{fig:opts_df_mnist_bw}
    \end{subfigure}
    \caption{Results for acceptable and measured inference accuracy loss and the optimal bitwidths of the quantized CNN resulting from dependent optimized search using a linear acceptable loss allocation scheme on a 23-layer branched CNN trained on MNIST.}
    \label{fig:opts_results_mnist_df}
\end{figure*}

\begin{figure*}
    \centering
    \begin{subfigure}[t]{0.45\textwidth}
        \centering
        \includegraphics[width=\textwidth]{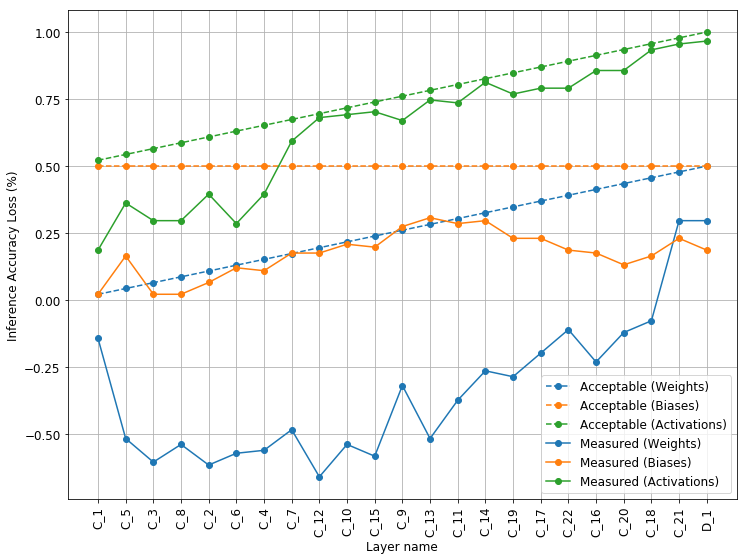}
        \caption{Acceptable and measured inference accuracy loss $\Delta a^D_{l, p}$ after sequentially quantizing the parameters $\mathbf{p}_l$ in the order ($\mathbf{W} \rightarrow \mathbf{B} \rightarrow \mathbf{A}$) from layers 1 to L.}
        \label{fig:opts_df_fashion_acc_loss}
     \end{subfigure}
    \hspace{4mm}
    \begin{subfigure}[t]{0.47\textwidth}
        \centering
        \includegraphics[width=\textwidth]{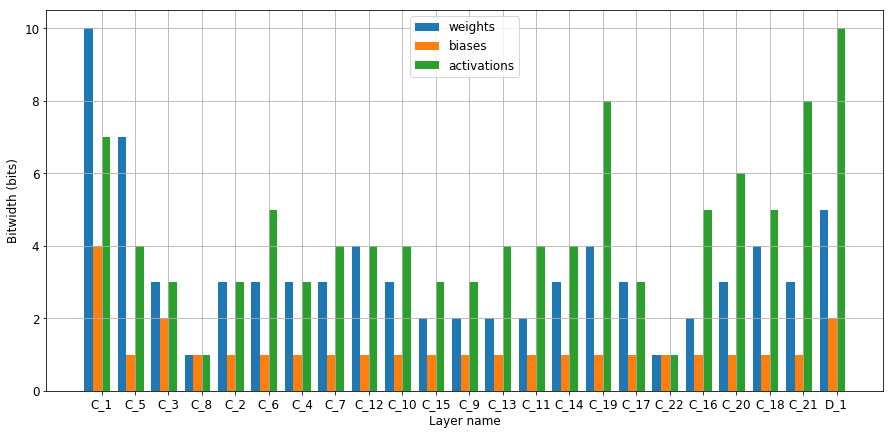}
        \caption{Optimal bitwidths $BW^*_{l, p}$ of parameters of each layer.}
        \label{fig:opts_df_fashion_bw}
    \end{subfigure}
    \caption{Results for acceptable and measured inference accuracy loss and the optimal bitwidths of the quantized CNN resulting from dependent optimized search using a linear acceptable loss allocation scheme on a 23-layer branched CNN trained on Fashion-MNIST.}
    \label{fig:opts_results_fashion_df}
    
    \centering
    \begin{subfigure}[t]{0.45\textwidth}
        \centering
        \includegraphics[width=\textwidth]{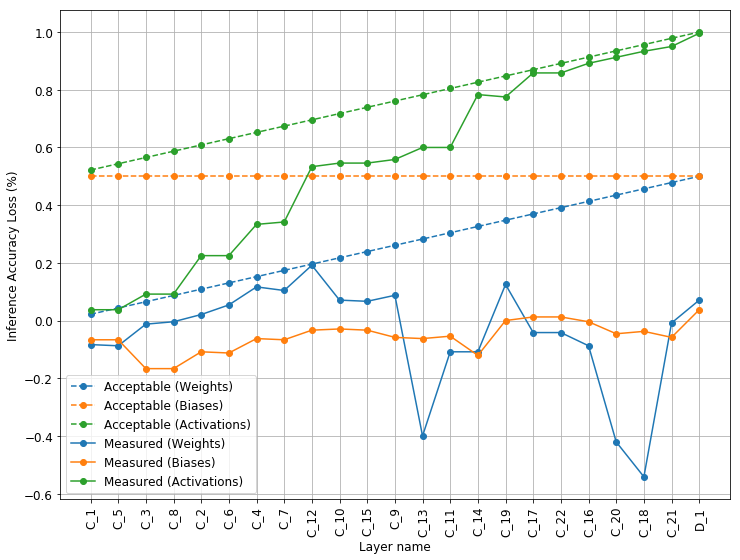}
        \caption{Acceptable and measured inference accuracy loss $\Delta a^D_{l, p}$ after sequentially quantizing the parameters $\mathbf{p}_l$ in the order ($\mathbf{W} \rightarrow \mathbf{B} \rightarrow \mathbf{A}$) from layers 1 to L.}
        \label{fig:opts_df_svhn_acc_loss}
     \end{subfigure}
    \hspace{4mm}
    \begin{subfigure}[t]{0.47\textwidth}
        \centering
        \includegraphics[width=\textwidth]{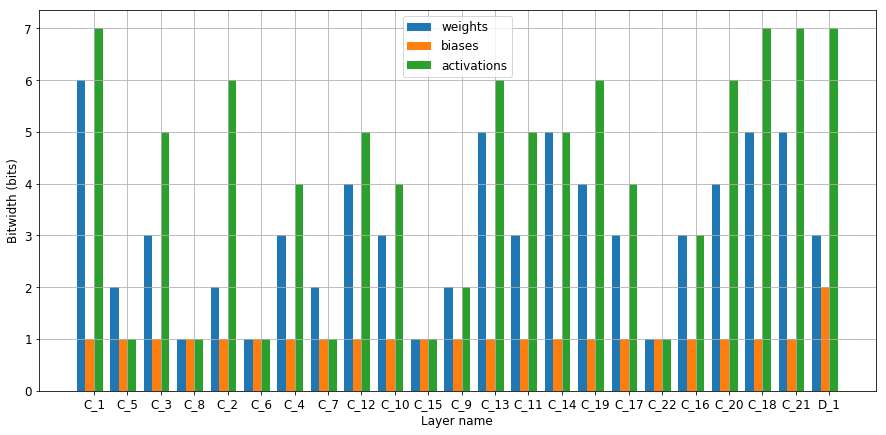}
        \caption{Optimal bitwidths $BW^*_{l, p}$ of parameters of each layer.}
        \label{fig:opts_df_svhn_bw}
    \end{subfigure}
    \caption{Results for acceptable and measured inference accuracy loss and the optimal bitwidths of the quantized CNN resulting from dependent optimized search using a linear acceptable loss allocation scheme on a 23-layer branched CNN trained on SVHN.}
    \label{fig:opts_results_svhn_df}
\end{figure*}

\begin{figure*}[t!]
    \centering
    \begin{subfigure}[t]{0.45\textwidth}
        \centering
        \includegraphics[width=\textwidth]{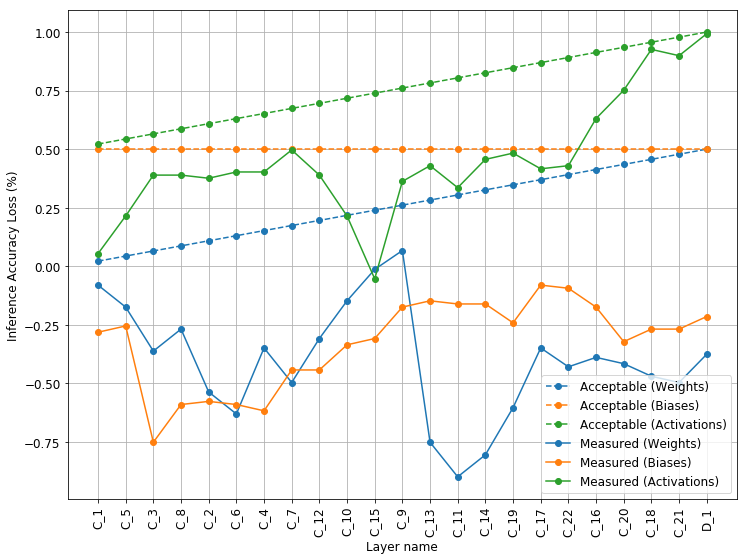}
        \caption{Acceptable and measured inference accuracy loss $\Delta a^D_{l, p}$ after sequentially quantizing the parameters $\mathbf{p}_l$ in the order ($\mathbf{W} \rightarrow \mathbf{B} \rightarrow \mathbf{A}$) from layers 1 to L.}
        \label{fig:opts_df_cifar10_acc_loss}
     \end{subfigure}
    \hspace{4mm}
    \begin{subfigure}[t]{0.47\textwidth}
        \centering
        \includegraphics[width=\textwidth]{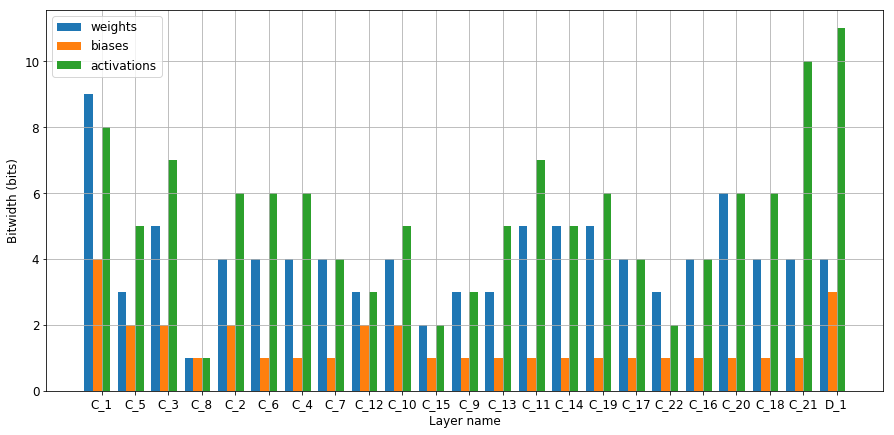}
        \caption{Optimal bitwidths $BW^*_{l, p}$ of parameters of each layer.}
        \label{fig:opts_df_cifar10_bw}
    \end{subfigure}
    \caption{Results for acceptable and measured inference accuracy loss and the optimal bitwidths of the quantized CNN resulting from dependent optimized search using a linear acceptable loss allocation scheme on a 23-layer branched CNN trained on CIFAR10.}
    \label{fig:opts_results_cifar10_df}
\end{figure*}

Fig.~\ref{fig:opts_results_fashion_seq} - Fig.~\ref{fig:opts_results_cifar10_df} present additional results for our final method. 

\subsection{Architectures of Branched and Sequential Model}
\begin{figure*}
    \centering
    \includegraphics[width=\textwidth]{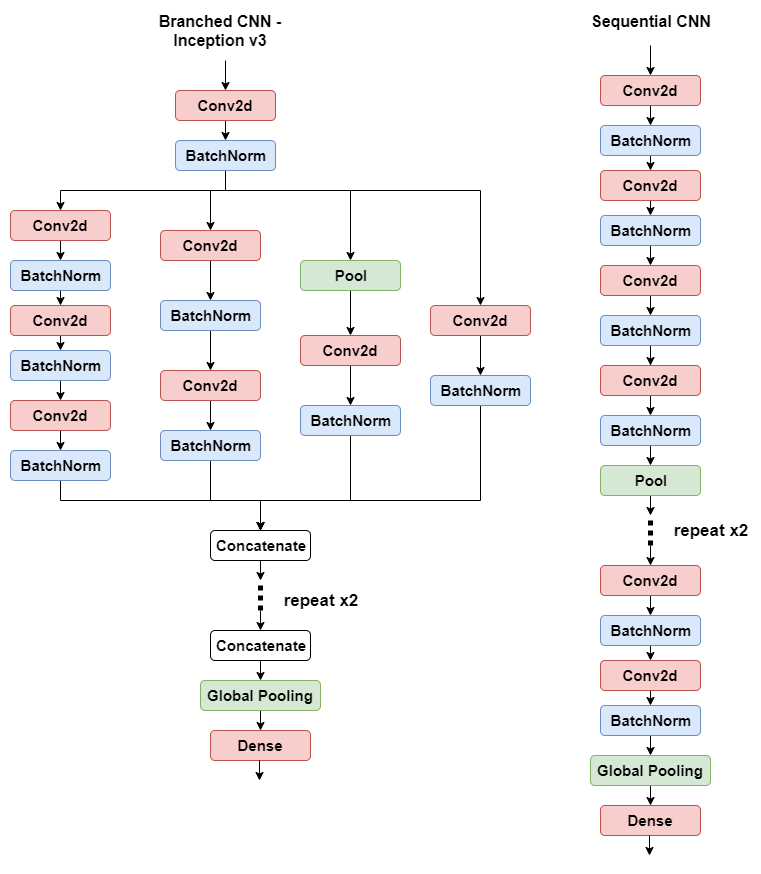}
    \caption{Architectures of the Branched model (left) and Sequential model (right) tested on the four datasets}
    \label{fig:architectures}
\end{figure*}

Visual illustrations of the architectures used for our experiments are shown in Fig.~\ref{fig:architectures}

\end{document}